%% file: main.tex
\documentclass[twocolumn]{article}

\usepackage[english]{babel}
\usepackage{microtype}

\usepackage{iccv}
\iccvfinalcopy

\usepackage{times}
\usepackage{epsfig}
\usepackage{graphicx}
\usepackage{amsmath}
\usepackage{amssymb}
\usepackage{tikz}

\usepackage{pgfplots}
\DeclareUnicodeCharacter{2212}{−}
\usepgfplotslibrary{groupplots,dateplot}
\usetikzlibrary{patterns,shapes.arrows}
\pgfplotsset{compat=newest}

\usepackage{bm}
\usepackage{enumitem}
\usepackage{acro}

\usepackage[export]{adjustbox}

\usepackage{array}
\usepackage{booktabs}
\usepackage{fontawesome}
\usepackage{multirow}
\usepackage{xcolor,colortbl}
\definecolor{Gray}{gray}{0.9}
\definecolor{Textgray}{gray}{0.4}
\newcolumntype{m}{>{\columncolor{Gray}}c}

\usepackage[breaklinks=true,letterpaper=true,colorlinks,bookmarks=false]{hyperref}

\newcommand{\PAR}[1]{\vskip4pt \noindent{\bf #1~}}

\DeclareAcronym{ood}{
  short = OoD,
  long  = out-of-distribution,
}
\DeclareAcronym{id}{
  short = ID,
  long  = in-distribution,
}
\DeclareAcronym{cnn}{
  short = CNN,
  long  = convolutional neural network,
}
\DeclareAcronym{ap}{
  short = AP,
  long  = average precision,
}
\DeclareAcronym{tpr}{
  short = TPR,
  long  = true positive rate,
}
\DeclareAcronym{fpr}{
  short = FPR,
  long  = false positive rate,
}
\DeclareAcronym{roc}{
  short = ROC,
  long  = receiver operating curve,
}
\DeclareAcronym{iou}{
  short = IoU,
  long  = intersection over union,
}
\DeclareAcronym{nn}{
  short = NN,
  long  = nearest neighbour,
}
\DeclareAcronym{yj}{
  short = J,
  long  = Youden's J,
}
\DeclareAcronym{mc}{
  short = MC,
  long  = Monte-Carlo,
}
\date{}

\begin{document}

\title{\vspace{-9mm}
The Fishyscapes Benchmark:\\Measuring Blind Spots in Semantic Segmentation
}

\author{Hermann Blum$^1$
\quad Paul-Edouard Sarlin$^2$
\quad Juan Nieto$^3$
\quad Roland Siegwart$^1$
\quad Cesar Cadena$^1$
}

\newcommand\blfootnote[1]{%
  \begingroup
  \renewcommand\thefootnote{}\footnote{#1}%
  \addtocounter{footnote}{-1}%
  \endgroup
}

\maketitle

\begin{abstract}
Deep learning has enabled impressive progress in the accuracy of semantic segmentation.
Yet, the ability to estimate uncertainty and detect failure is key for safety-critical applications like autonomous driving.
Existing uncertainty estimates have mostly been evaluated on simple tasks, and it is unclear whether these methods generalize to more complex scenarios.
We present Fishyscapes, the first public benchmark for anomaly detection in a real-world task of semantic segmentation for urban driving.
It evaluates pixel-wise uncertainty estimates towards the detection of anomalous objects.
We~adapt state-of-the-art methods to recent semantic segmentation models and compare uncertainty estimation approaches based on softmax confidence, Bayesian learning, density estimation, image resynthesis, as well as supervised anomaly detection methods.
%
%
Our results show that anomaly detection is far from solved even for ordinary situations, while our benchmark allows measuring advancements beyond the state-of-the-art.
Results, data and submission information can be found at \url{fishyscapes.com}.
\end{abstract}

\blfootnote{
\hspace{-14pt}$^1$ Autonomous Systems Lab, ETH Z\"urich\\
$^2$ Visual Geometry Group, ETH Z\"urich\\
$^3$ Microsoft Research\\
Correspondence: blumh@ethz.ch\\
This work was partially supported by the HILTI Group.}

\vspace{-8mm}
\section{Introduction}

Deep learning has had a high impact on the precision of computer vision methods~\cite{Chen2018-bp,he2017mask,Fu2018-kx,Sun2018-gu} and enabled semantic understanding in robotic applications~\cite{McCormac2018-ki,Florence2018-cd,Liang2018-ec}.
However, while these algorithms are usually compared on closed-world datasets with a fixed set of classes~\cite{Geiger2012-if,Cordts2016-qz}, the real-world is uncontrollable, and an incorrect reaction by an auto\-no\-mous agent to an unexpected input can have disastrous consequences~\cite{Bozhinoski2019-hx}.
    
\begin{figure}[t]
    \centering
\begin{tikzpicture}
\node[inner sep=0pt, label=\small{Input}] (rgb) at (0,0)
    {\includegraphics[width=.5\linewidth]{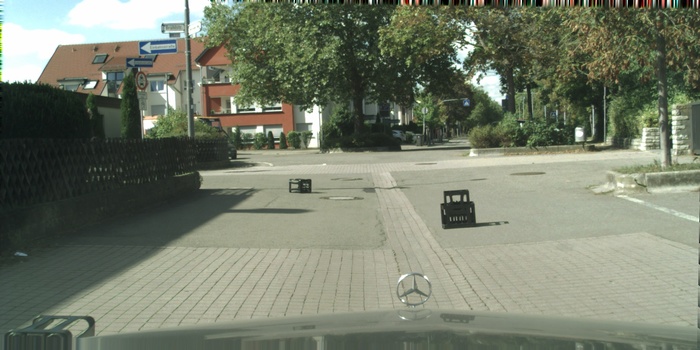}\hspace{-.5\linewidth}\includegraphics[width=.5\linewidth]{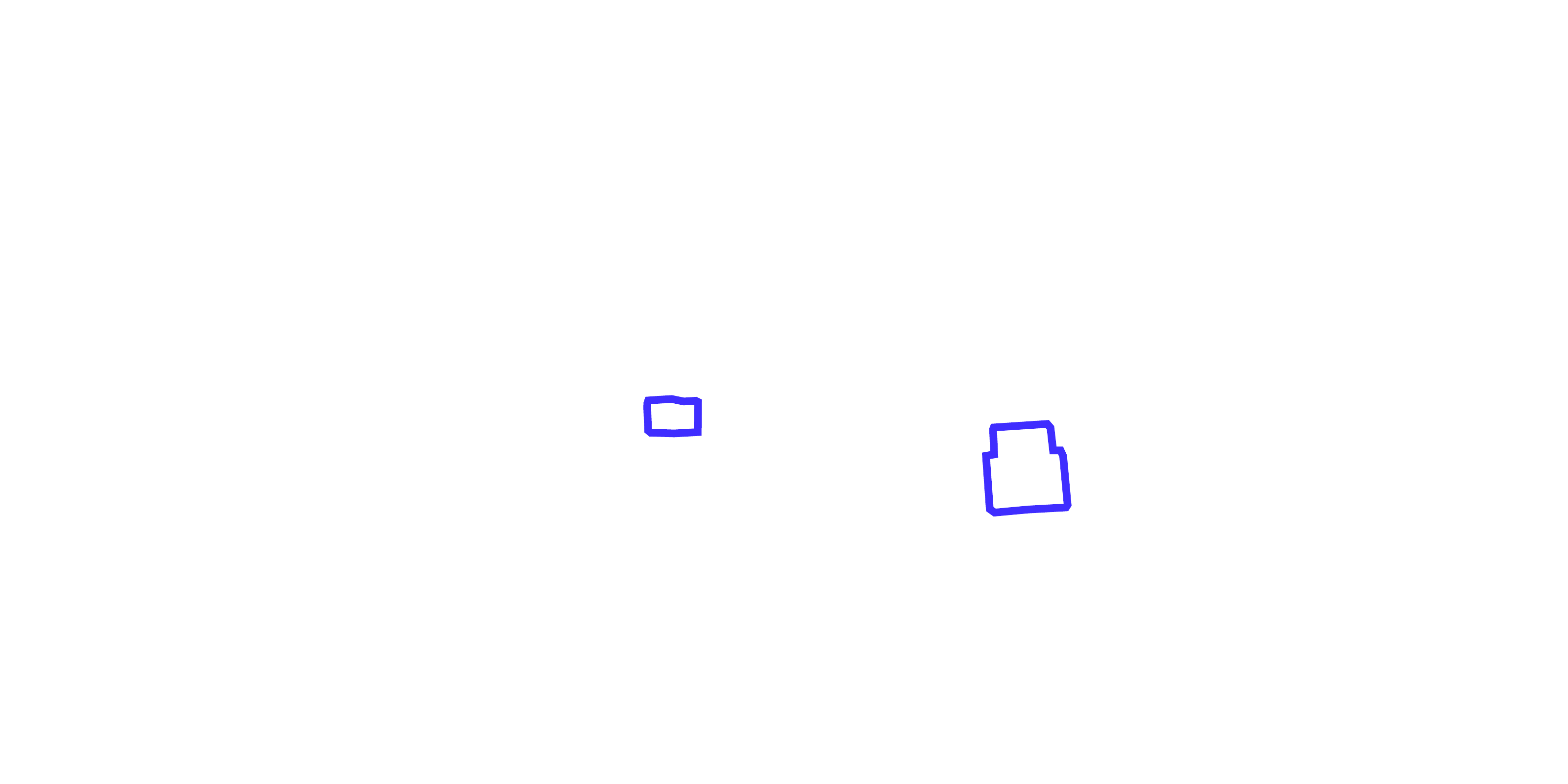}};
\node[inner sep=0pt, label=\small{Learned Embedding Density}] (mi) at (0.5\linewidth,0)
   {\includegraphics[width=.5\linewidth]{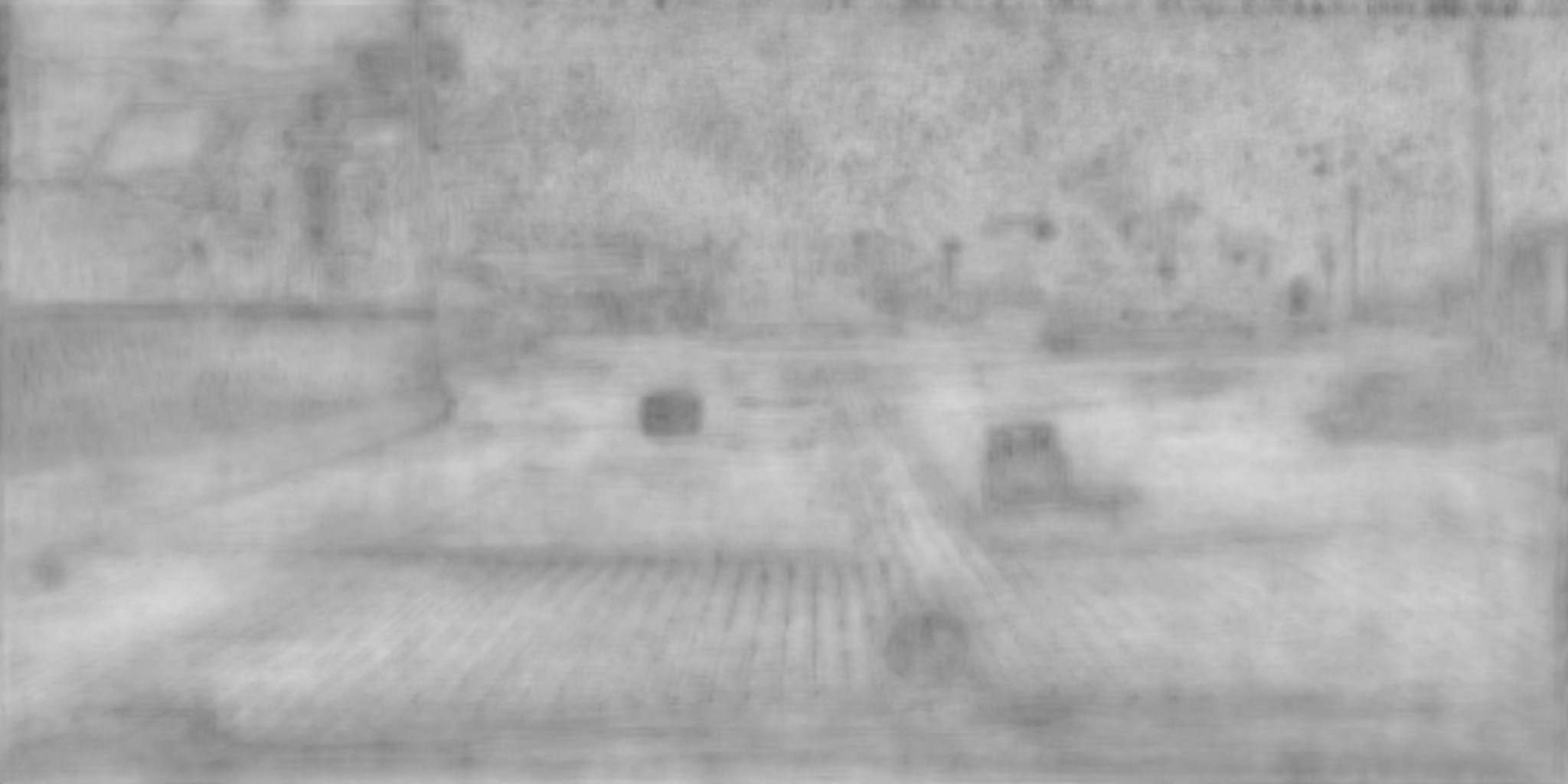}};
\node[inner sep=0pt, label=below:\small{Prediction}] (pred) at (0,-2.07)
    {\includegraphics[width=.5\linewidth]{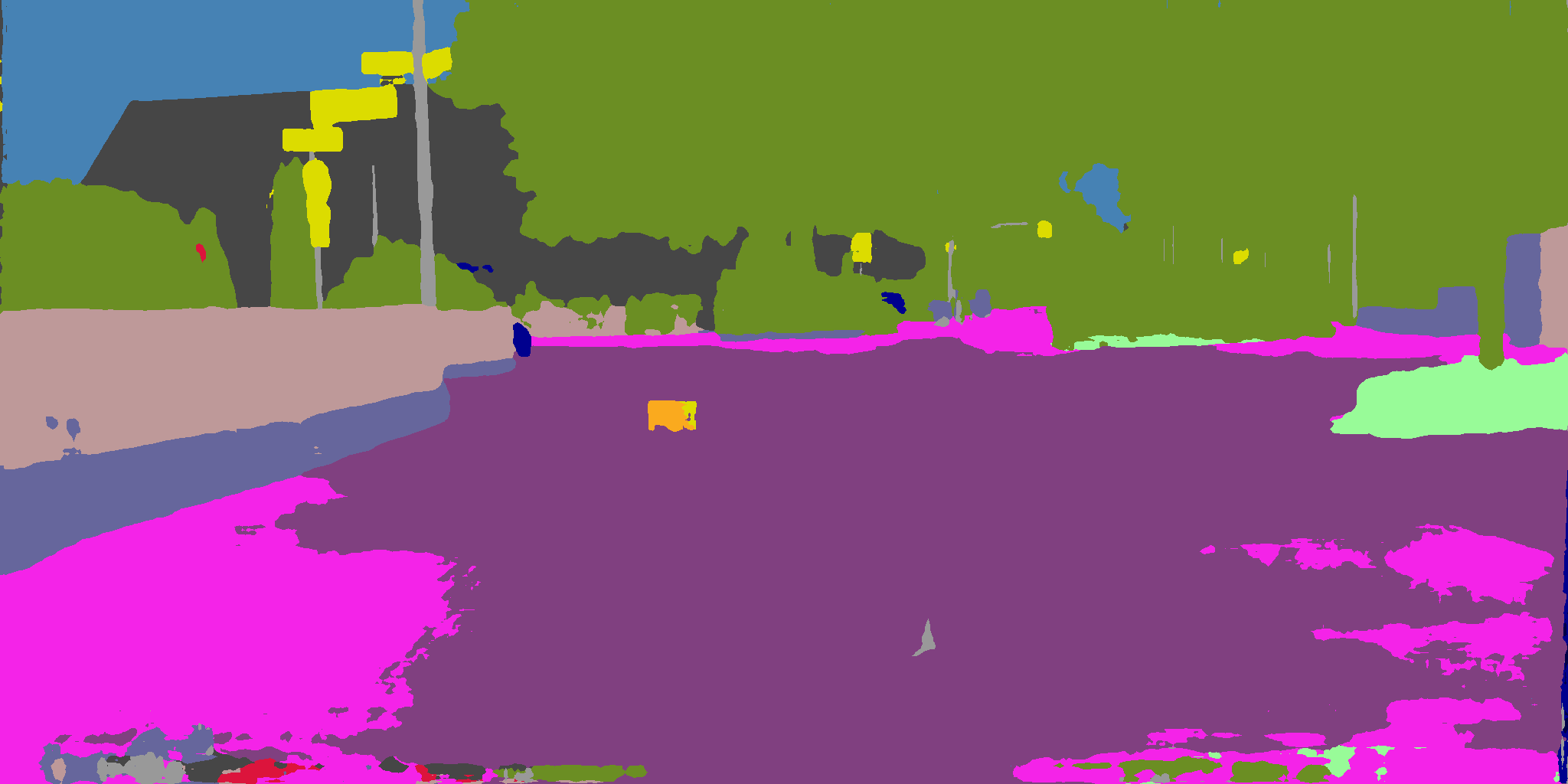}\hspace{-.5\linewidth}\includegraphics[width=.5\linewidth]{laf_0008_outline.pdf}};
\node[inner sep=0pt, white] (pred-capt) at (-1,-2.8)
    {\scriptsize{\textsf{streetsign}}};
\draw[->, white] (pred-capt) -- (-.4, -2.2);
\node[inner sep=0pt, white] (pred-capt2) at (1,-2.8)
    {\scriptsize{\textsf{road}}};
\draw[->, white] (pred-capt2) -- (.7, -2.45);
\node[inner sep=0pt, label=below:\small{MC Dropout}] (density) at (0.5\linewidth,-2.07)
    {\includegraphics[width=.5\linewidth]{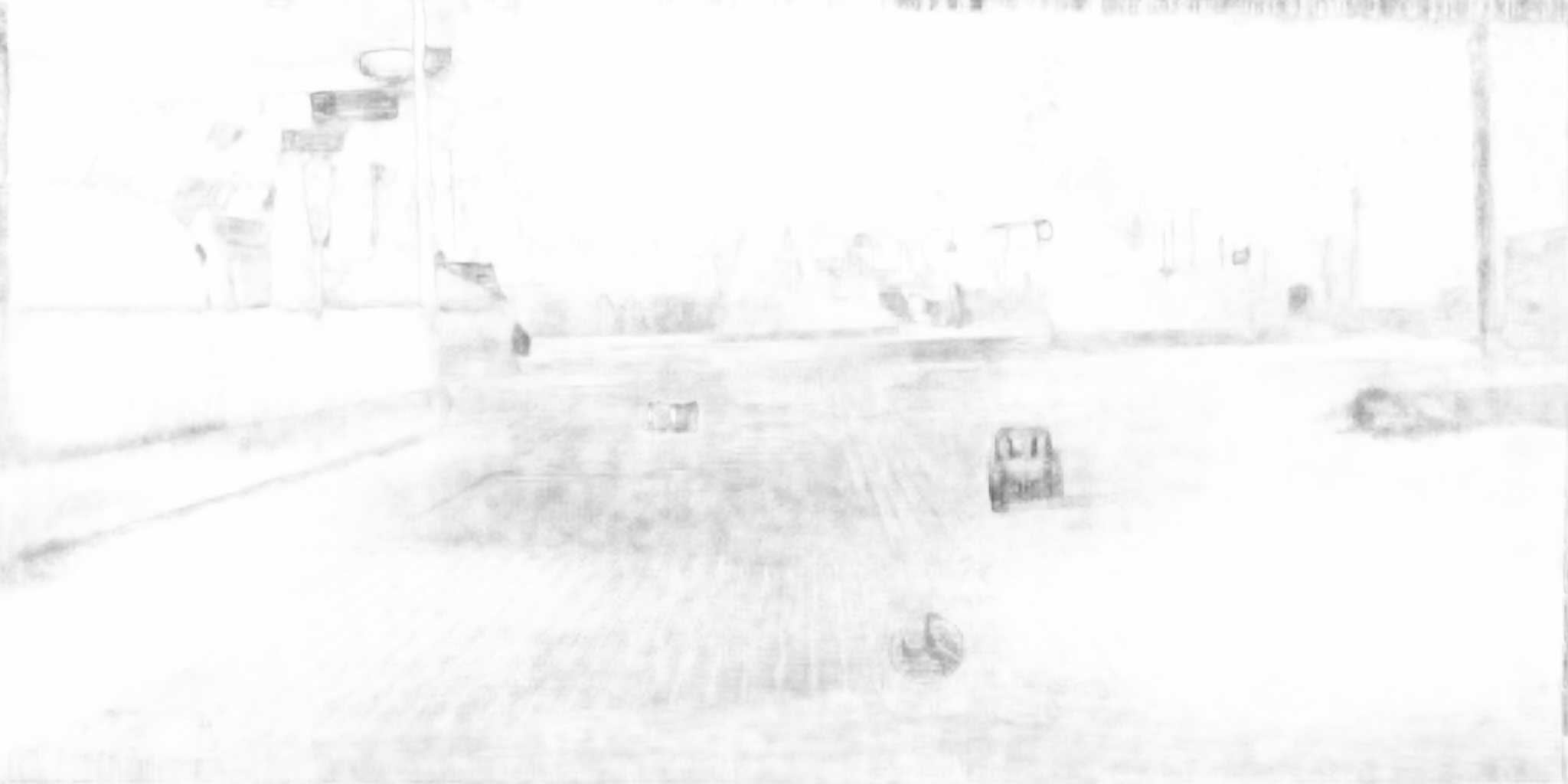}};
\end{tikzpicture}
    \caption{%
    When exposed to an object type unseen during training, a state-of-the-art semantic segmentation model~\cite{Chen2018-bp} predicts familiar labels (\emph{streetsign}, \emph{road}) with high confidence. To detect such failures, we evaluate various methods that assign a pixel-wise out-of-distribution score, where higher values are darker. The blue outline is added for illustration.
    }%
    \label{fig:teaser}
    \vspace{-4mm}
\end{figure}
    
As such, to reach full autonomy while ensuring safety and reliability, decision-making systems need information about outliers and uncertain or ambiguous cases that might affect the quality of the perception output.
As illustrated in Figure~\ref{fig:teaser}, deep \acp{cnn} react unpredictably for inputs that deviate from their training distribution.
In the presence of outlier objects, this is interpolated with the available classes at high confidence. 
Existing research to detect such behaviour is often labeled as \ac{ood}, anomaly, or novelty detection, and has so far focused on developing methods for image classification, evaluated on simple datasets like MNIST or CIFAR-10~\cite{Malinin2018-pl,Papernot2018-xz,Hendrycks2016-ua,Lee2018-si,pmlr-v80-ruff18a,golan,choi2018generative,sabokrou2018adversarially,pidhorskyi18}.
How these methods generalize to more elaborate network architectures and pixel-wise uncertainty estimation has not been assessed in prior work.

Motivated by these practical needs, we introduce `Fishy\-scapes', a benchmark that evaluates uncertainty estimates for semantic segmentation. The benchmark measures how well methods detect potentially hazardous anomalies in driving scenes.
Fishyscapes is based on data from Cityscapes~\cite{Cordts2016-qz}, a popular benchmark for semantic segmentation in urban driving.
Our benchmark consists of 
(i)~Fishyscapes Web, where images from Cityscapes are overlayed with objects that are regularly crawled from the web in an open-world setup,
and (ii)~Fishyscapes Lost \& Found, that builds up on a road hazard dataset collected with the same setup as Cityscapes~\cite{Pinggera2016-ki} and that we supplemented with labels.

To provide a broad overview, we adapt a variety of methods to semantic segmentation that were originally designed for image classification.
Because segmentation networks are much more complex and have high computational costs, this adaptation is not trivial, and we suggest different approximations to overcome these challenges.

Our experiments show that the embeddings of intermediate layers hold important information for anomaly detection. Based on recent work on generative models, we develop a novel method using density estimation in the embedding space. However, we also show that varying visual appearance can mislead feature-based and other methods. None of the evaluated methods achieves the accuracy required for safety-critical applications. We conclude that these remain open problems, with our benchmark enabling the community to measure progress and build upon the best performing methods so far.

To summarize, our contributions are the following:
\begin{itemize}[label={--},leftmargin=*]
    \item We introduce the first public benchmark evaluating pixel-wise uncertainty estimates in semantic segmentation, with a dynamic, self-updating dataset for anomaly detection.
    \item We report an extensive evaluation with diverse state-of-the-art approaches to uncertainty estimation, adapted to the semantic segmentation task, and present a novel me\-thod for anomaly detection.
    \item We show a clear gap between the alleged capabilities of established methods and their performance on this real-world task, thereby confirming the necessity of our benchmark to support further research in this direction.
\end{itemize}

\section{Related Work}
\label{sec:related_work}
Here we review the most relevant works in semantic segmentation and their benchmarks, and methods that aim at providing a confidence estimate of the output of deep networks.
\subsection{Semantic Segmentation}
\PAR{State-of-the-art models} are fully-convolutional deep networks trained with pixel-wise supervision. Most works~\cite{Ronneberger2015-uf,Badrinarayanan2017-rd,Chen2016-nv,Chen2018-bp} adopt an encoder-decoder architecture that initially reduces the spatial resolution of the feature maps, and subsequently upsamples them with learned transposed convolution, fixed bilinear interpolation, or unpooling.
Additionally, dilated convolutions or spatial pyramid pooling enlarge the receptive field and improve the accuracy.

\PAR{Popular benchmarks} compare methods on the segmentation of objects~\cite{Everingham2010-ci} and urban scenes. 
In the latter case, Cityscapes~\cite{Cordts2016-qz} is a well-established dataset depicting street scenes in European cities with dense annotations for a limited set of classes.
Efforts have been made to provide datasets with increased diversity, either in terms of environments, with WildDash~\cite{Zendel2018-ru}, which incorporates data from numerous parts of the world, or with Mapillary~\cite{Neuhold2017-ca}, which adds many more classes. Recent data releases add multi-sensor and multi-modality recordings on top of that~\cite{Sun2019-ca,Geyer2020-fk,Caesar2019-yd}.
Like ours, some datasets are explicitly derived from Cityscapes, the most relevant being Foggy Cityscapes~\cite{Sakaridis2018-iz}, which overlays synthetic fog onto the original dataset to evaluate more difficult driving conditions.
The Robust Vision Challenge\footnote{\url{http://www.robustvision.net/}} also assesses generalization of learned models across different datasets.

\PAR{Robustness and reliability} are only evaluated by these benchmarks through ranking methods according to their accuracy, without taking into accounts the uncertainty of their predictions.
Additionally, despite the fact that one cannot assume that models trained with closed-world data will only encounter known classes, these scenarios are rarely quantitatively evaluated. To our knowledge, WildDash~\cite{Zendel2018-ru} is the only public benchmark that explicitly reports uncertainty w.r.t. \ac{ood} examples. These are however drawn from a very limited set of full-image outliers, while we introduce a diverse set of objects, as WildDash mainly focuses on accuracy. Complementarily, the Dark Zurich dataset~\cite{Sakaridis2020-vi} allows for uncertainty-aware evaluation of semantic segmentation models with regard to deprived sensor inputs, i.e. evaluating aleatoric uncertainty.

Bevandic et al.~\cite{Bevandic2019-qp} experiment with \ac{ood} objects for semantic segmentation by overlaying objects on Cityscapes images in a manner similar to ours. They however assume the availability of a large \ac{ood} dataset, which is not realistic in an open-world context, and thus mostly evaluate supervised methods. In contrast, we assess a wide range of methods that do not require \ac{ood} data.
Mukhoti \& Gal~\cite{Mukhoti2018-af} introduce a new metric for uncertainty evaluation and are the first to quantitatively assess misclassification for segmentation. Yet they only compare few methods on normal \ac{id} data.
The MVTec benchmark~\cite{Bergmann2019-ho} compares a range of anomaly segmentation methods on images of single objects to find industrial production anomalies. It mostly compare methods that focus on low-power computing. Following our work, the CAOS benchmark~\cite{Hendrycks2019-jc} also compares anomaly segmentation methods in simulated and real-world driving scenes. While their results confirm our finding that most established methods scale poorly to semantic segmentation, their methodology lacks open-world testing, which we argue later is important for true anomaly detection.

\subsection{Uncertainty estimation}
There is a large body of work that aims at detecting \ac{ood} data or misclassification by defining uncertainty or confidence estimates.

\PAR{Probabilistic modeling} of a neural network's output is a straightforward approach in uncertainty estimation. The softmax score, i.e. the classification probability of the predicted class, was shown to be a first baseline~\cite{Hendrycks2016-ua}, although sensitive to adversarial examples~\cite{Goodfellow2014-fk}.
Its performance was improved by  ODIN~\cite{Liang2017-mj}, which applies noise to the input with the Fast Gradient Sign Method (FGSM)~\cite{Goodfellow2014-fk} and calibrates the score with temperature scaling~\cite{Guo2017-kg}. Probabilistic modelling has been extended further in Deep Belief Networks that propagate activation distributions throughout the network~\cite{Frey1999-eo,Loquercio2020-sh}.

\PAR{Bayesian deep learning} \cite{Gal2016-mx,Kendall2017-jy} adopts a probabilistic view by designing deep models whose outputs and weights are probability distributions instead of point estimates.
Uncertainties are then defined as dispersions of such distributions, and can be of several types. \emph{Epistemic} uncertainty, or model uncertainty, corresponds to the uncertainty over the model parameters that best fit the training data for a given model architecture.
As evaluating the posterior over the weights is intractable in deep non-linear networks, recent works perform \ac{mc} sampling with dropout~\cite{Gal2016-vm} or ensembles~\cite{Lakshminarayanan2017-zk}.
\emph{Aleatoric} uncertainty, or data uncertainty, arises from the noise in the input data, such as sensor noise. 
Both have been applied to semantic segmentation~\cite{Kendall2017-jy}, and successively evaluated for misclassification detection~\cite{Mukhoti2018-af}, but only on \ac{id} data and not for \ac{ood} detection. 
Malinin \& Gales~\cite{Malinin2018-pl} later single out  \emph{distributional} uncertainty to represent model misspecification with respect to \ac{ood} inputs.
Their approach however was only applied to image classifications on toy datasets, and requires \ac{ood} data during the training stage.
To address the latter constraint, Lee et al.~\cite{Lee2017-vv} earlier proposed a Generative Adversarial Network (GAN) that generates \ac{ood} data as boundary samples.
This is however very challenging to scale to complex and high-dimensional data like high-resolution images of urban scenes. Recently, Bayesian methods investigated the inductive bias of network structures beyond weights~\cite{Wilson2020-cc}. For example, \cite{Antoran2020-zs} extracts meaningful uncertainties from an `ensemble' of network activations at varying depth, and \cite{Yehezkel_Rohekar2019-ef} employs a sampling scheme for architectures.

\PAR{\ac{ood} and novelty detection} is often tackled by non-Bayesian approaches.
As such, feature introspection amounts to measuring discrepancies between distributions of deep features of training data and \ac{ood} samples, using either \ac{nn} statistics~\cite{Papernot2018-xz,Mandelbaum2017-ti} or Gaussian approximations~\cite{Lee2018-si,Van_Amersfoort2020-vq}.
These methods have the benefit of working on any classification model without requiring specific training. Recently, connections between feature density and Bayesian uncertainties have been investigated~\cite{Postels2020-zx}.
On the other hand, approaches specifically tailored to perform \ac{ood} detection include one-class classification~\cite{pmlr-v80-ruff18a,golan}, which aim at creating discriminative embeddings, density estimation~\cite{choi2018generative,nalisnick2018deep}, which estimate the likelihood of samples w.r.t to the true data distribution, and generative reconstruction~\cite{sabokrou2018adversarially,pidhorskyi18,Gong2019-zf}, which use the quality of auto-encoder reconstructions to discriminate \ac{ood} samples.
Richter et al.~\cite{Richter2017-wg} apply the latter to simple real images recorded by a robotic car and successfully detect new environments.

\input{qualitative_examples}

\section{Benchmark Design}
\label{sec:benchmark_design}
Because it is not possible to produce ground truth for uncertainty values, evaluating estimators is not a straightforward task. We thus compare them on the proxy classification task~\cite{Hendrycks2016-ua} of detecting anomalous inputs. The uncertainty estimates are seen as scores of a binary classifier that compares the score against a threshold and whose performance reflects the suitability of the estimated uncertainty for anomaly detection.

Such an approach however introduces a major issue for the design of a public \ac{ood} detection benchmark.
With publicly available \ac{id} training data $A$ and \ac{ood} inputs~$B$, it is not possible to distinguish between an uncertainty method that informs a classifier to discriminate $A$ from any other input, and a classifier trained to discriminate $A$ from $B$.
The latter option clearly does not represent progress towards the goal of general uncertainty estimation, but rather overfitting.

To this end, we (i)~only release a small validation set with associated ground truth masks, while keeping larger test sets hidden, (ii)~continuously evaluate submitted methods against a dynamically changing, synthetic dataset, and (iii)~compare the performance on the dynamic dataset with evaluations on real-world data. Additionally, all submissions to the benchmark must indicate whether any \ac{ood} data was used during training, which is cross-checked with linked publications.

Examples from all benchmark datasets are shown in figure~\ref{fig:data-overview}.

\subsection{Does the method work in an open world?}
\label{sec:dynamic_dataset}
The open world scenario describes the problem that an autonomous agent who is freely interacting with the world has to be able to deal with the unexpected at all times. To test perception methods in an open world scenario, a benchmark therefore needs to present truly unexpected inputs. We argue that this is never truly possible with a fixed dataset that by design has limited diversity, and over time may simply identify those methods that deal best with the kind of objects included in the dataset. Instead, we propose a dynamically changing dataset that samples diverse objects at every iteration. 

In general, there are three options to generate such dynamic datasets: At every iteration, one may (i) capture new data in the wild and annotate, (ii) render new objects in simulation, or (iii) capture new objects in the wild, but blend them into already annotated scenes. While data from the wild is essential to test methods in realistic settings, annotation for semantic segmentation is very expensive and not a sustainable way to generate new datasets multiple times per year. Between (ii) and (iii) there is an essential trade-off. Rendering in 3D ensures physically viable object placement and consistent lighting. Images of diverse objects in the wild are much better available than textured 3D models and can be blended into real-world scenes. We acknowledge that there is an ongoing debate whether photorealtistic rendering engines or modern blending techniques achieve more realistic images, which was touched upon by a response-work to this benchmark~\cite{Hendrycks2019-jc}. In this work, we decided to base our dataset FS Web on approach (iii).
In the following, we describe a blending-based reference dataset FS Static and the dynamically changing dataset FS Web.

\PAR{FS Static} is based on the validation set of Cityscapes~\cite{Cordts2016-qz}. It has a limited visual diversity, which is important to make sure that it contains none of the overlayed objects.
In addition, background pixels originally belonging to the void class\footnote{\emph{void} in cityscapes is defined as: forms of horizontal ground-level structures that do not match any class, things that might not be there anymore the next day/hour/minute (e.g. movable trash bin, buggy, bag, wheelchair, animal), clutter in the background that is not distinguishable, or any objects that do not match a class (e.g. visible parts of the ego vehicle, mountains, street lights, back side of signs).} are excluded from the evaluation, as they may be borderline \ac{ood}.
Anomalous objects are extracted from the ge\-ne\-ric Pascal VOC~\cite{Everingham2010-ci} dataset using the associated segmentation masks. We only overlay objets from classes that cannot be found in Cityscapes: \textit{aeroplane, bird, boat, bottle, cat, chair, cow, dog, horse, sheep, sofa, tvmonitor}.
Objects cropped by the image borders or objects that are too small to be seen are filtered out.
We randomly size and position the objects on the underlying image, making sure that none of the objects appear on the ego-vehicle.
Objects from mammal classes have a higher probability of appearing on the lower-half of the screen, while classes like birds or airplanes have a higher probability for the upper half.
The placing is not further limited to ensure each pixel in the image, apart from the ego-vehicle, is comparably likely to be anomalous.
To match the image characteristics of cityscapes, we employ a series of postprocessing steps similar to those described in~\cite{Abu_Alhaija2018-dp}, without those steps that require 3D models of the objects to e.g. adapt shadows and lighting.
To make the task of anomaly detection harder, we add synthetic fog~\cite{foggycityscapes1,foggycityscapes2} on the in-distribution pixels with a per-image probability.
This prevents fraudulent methods to compare the input against a fixed set of Cityscapes images.
The dataset is split into a minimal public validation set of 30 images and a hidden test set of 1000 images. It contains in total around 4.5e7 \ac{ood} and 1.8e9 \ac{id} pixels. The validation set only contains a small disjoint set of pascal objects to prevent few-shot learning on our data creation method.

\begin{figure}[t]
    \centering
\begin{tikzpicture}
\node[rotate=90] at (-.27\linewidth,0) {\small\strut no blending};
\node[rotate=90] at (-.27\linewidth,-2.4) {\small\strut blending v1};
\node[rotate=90] at (-.27\linewidth,-4.8) {\small\strut blending v2};

\node[inner sep=0pt] at (0,0)
    {\adjincludegraphics[width=.45\linewidth,trim={{.2\width} {.05\height} {.3\width} {.3\height}},clip]{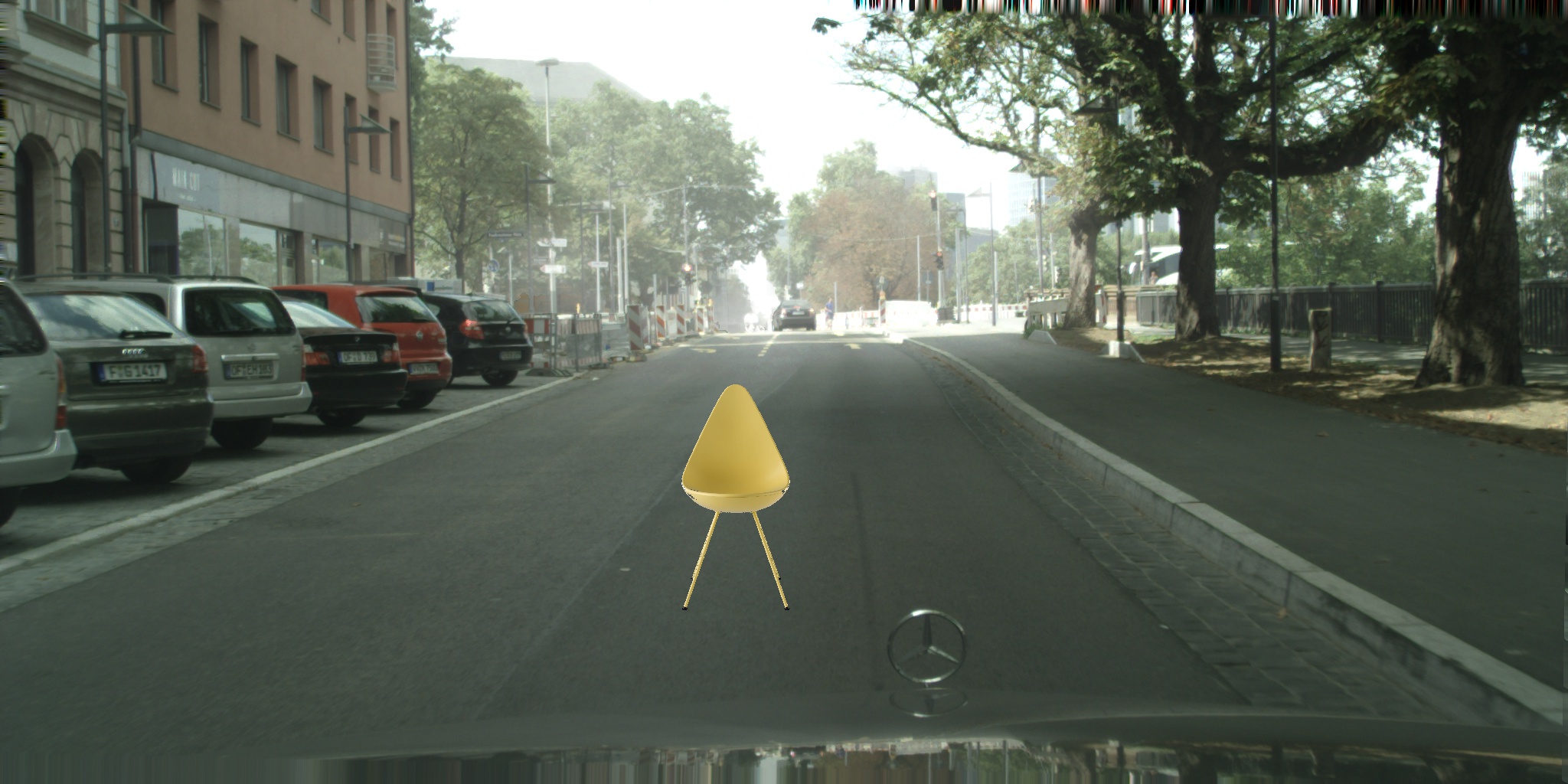}};
\node[inner sep=0pt] at (0.45\linewidth,0)
   {\adjincludegraphics[width=.45\linewidth,trim={{.2\width} {.05\height} {.3\width} {.3\height}},clip]{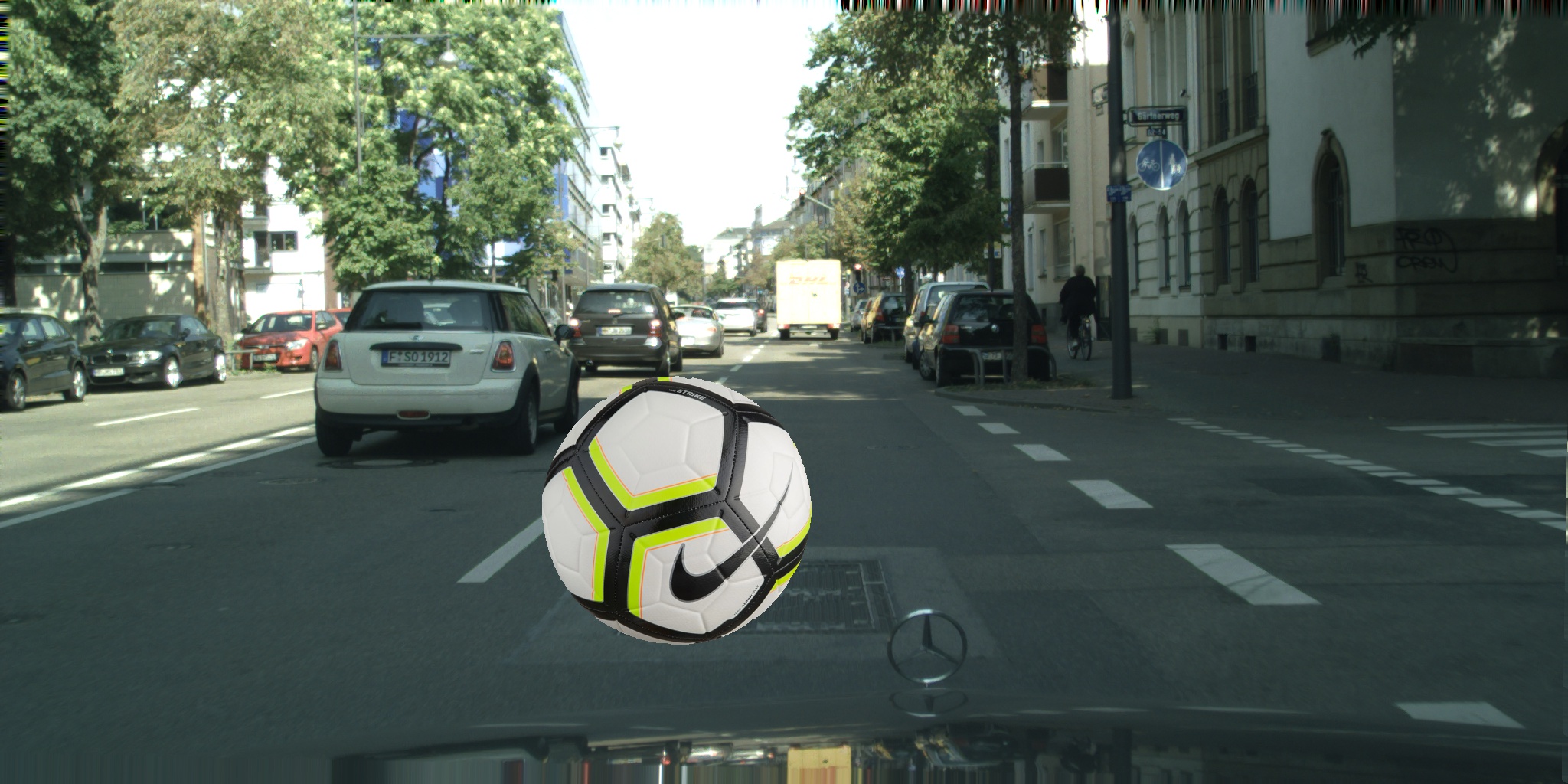}};
\node[inner sep=0pt] at (0,-2.4)
    {\adjincludegraphics[width=.45\linewidth,trim={{.2\width} {.05\height} {.3\width} {.3\height}},clip]{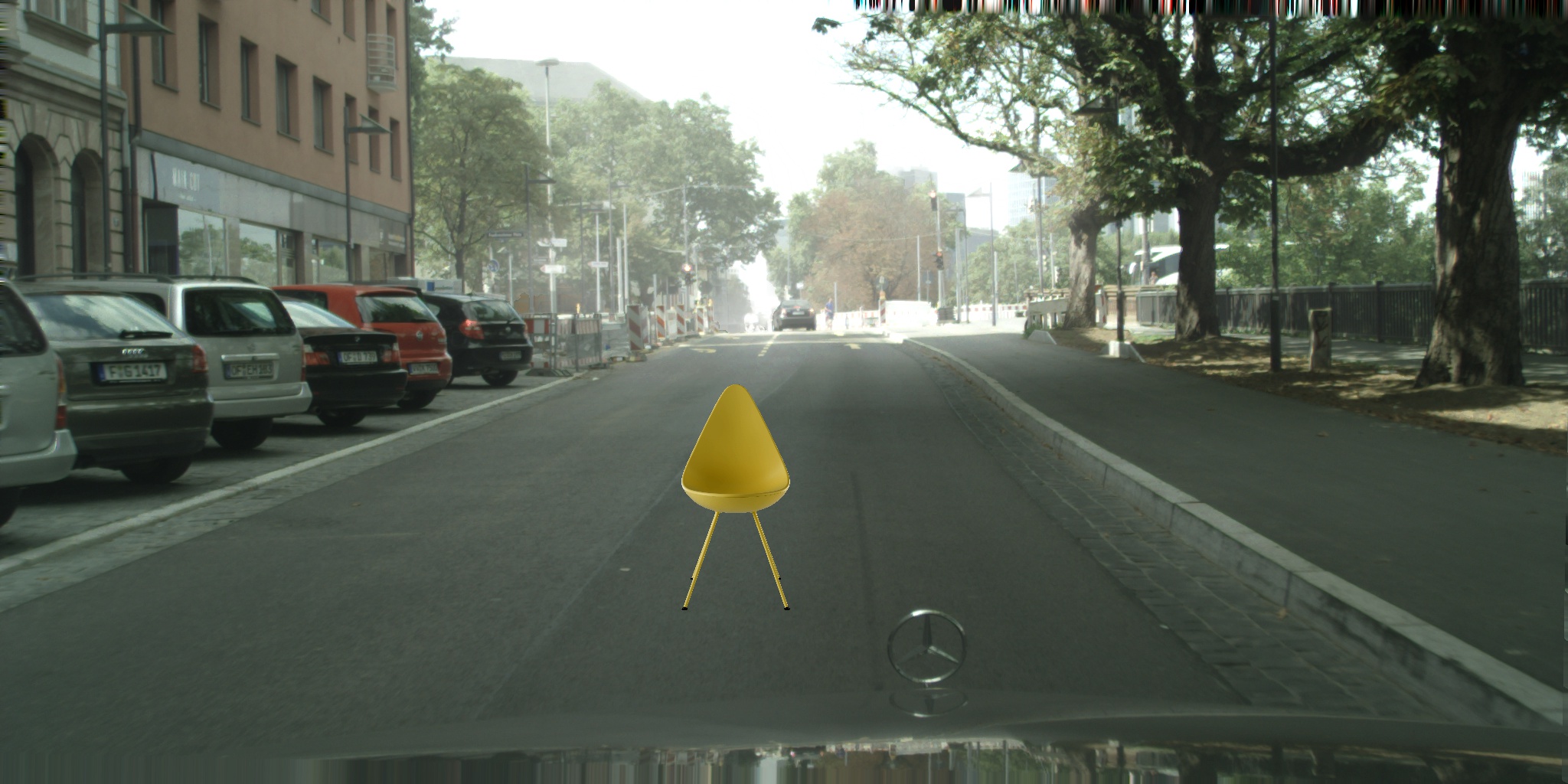}};
\node[inner sep=0pt] at (0.45\linewidth,-2.4)
   {\adjincludegraphics[width=.45\linewidth,trim={{.2\width} {.05\height} {.3\width} {.3\height}},clip]{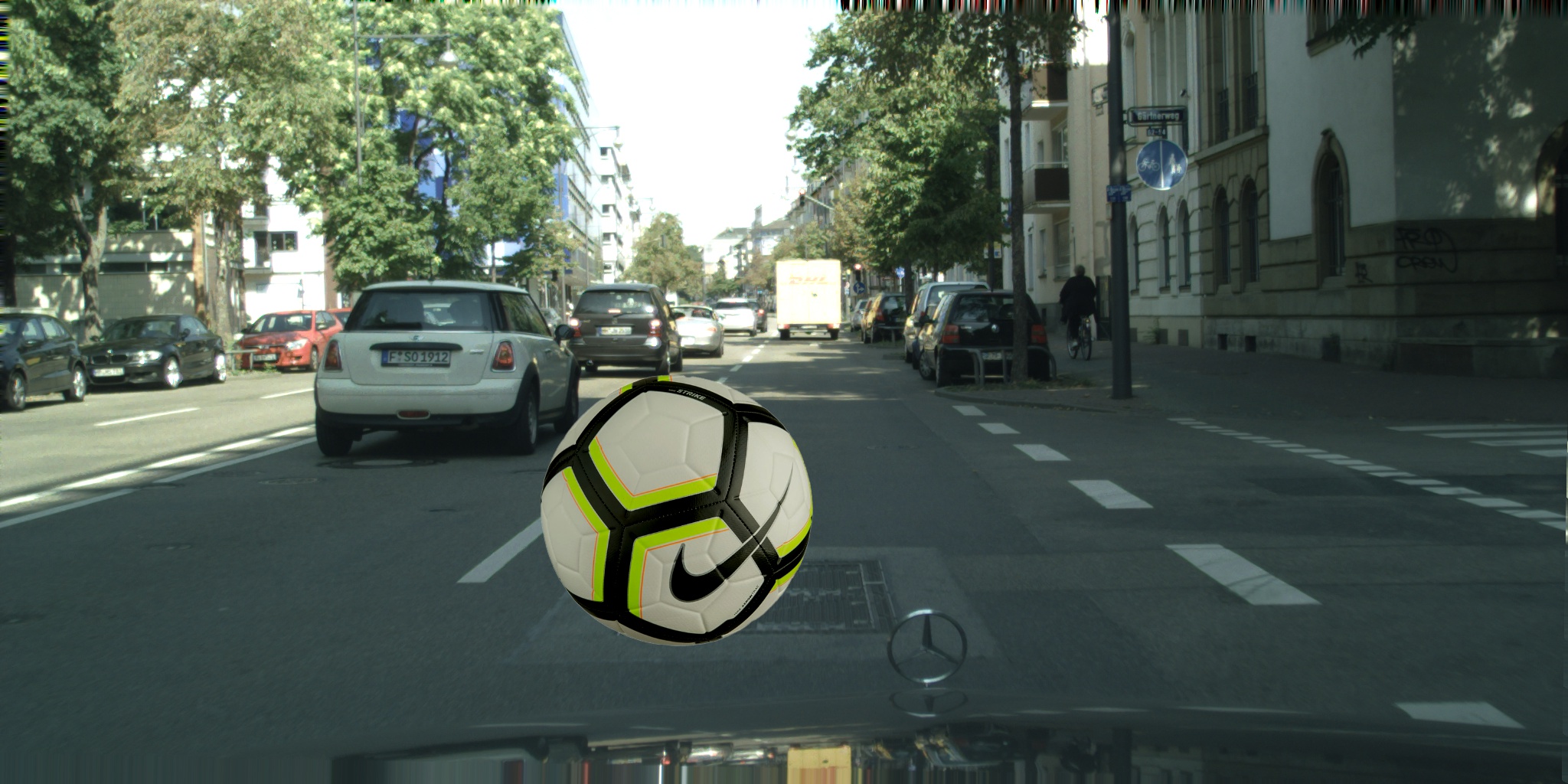}};
   \node[inner sep=0pt] at (0,-4.8)
    {\adjincludegraphics[width=.45\linewidth,trim={{.2\width} {.05\height} {.3\width} {.3\height}},clip]{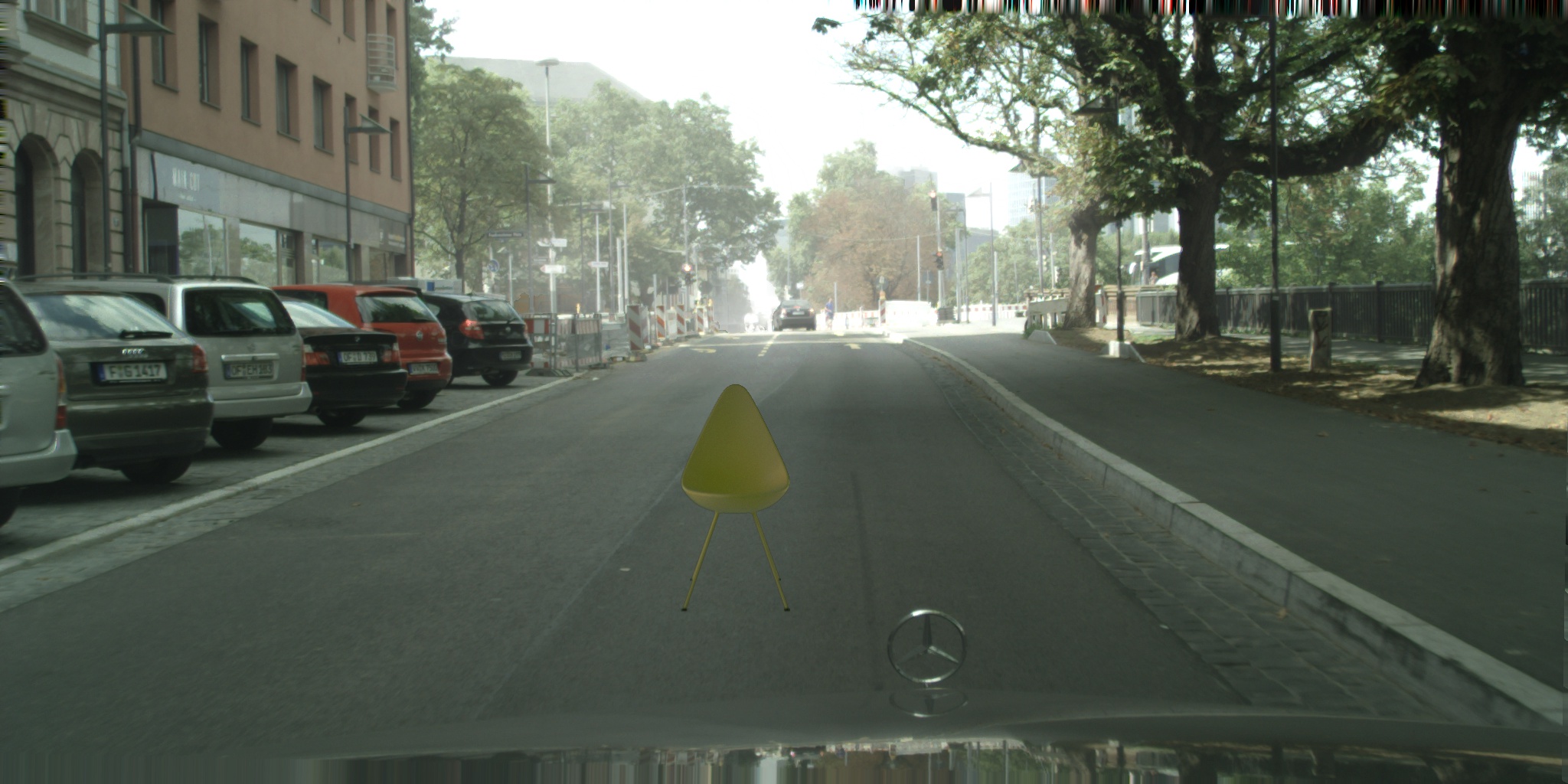}};
\node[inner sep=0pt] at (0.45\linewidth,-4.8)
   {\adjincludegraphics[width=.45\linewidth,trim={{.2\width} {.05\height} {.3\width} {.3\height}},clip]{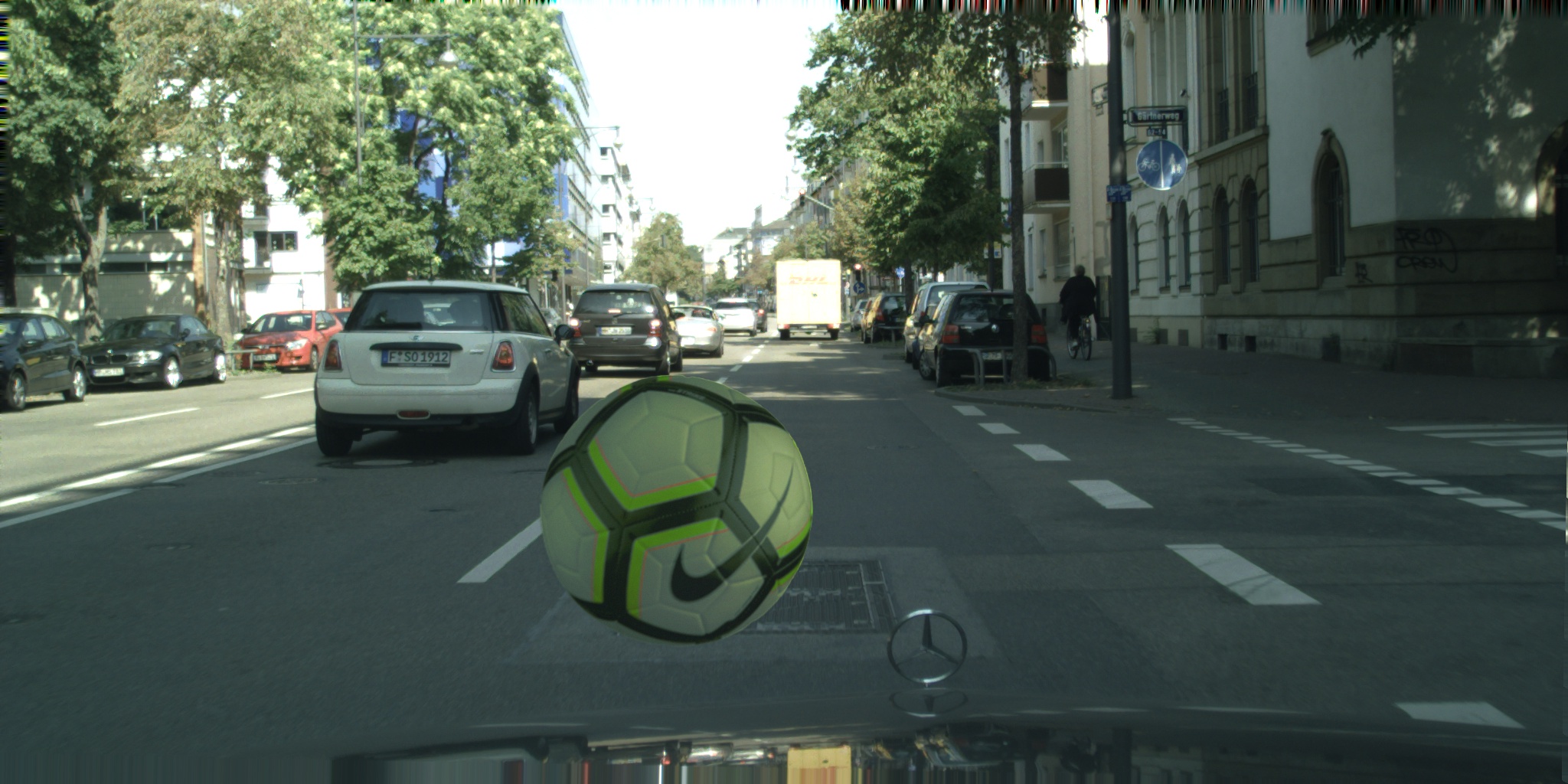}};
\end{tikzpicture}
    \caption{%
    Illustration of the blending process and improvements (v2) applied in June 2019. While color adaptation to the predominantly gray Cityscapes images is visually most obvious, important improvements in v2 include depth and motion blur, as well as glow effects.
    }%
    \label{fig:blending}
    \vspace{-5mm}
\end{figure}

\PAR{FS Web} is built similarly to FS Static, but with overlay objects crawled from the internet using a changing list of keywords.
Our script searches for images with transparent background, uploaded in a recent timeframe, and filters out images that are too small.
The only manual process is filtering out images that are not suitable, e.g. with decorative borders or watermarks. The dataset for March 2019 contains 4.9e7 \ac{ood} and 1.8e9 \ac{id} pixels.
As the diversity of images and color distributions for the images from the web is much greater than those from Pascal VOC, we also adapt our overlay procedure. In total, we follow these steps, some of which were added from June 2019 onwards (marked with *):
\begin{itemize}[label={--},leftmargin=*, nosep]
    \item in case the image does not already have a smooth alpha channel, smooth the mask of the objects around the borders for a small transparency gradient
    \item adapt the brightness of the object towards the mean brightness of the overlayed pixels
    \item apply the inverse color histogram of the Cityscapes image to shift the color distribution towards the one found on the underlying image*
    \item radial motion blur*
    \item depth blur based on the position in the image*
    \item color noise
    \item glow effects to simulate overexposure*
\end{itemize}
Figure~\ref{fig:blending} shows an illustration of the blending results.

As discussed, the blending process is part of a trade-off to make an open-world dataset feasible. To further ensure that methods do not overfit to any artifacts created by the blending process, but detect anomalies based on their semantics and appearance, we include a sample of \ac{id} objects in the blending dataset. For this, we create a database from objects in the Cityscapes training dataset (\emph{car, person, truck, bus, train, bike}) where we manually filter out any occluded instances. We then decide at random for every image whether to blend an anomalous object or a Cityscapes object, where we skip random placement and histogram adaptation for the latter. This addition was introduced in FS Web Jan 2020. An example can be seen in figure~\ref{fig:data-overview}.

As indicated, the postprocessing was improved between iterations of the dataset. Because the purpose of the FS Web dataset is to measure any possible overfitting of the methods through a dynamically changing dataset, we will continue to refine also this image overlay procedure, updating our method with recent research results. Any update to the blending is also applied to the FS Static validation set, allowing submissions to validate the effect of blending improvements.

\subsection{Does the method work on real images?}

As discussed in section~\ref{sec:dynamic_dataset}, capturing and annotating driving scenes multiple times per year is not sustainable, which made it necessary to use synthetic data generation for the dynamic dataset. However, for safe deployment it is equally important to test methods under real-world conditions. This is the purpose of the FS Lost \& Found dataset in our benchmark.

\PAR{FS Lost \& Found} is based on the original Lost \& Found dataset~\cite{Pinggera2016-ki}. However, the original dataset only includes annotations for the anomalous objects and a coarse annotation of the road. It does not allow for appropriate evaluation of anomaly detection, as objects and road are very distinct in texture and it is more challenging to evaluate the anomaly score of the objects compared to eg. building structures. In order to make use of the full image, we add pixel-wise annotations that distinguish between \emph{objects} (the anomalies), \emph{background} (classes contained in Cityscapes) and \emph{void} (anything not contained in Cityscapes classes that still appears in the training images). Additionally, we filter out those sequences where the `road hazards' are children or bikes, because these are part of regular Cityscapes data and not anomalies. We subsample the repetitive sequences, labelling at least every sixth image, and remove images that do not contain objects. In total, we present a public validation set of 100 images and a testset of 275 images, based on disjoint sets of locations.\\
While the Lost \& Found images were captured with the same setup as Cityscapes, the distribution of street scenery is very different. The images were captured in small streets of housing areas, industrial areas, or on big parking lots. The anomalous objects are usually very small and are not equally distributed on the image. Nevertheless, the dataset allows to test for real images as opposed to synthetic data, therefore preventing any overfitting on synthetic image processing. This is especially important for parameter tuning on the validation set.

\subsection{Metrics}

We consider metrics associated with a binary classification task. Since the \ac{id} and \ac{ood} data is unbalanced, metrics based on the \ac{roc} are not suitable~\cite{Saito2015-du}.
We therefore base the ranking and primary evaluation on the \ac{ap}.
However, as the number of false positives in high-recall areas is particularly relevant for sa\-fe\-ty-critical applications, we additionally report the false positive rate at 95\% recall ($\textrm{FPR}_\textrm{95}$). This metric was also used in~\cite{Hendrycks2016-ua} and emphasizes safety.

Semantic classification is not the goal of our benchmark, but uncertainty estimation and outlier detection should not come at high cost of segmentation accuracy.
We therefore additionally report the mean \ac{iou} of the semantic segmentation on the Cityscapes validation set.

For safety-critical systems, it is not only important to detect anomalies, but also to be fast enough to allow for a reaction. We therefore report the inference time of joint segmentation and anomaly detection per single frame. Times are measured over 500 images of the Cityscapes validation set on a GeForce 1080 Ti GPU.

\section{Evaluated Methods}
We now present the methods that are evaluated in Fishy\-scapes. In a first part, we describe the existing baselines and how we adapted them to the task of semantic segmentation. We then propose a novel method based on learned embedding density. Finally, we list those methods that were submitted to the public benchmark so far.
All approaches are applied to the state-of-the-art semantic segmentation model DeepLab-v3+~\cite{Chen2018-bp}. Further implementation details are listed in the supplementary material.

\subsection{Baselines}
\label{sec:method:baselines}
\PAR{Softmax.} The maximum softmax probability is a commonly used baseline and was evaluated in~\cite{Hendrycks2016-ua} for \ac{ood} detection.
We apply the metric pixel-wise and additionally measure the softmax entropy, as proposed by~\cite{Lee2017-vv}, which captures more information from the softmax.

\PAR{OoD training.} While we generally strive for methods that are not biased by data, learning confidence from data is an obvious baseline and was explored in~\cite{DeVries2018-at}. As we are not supposed to know the true \ac{ood} distribution, we do not use Pascal VOC, but rather approximate unknown pixels with the Cityscapes \emph{void} class.
In our evaluation, we (i)~train a model to maximise the softmax entropy for \ac{ood} pixels, or (ii)~introduce void as an additional output class and train with it.
The uncertainty is then measured as (i)~the softmax entropy, or (ii)~the score of the void class.

\PAR{Bayesian DeepLab} was introduced by Mukhoti \& Gal~\cite{Mukhoti2018-af}, following Kendall \& Gal~\cite{Kendall2017-jy}, and is the only uncertainty estimate already applied to semantic segmentation in the literature. The epistemic uncertainty is modeled by adding Dropout layers to the encoder, and approximated by $T$ \ac{mc} samples, while the aleatoric uncertainty corresponds to the spread of the categorical distribution. The total uncertainty is the predictive entropy of the distribution $\*y$,
\begin{equation}
    \mathbb{\hat{H}}\left[\*y|\*x\right] = 
        -\sum_c\left(\frac{1}{T}\sum_t y_c^t\right)\log\left(\frac{1}{T}\sum_t y_c^t\right),
\end{equation}
where $y_c^t$ is the probability of class $c$ for sample $t$. The epistemic uncertainty is measured as the mutual information (MI) between $\*y$ and the weights $\*w$,
\begin{equation}
    \mathbb{\hat{I}}\left[\*y, \*w | \*x\right] = 
        \mathbb{\hat{H}}\left[\*y|\*x\right]
        - \frac{1}{T}\sum_{c, t} y_c^t\log y_c^t.
\end{equation}

\PAR{Dirichlet DeepLab.} Prior Networks~\cite{Malinin2018-pl} extend the framework of~\cite{Gal2016-mx} by considering the predicted logits $\*z$ as log concentration parameters $\bm{\alpha}$ of a Dirichlet distribution, which is a prior of the predictive categorical distribution $\*y$. Intuitively, the spread of the Dirichlet prior should model the distributional uncertainty, and remain separate from the data uncertainty modelled by the spread of the categorical distribution. To this end, Malinin \& Gales~\cite{Malinin2018-pl} advocate to train the network with the objective:
\begin{equation}
\begin{split}
    \mathcal{L(\theta)}
    &= \mathbb{E}_{p_{\mathrm{in}}}\left[
        \mathrm{KL}\left[
            \mathrm{Dir}(\bm{\mu}|\bm{\alpha}_\mathrm{in})
            ||p(\bm{\mu}|\*x; \bm{\theta})\right]
        \right] \\
    &+ \mathbb{E}_{p_{\mathrm{out}}}\left[
        \mathrm{KL}\left[
            \mathrm{Dir}(\bm{\mu}|\bm{\alpha}_\mathrm{out})
            ||p(\bm{\mu}|\*x; \bm{\theta})\right]
        \right] \\
    &+ \mathrm{CrossEntropy}(\*y, \*z).
\end{split}
\label{eq:dirichlet}
\end{equation}
The first term forces \ac{id} samples to produce sharp priors with a high concentration $\bm{\alpha}_\mathrm{in}$, computed as the product of smoothed labels and a fixed scale $\alpha_0$. The second term forces \ac{ood} samples to produce a flat prior with $\bm{\alpha}_\mathrm{out}=\bm{1}$, effectively maximizing the Dirichlet entropy, while the last one helps the convergence of the predictive distribution to the ground truth. We model pixel-wise Dirichlet distributions, approximate \ac{ood} samples with void pixels, and measure the Dirichlet differential entropy.

\PAR{kNN Embedding.} Different works~\cite{Papernot2018-xz,Mandelbaum2017-ti} estimate uncertainty using kNN statistics between inferred embedding vectors and their neighbors in the training set. They then compare the classes of the neighbors to the prediction, where discrepancies indicate uncertainty.
In more details, a given trained encoder maps a test image $\*x'$ to an embedding $\*z'_l=\*f_l(\*x')$ at layer $l$, and the training set $\*X$ to a set of neighbors $\*Z_l := \*f_l(\*X)$.
Intuitively, if $\*x'$ is \ac{ood}, then $\*z'$ is also differently distributed and has e.g. neighbors with different classes.
Adapting these methods to semantic segmentation faces two issues: (i)~The embedding of an intermediate layer of DeepLab is actually a map of embeddings, resulting in more than 10,000 kNN queries for each layer, which is computationally infeasible. We follow~\cite{Mandelbaum2017-ti} and pick only one layer, selected using the FS Lost \& Found validation set.
(ii)~The embedding map has a lower resolution than the input and a given training embedding $\*z_l^{(i)}$ is therefore not associated with one, but with multiple output labels. As a baseline approximation, we link $\*z_l^{(i)}$ to all classes in the associated image patch. 
The relative density~\cite{Mandelbaum2017-ti} is then:
\begin{equation}
    D(\*z') = \frac{
        \sum \limits_{i \in K, c' = c_i} \exp \left( - \frac{\*z'\*z^{(i)}}{|\*z'|\, |\*z^{(i)}|}\right)
    }{
        \sum \limits_{i \in K} \exp \left( - \frac{\*z'\*z^{(i)}}{|\*z'|\, |\*z^{(i)}|}\right)
    }.
\end{equation}
Here, $c_i$ is the class of $\*z^{(i)}$ and $c'$ is the class of $\*z'$ in the downsampled prediction. In contrast to~\cite{Mandelbaum2017-ti}, we found that the cosine similarity from~\cite{Papernot2018-xz} works well without additional losses.
Finally, we upsample the density of the feature map to the input size, assigning each pixel a density value.

As the class association is unclear for encoder-decoder architectures, we also evaluate the density estimation with k neighbors independent of the class:
\begin{equation}
    D(\*z') = \sum \limits_{i \in K} \exp \left( - \frac{\*z'\*z^{(i)}}{|\*z'|\, |\*z^{(i)}|}\right).
\end{equation}
This assumes that an \ac{ood} sample $\*x'$, with a low density w.r.t $\*X$, should translate into $\*z'$ with a low density w.r.t. $\*Z_l$.

\subsection{Learned Embedding Density}
We now introduce a novel approach that takes inspiration from density estimation methods while greatly improving their scalability and flexibilty.

Density estimation using k\ac{nn} has two weaknesses. First, the estimation is a very coarse isotropic approximation, while the distribution in feature space might be significantly more complex. Second, it requires to store the embeddings of the entire training set and to run a large number of \ac{nn} searches, both of which are costly, especially for large input images. On the other hand, recent works~\cite{choi2018generative,nalisnick2018deep} on OoD detection leverage more complex generative models, such as normalizing flows~\cite{Dinh2016-sl,Kingma2018-jp,Dinh2014-mx}, to directly estimate the density of the input sample $\*x$. This is however not directly applicable to our problem, as (i) learning generative models of images that can capture the entire complexity of e.g. urban scenes is still an open problem; and (ii) the pixel-wise density required here should be conditioned on a very  (ideally infinitely) large context, which is computationally intractable.

Our approach mitigates these issues by learning the density of $\*z$. We start with a training set $\*X$ drawn from the unknown true distribution $\*x \sim p^*(\*x)$, and corresponding embeddings $\*Z_l$. A normalizing flow with parameters $\bm\theta$ is trained to approximate $p^*(\*z_l)$ by minimizing the negative log-likelihood (NLL) over all training embeddings in $\*Z_l$:
\begin{equation}
    \mathcal{L}(\*Z_l) = -\frac{1}{|\*Z_l|}
        \sum_i \log p_{\bm\theta}(\*z_l^{(i)}).
\end{equation}
The flow is composed of a bijective function $\*g_{\bm\theta}$ that maps an embedding $\*z_l$ to a latent vector $\bm \eta$ of identical dimensionality and with Gaussian prior $p(\bm\eta) = \mathcal N(\bm\eta;0,\*I)$. Its loglikelihood is then expressed as
\begin{equation}
    \log p_{\bm\theta}(\*z_l) = \log p(\bm\eta)
        + \log\left|\det\left(\frac{d\*g_{\bm\theta}}{d\*z}\right)\right|,
\end{equation}
and can be efficiently evaluated for some constrained $\*g_{\bm\theta}$. At test time, we compute the embedding map of an input image, and estimate the NLL of each of its embeddings. In our experiments, we use the  Real-NVP bijector~\cite{Dinh2016-sl}, composed of a succession of affine coupling layers, batch normalizations, and random permutations.

The benefits of this method are the following: (i) A normalizing flow can learn more complex distributions than the simple k\ac{nn} kernel or mixture of Gaussians used by~\cite{Lee2018-si}, where each embedding requires a class label, which is not available here; (ii) Features follow a simpler distribution than the input images, and can thus be correctly fit with simpler flows and shorter training times; (iii) The only hyperparameters are related to the architecture and the training of the flow, and can be cross-validated with the NLL of \ac{id} data without any \ac{ood} data; (iv) The training embeddings are efficiently summarized in the weights of the generative model with a very low memory footprint.

Input preprocessing~\cite{Liang2017-mj} can be trivially applied to our approach. Since the NLL estimator is an end-to-end network, we can compute the gradients of the average NLL w.r.t. the input image by backpropagating through the flow and the encoder.

A flow ensemble can be built by training separate density estimators over different layers of the segmentation model, similar to~\cite{Lee2018-si}. However, the resulting NLL estimates cannot be directly aggregated as is, because the different embedding distributions have varying dispersions and dimensions, and thus densities with very different scales. We propose to normalize the NLL $N(\*z_l)$ of a given embedding by the average NLL of the training features for that layer:
\begin{equation}
    \bar{N}(\*z_l) = N(\*z_l) - \mathcal{L}(\*Z_l).
\end{equation}
This is in fact a \ac{mc} approximation of the differential entropy of the flow, which is intractable. In the ideal case of a multivariate Gaussian, $\bar{N}$ corresponds to the Mahalanobis distance used by~\cite{Lee2018-si}. We can then aggregate the normalized, resized scores over different layers. We experiment with two strategies: (i) Using the minimum 
detects a pixel as OoD only if it has low likelihood through all layers, thus accounting for areas in the feature space that are in-distribution but contain only few training points; (ii) Following~\cite{Lee2018-si}, taking a weighted average %
, with weights given by a logistic regression fit on the FS Lost \& Found validation set, captures the interaction between the layers.

\subsection{Submitted Methods}
The following methods were submitted to our benchmark since it went online in August 2019. They were not implemented or trained by us, but we include an overview since they are part of the benchmark results.

\PAR{An outlier head} can be added in a multi-task fashion to many semantic segmentation architectures. \cite{Bevandic2019-qp} trains the head in a supervised fashion on both \ac{id} and \ac{ood} data samples. The training is executed simultaneously with the segmentation training. The outlier detection head then returns a pixel-wise anomaly score. Submitted were three variants of this method where the exact descriptions are in submission for publication.

\PAR{Image Resynthesis} uses reconstruction to estimate the fit of an input to the training data distribution of a generative model. While auto-encoders such as described in section~\ref{sec:related_work} scale poorly to the level of detail in urban driving, good results have been achieved with generative adversarial networks~\cite{Wang2017-uj,Isola2017-pk} that synthesize driving scenes from semantic segmentation. \cite{Lis2019-hq} uses such a method to find outliers by comparing the original and resynthesized image, where they train the comparison on flipped semantic labels in the \ac{id} data and therefore do not require outliers in training. While the original work~\cite{Lis2019-hq} experimented with lower resolution segmentation data,~\cite{Giancarlo2020} submitted an adapted, scaled-up model.

\PAR{Synboost} is a modular approach that combines introspective uncertainties and input reconstruction into a pixel-wise dissimilarity score. Further details are described in~\cite{Giancarlo2020}.

\begin{table*}[t]
    \centering
    \scriptsize{
    \setlength\tabcolsep{3pt}
    \begin{tabular}{ll|mc|mc|mc|mc|cccc}
    \toprule
    & &
    \multicolumn{2}{c|}{FS Lost \& Found} &
    \multicolumn{2}{c|}{FS Web Oct 20} &
    \multicolumn{2}{c|}{FS Web Jan 20} &
    \multicolumn{2}{c|}{FS Static} &
    \multirow{2}{*}{\parbox[b]{4em}{\centering re-training}} & \multirow{2}{*}{\parbox[b]{4.5em}{\centering \ac{ood}\\ data}} & 
    \multirow{2}{*}{\parbox[b]{4.5em}{\centering Cityscapes\\mIoU}} &
    \multirow{2}{*}{\parbox[b]{4.5em}{\centering time\\\lbrack s\rbrack}}\\
    
    method & score & AP$\,\uparrow$ & $\textrm{FPR}_\textrm{95}\,\downarrow$ &
    AP$\,\uparrow$ & $\textrm{FPR}_\textrm{95}\,\downarrow$ &
    AP$\,\uparrow$ & $\textrm{FPR}_\textrm{95}\,\downarrow$ &
    AP$\,\uparrow$ & $\textrm{FPR}_\textrm{95}\,\downarrow$ & 
     & & \\
    \toprule
    
    Random & random uncertainty & 00.3 & 95.0 & 01.6 & 95.0 & 01.5 & 95.0 & 02.5 & 95.0 &  \faClose & \faClose & 80.3 & -\\
    \midrule
    \multirow{2}{*}{Softmax}
        & max-probability & 01.8 & 44.8 & 11.8 & 40.1 & 11.1 & 37.8 & 12.9 & 39.8 & 
        \multirow{2}{*}{\faClose} & \multirow{2}{*}{\faClose} & \multirow{2}{*}{80.3} & 0.05\\
        & entropy  & 02.9 & 44.8 & 16.6 & 39.8 & 15.6 & 37.5 & 15.4 & 39.8 &&& & 0.29\\
    \midrule
    \multirow{2}{*}{\parbox[b]{8em}{kNN Embedding}}
        & density  & 03.5 & 30.0 & 24.9 & 35.2 & 22.3 & 31.7 & 44.0 & 20.2 & \multirow{2}{*}{\faClose} & \multirow{2}{*}{\faClose} & \multirow{2}{*}{80.3} & 7.89\\
        & relative class density  & 00.8 & 100.0 & 09.6 & 100.0 & 8.3 & 100.0  & 15.8 & 100.0 & & & & 6.45\\
    \midrule
    Image Resynthesis & resynthesis difference & 05.7 & 48.1 & 12.5 & 51.3 & & & 29.6 & 27.1 & \faClose & \faClose & 81.4 & 2.39\\
    \midrule
    \multirow{3}{*}{\parbox[b]{8em}{Learned\\ Embedding\\ Density (ours)}}
        & single-layer NLL  & 03.0 & \emph{32.9} & 21.4 & 44.0 & 18.8 & 41.1 & 40.9 & 21.3 & \multirow{3}{*}{\faClose} & \faClose & \multirow{3}{*}{80.3} & 0.29\\
        & minimum NLL & 04.3 & 47.2 & 32.6 & 55.8 & 30.2 & 49.5 & \emph{62.1} & 17.4 && \faClose & & 1.53\\
        & logistic regression & 04.7 & 24.4 & 29.2 & 38.8 & 26.0 & 33.8 & 57.2 & 13.4 && \faCheck & & 1.53\\
    \midrule
    SynBoost & dissimilarity score & \textbf{43.2} & \textbf{15.8} & 61.3 & 18.9 & & & 72.6 & 18.8 & \faClose & \faCheck & 81.4 & 2.32\\
    \midrule
    Bayesian DeepLab
        & mutual information & \emph{09.8} & 38.5 & \emph{35.8} & \emph{25.7} & \emph{33.8} & \emph{19.7} & 48.7 & \emph{15.5} &
        \faCheck & \faClose & 73.8 & 3.62\\
    \midrule
    \multirow{3}{*}{Outlier Head} 
        & fixed patches & 15.7 & 76.9 & 65.3 & 19.6 & \textbf{61.4} & 28.2 & 82.9 & 05.1 & \multirow{3}{*}{\faCheck} & \multirow{3}{*}{\faCheck} & 77.7 & 0.12\\
        & random patches & 21.2 & 36.9 & \textbf{67.2} & \textbf{10.3} & & & \textbf{86.2} & \textbf{02.4} & & & 77.3 & 0.12\\
        & combined probability & 30.9 & 22.2 & 64.0 & 18.8 & & & 84.0 & 10.3 & & & 77.3 & 0.12\\
    \midrule
    \multirow{2}{*}{\ac{ood} training}
            & max-entropy & 01.7 & 30.6 & 28.2 & 22.6 & 28.7 & 18.4 & 27.5 & 23.6 &
            \multirow{2}{*}{\faCheck} & \multirow{2}{*}{\faCheck} & 79.0 & 0.29\\
            & void classifier & 10.3 & 22.1 & 43.0 & 14.0 & 42.8 & \textbf{12.0} & 45.0 & 19.4 &&& 70.4 & 0.05\\
    \midrule
    Dirichlet DeepLab
        & prior entropy & 34.3 & 47.4 & 30.0 & 76.6 & 33.9 & 75.5 & 31.3 & 84.6 & \faCheck & \faCheck & 70.5 & 0.05\\
    \bottomrule
    \end{tabular}
    }
    \vspace{5pt}
    \caption{\textbf{Benchmark Results}. The gray columns mark the primary metric of the benchmark. Methods are only evaluated on those FS Web datasets with object images appearing on the web after their submission date. For every metric and dataset, the best performance is marked bold and the best performance without \ac{ood} training is marked italic.
    }
    \label{tab:results}
    \vspace{-3mm}
\end{table*}

\begin{figure}
    \centering
    \input{fsweb_over_time}
    \vspace{-3mm}
    \caption{Performance evolution over the different iterations of the FS Web dataset. We only plot the best-performing variant of each method. Methods that train on \ac{ood} data are plotted with dashed lines. Notable changes are the better blending method in June 19 and the inclusion of blended \ac{id} objects in January 20, which changed the data-balance.}
    \label{fig:fsweb}
    \vspace{-3mm}
\end{figure}

\input{good_and_bad_examples}

\section{Discussion of Results}
We show in Table~\ref{tab:results} the results of our benchmark as of December 2020 for the aforementioned datasets and methods. Qualitative examples of all methods are shown in figure~\ref{fig:good-and-bad}.

\PAR{Softmax Confidence.} Confirming findings on simpler tasks~\cite{Lee2018-si}, the softmax confidence is not a reliable score for anomaly detection. While training with \ac{ood} data clearly improves the softmax-based detection, it is not much better than Bayesian DeepLab, that does not require such data.
 
\PAR{Difference between datasets.} For most methods, there is a clear performance gap between the data from Lost \& Found and the other datasets. We attribute this to two factors. First, the dataset contains a lot of images with only very small objects. This is indicated by the AP of the random classifier, which equals to the fraction of anomalous pixels. Second, the qualitative examples show the more challenging nature of the Lost \& Found dataset with e.g. false positives for the void classifier or outlier head, and cases where small anomalous objects are not detected at all e.g. for the Bayesian DeepLab or Softmax Entropy.\\
We further investigate the results on FS Web over time in figure~\ref{fig:fsweb}. While most methods follow overall trends that can be attributed to the difficulty of the individual objects or differences in data balance, it becomes clear that (i) embedding based methods were picking up blending artifacts in FS Web March 2019, and (ii) Dirichlet DeepLab is performing very inconsistently. (i) appears to be fixed with the advanced blending from June 2019, since the introduction of blended \ac{id} objects did not have any effect on embedding based methods. (ii) could indicate a degree of overfitting to specific object types, because Dirichlet DeepLab is trained on \ac{ood} data.

\PAR{Semantic Segmentation Accuracy.} The data in table~\ref{tab:results}  illustrates a tradeoff between anomaly detection and segmentation performance. Methods like Bayesian DeepLab or Outlier Head are consistently among the best methods on all datasets, but need to train with special losses that reduce the segmentation accuracy by up to 10\%. If segmentation accuracy is important, methods that do not require any retraining are particularly interesting.

\PAR{Supervision with \ac{ood} data} appears to be important for good anomaly detection. On every dataset, the best method required \ac{ood} data and is at least 38\% better than any `unsupervised' method. While training with \ac{ood} data can in principle lead to overfitting to specific objects, the results on FS Web, which was designed specifically to resemble open-world settings, show that the Outlier Head or Dissimilarity Ensemble are very robust to diverse anomalies.\\
We however want to emphasize that anomaly detection and uncertainty estimation are very different principles. Our benchmark therefore serves the dual purpose of finding either the best anomaly segmentation method or well-scalable uncertainty estimates, that are simply tested on the proxy task of anomaly detection. Comparing Bayesian DeepLab and the void classifier shows that good uncertainty estimation methods can even compete with some supervised methods, but so far not with specifically designed anomaly segmentation methods.

\PAR{Inference time} differs significantly between methods. Methods can be broadly sorted into two categories, where the first do a single pass through a (sometimes modified) DeepLabv3+ architecture and the second category applies additional processing on top of this forward pass. Our measurements show that methods in the second category have up to two orders of magnitude higher inference time. The only exception marks the single-layer embedding density, where inference time is comparable to single pass methods. While nearly all methods\footnote{Image Resynthesis and SynBoost were submitted as pytorch models.} were executed as optimised tensorflow graphs, measurements are still dependent on the implementation details and possible parallelization is limited by GPU memory constraints. For example, the difference between softmax max-prob, softmax entropy, and dirichlet entropy can only be explained with inefficiencies in the softmax entropy implementation that cause a difference of more than 0.2 s.

\PAR{Challenges in Method Adaptation.} The results reveal that some methods cannot be easily adapted to semantic segmentation. For example, retraining required by special losses can impair the segmentation performance, and we found that these losses (e.g. for Dirichlet DeepLab) were often unstable during training or did not converge. Other challenges rise from the complex network structures which complicate the translation of class-based embedding methods such as deep k-nearest neighbor~\cite{Papernot2018-xz} to segmentation. This is illustrated by the performance of our simple implementation.

\section{Conclusion}

In this work, we introduced \textit{Fishyscapes}, a benchmark for anomaly detection in semantic segmentation for urban driving.
Comparing state-of-the-art methods on this complex task for the first time, we draw multiple conclusions:
\begin{itemize}[label={--},leftmargin=*]
    \item The softmax output from a standard classifier is a bad indicator for anomaly detection.
    \item Most of the better performing methods required special losses that reduce the semantic segmentation accuracy.
    \item Supervision of anomaly segmentation methods with \ac{ood} data consistently outperformed unsupervised methods even in open-world scenarios.
\end{itemize}

Overall, the methods compared in our benchmark so far leave a lot of room for improvement. To safely deploy semantic segmentation methods in autonomous cars, further research is required. As a public benchmark, \textit{Fishyscapes} supports the evaluation of new methods on urban driving scenarios.

\bibliographystyle{spmpsci}
\bibliography{PaperpileMar07.bib}

\appendix
\section*{Appendix}
\input{appendix.tex}
\end{document}

%% file: qualitative_examples.tex
\begin{figure*}[htb!]
\centering
\def\iwidth{.19\linewidth}
\def\iheight{1.66}
\def\iiheight{\iheight*2.4}
\def\iiiheight{\iheight*3.4}
\def\iiiiheight{\iheight*4.4}
\def\iiiibheight{\iheight*5.4}
\def\vheight{\iheight*6.8}
\def\viheight{\iheight*7.8}
\def\viiheight{\iheight*8.8}

\begin{tikzpicture}
\node[rotate=90] at (-2,-0.8) {\scriptsize\strut Fishyscapes Static};
\node[rotate=90] at (-2,-6.5) {\scriptsize\strut Fishyscapes Web October 2020};
\node[rotate=90] at (-2,-\viheight) {\scriptsize\strut Fishyscapes Lost \& Found};

\node[inner sep=0pt, label=\scriptsize{Input\strut}] at (0,0)
    {\includegraphics[width=\iwidth]{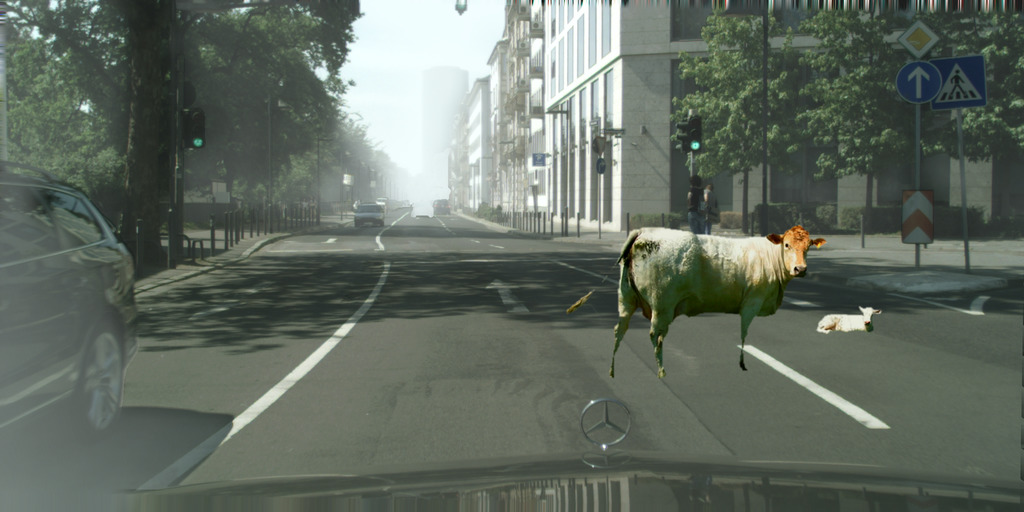}};
\node[inner sep=0pt, label=\scriptsize{DeepLabv3+ Prediction\strut}] () at (\iwidth,0)
    {\includegraphics[width=\iwidth]{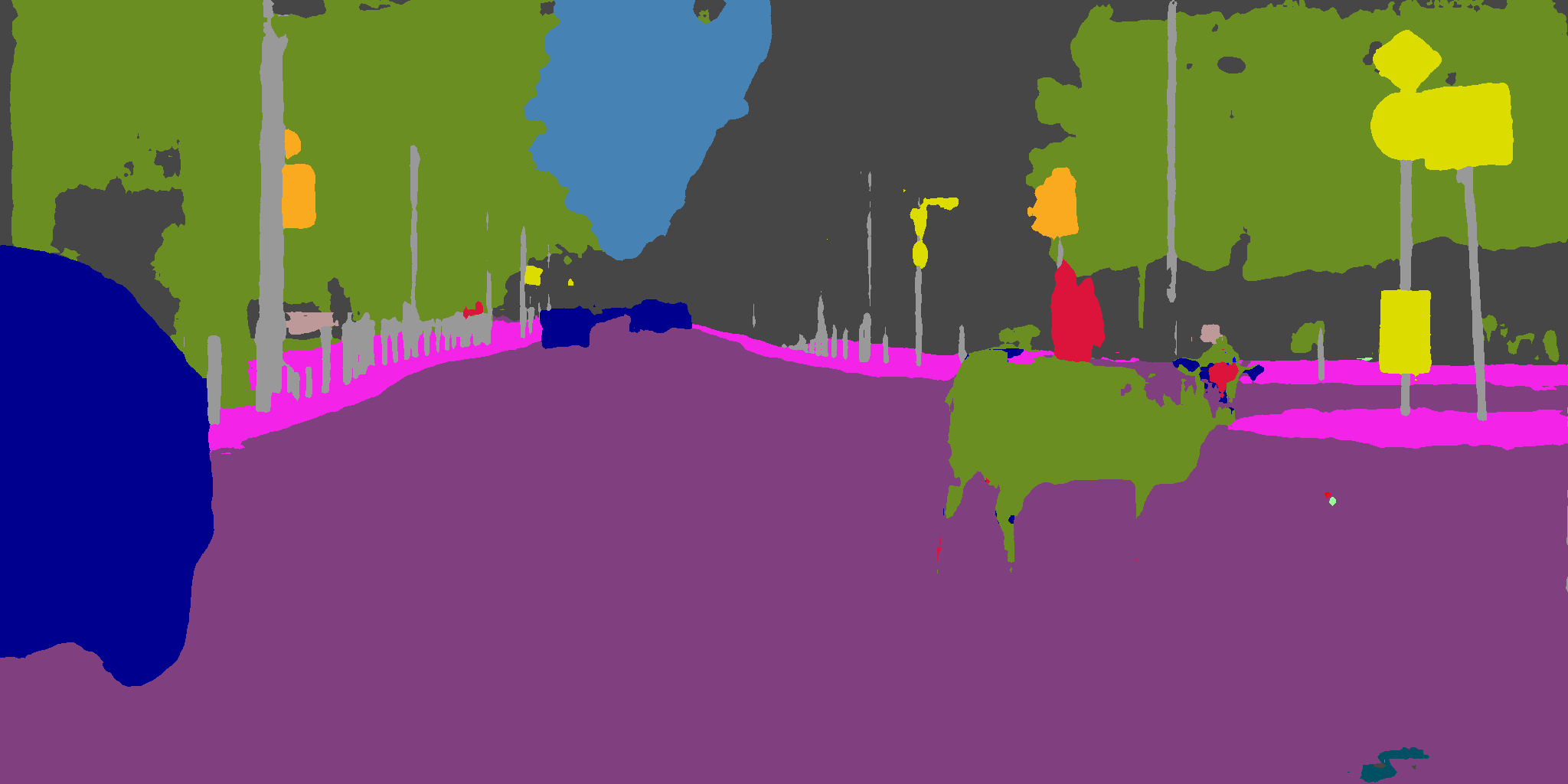}};
\node[inner sep=0pt, label=\scriptsize{Ground Truth\strut}] (a1) at (\iwidth*2,0) {\includegraphics[width=\iwidth]{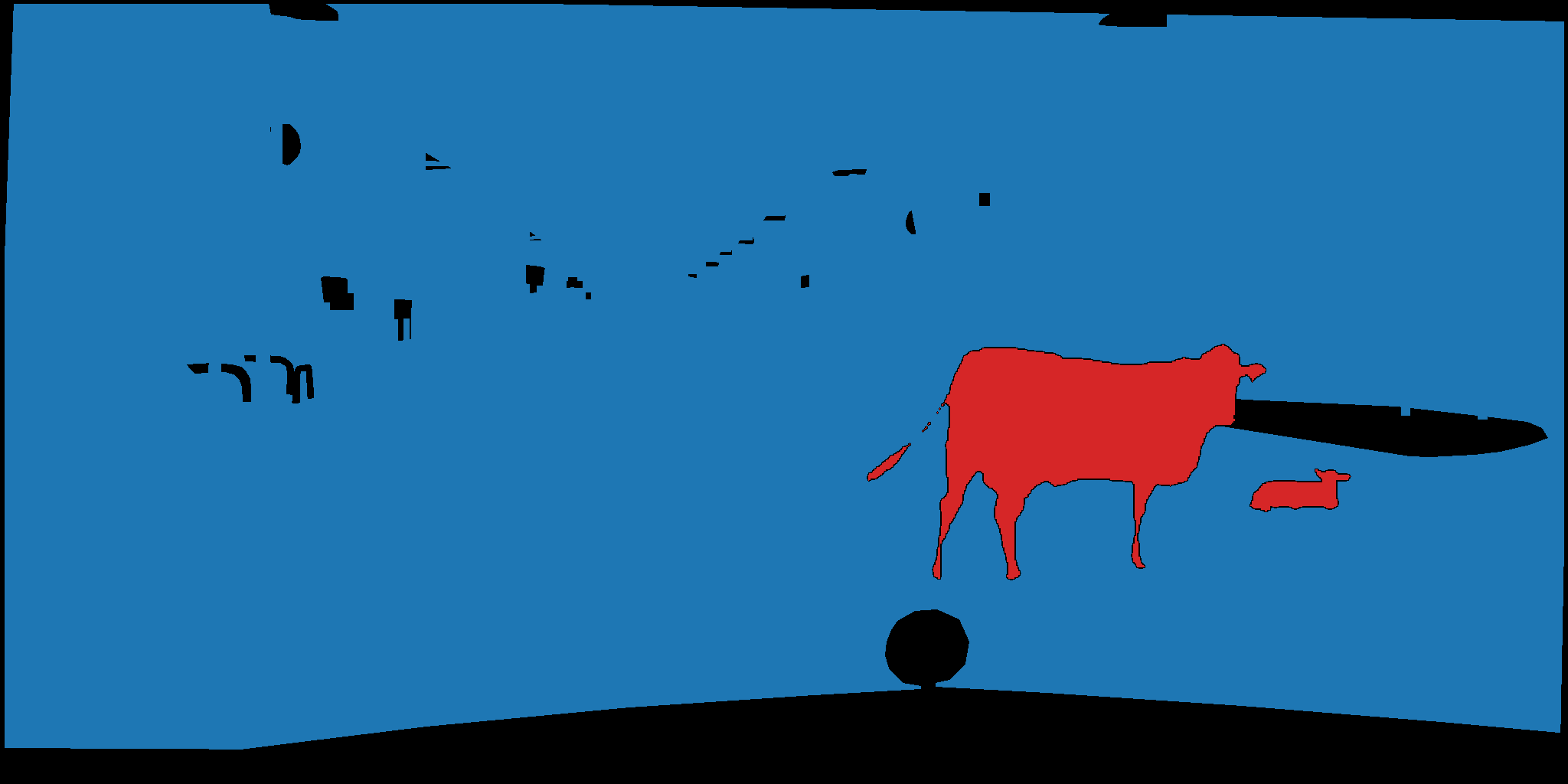}};
\node[inner sep=0pt, label=\scriptsize{Outlier Head (random patches)\strut}] () at (\iwidth*3,0)
    {\includegraphics[width=\iwidth]{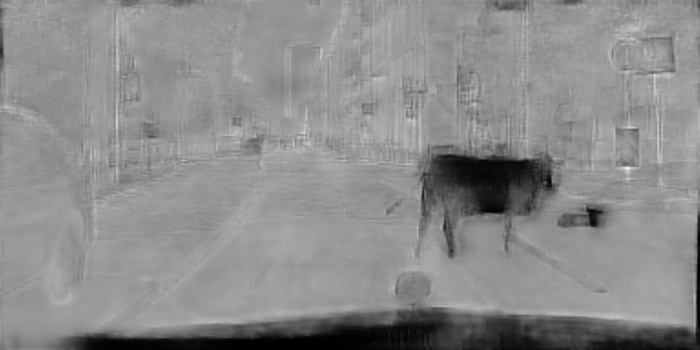}};
\node[inner sep=1pt,below left] at (\iwidth*3.5,\iheight/2)
    {\tiny{\textcolor{blue}{93.4}}};
\node[inner sep=0pt, label=\scriptsize\strut Embedding Density (min nll)] () at (\iwidth*4,0)
    {\includegraphics[width=\iwidth]{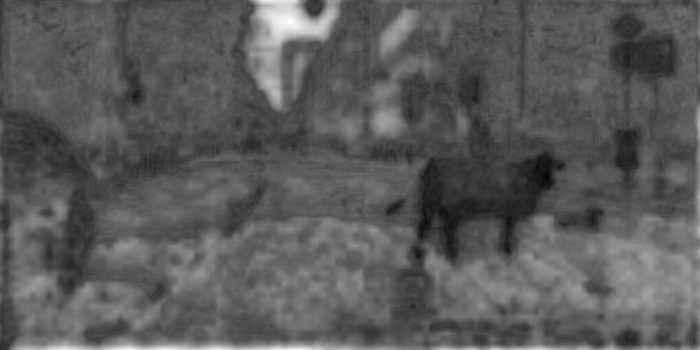}};
\node[inner sep=1pt,below left] at (\iwidth*4.5,\iheight/2)
    {\tiny{\textcolor{yellow}{69.1}}};

\node[inner sep=0pt] at (0,-\iheight)
    {\includegraphics[width=\iwidth]{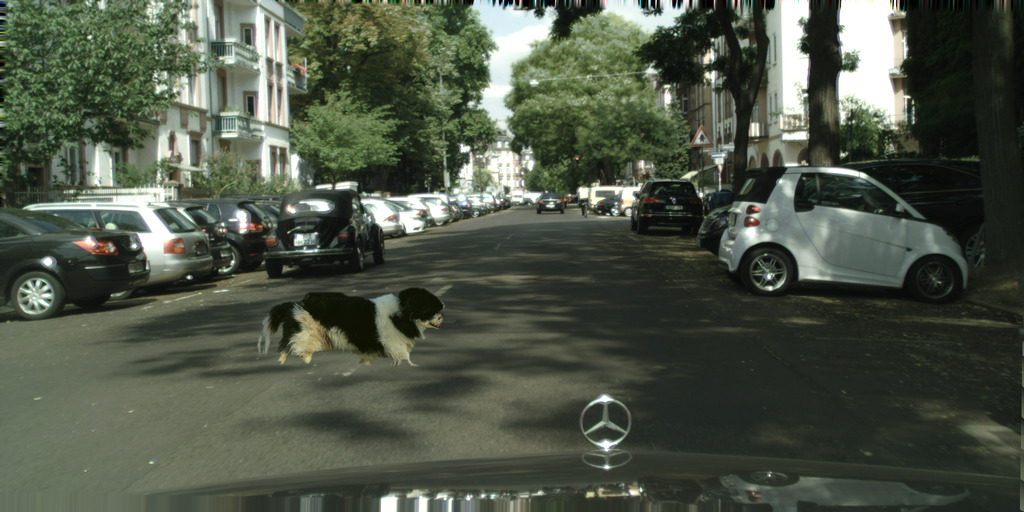}};
\node[inner sep=0pt] at (\iwidth,-\iheight)
    {\includegraphics[width=\iwidth]{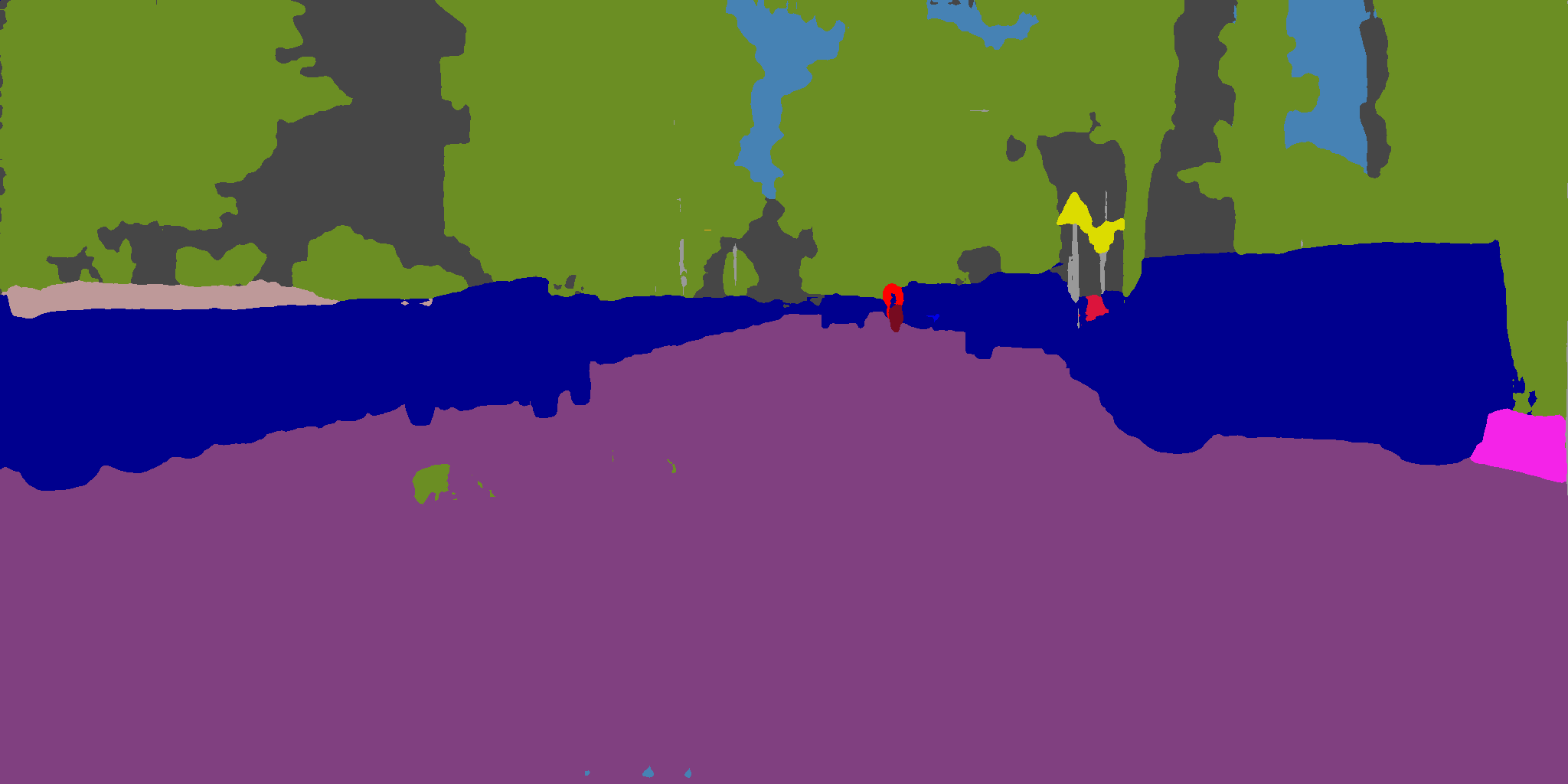}};
\node[inner sep=0pt] at (\iwidth*2,-\iheight)
    {\includegraphics[width=\iwidth]{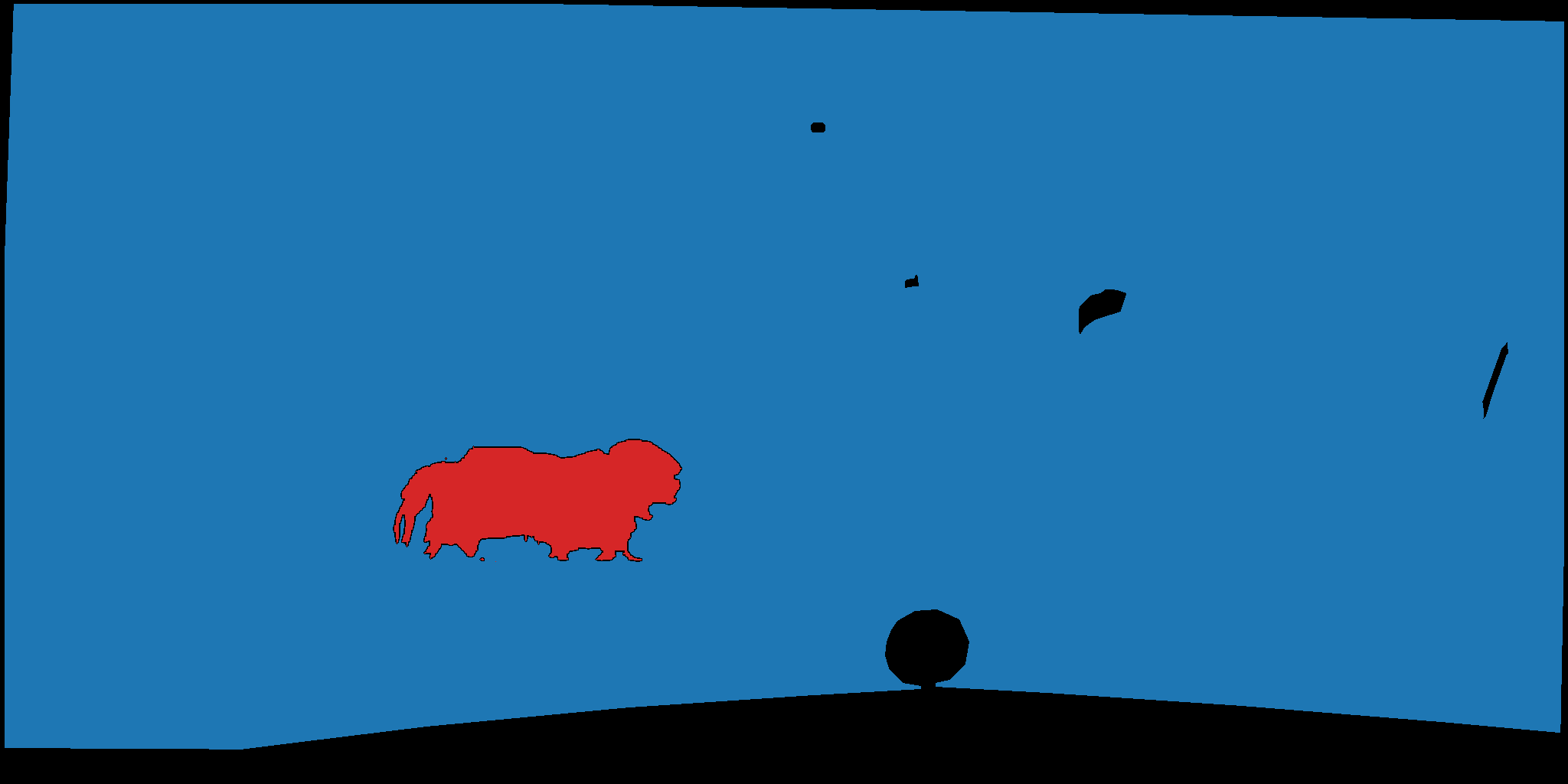}};
\node[inner sep=0pt] at (\iwidth*3,-\iheight)
    {\includegraphics[width=\iwidth]{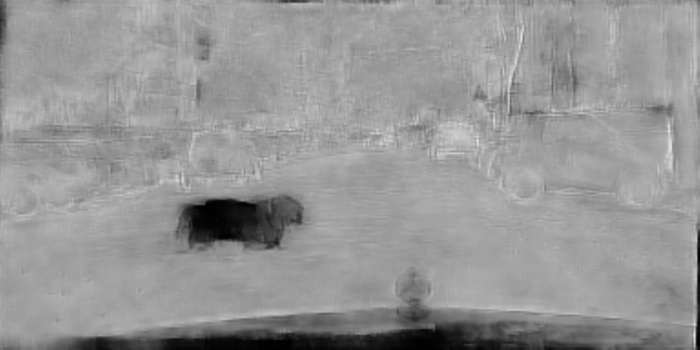}};
\node[inner sep=1pt,below left] at (\iwidth*3.5,-\iheight/2)    
    {\tiny{\textcolor{blue}{97.1}}};
\node[inner sep=0pt] at (\iwidth*4,-\iheight)
    {\includegraphics[width=\iwidth]{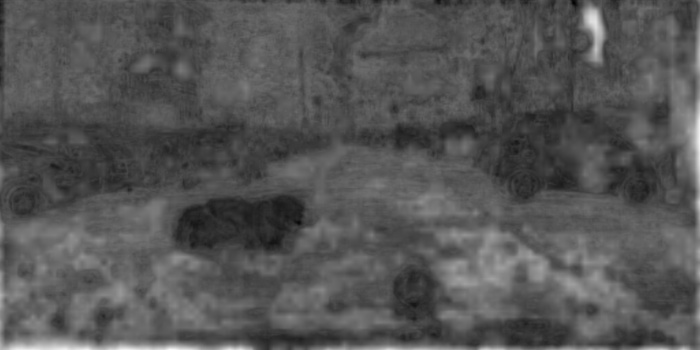}};
\node[inner sep=1pt,below left] at (\iwidth*4.5,-\iheight/2)
    {\tiny{\textcolor{yellow}{65.9}}};
    
\node[inner sep=0pt, label=\scriptsize{Input\strut}] at (0,-\iiheight)
    {\includegraphics[width=\iwidth]{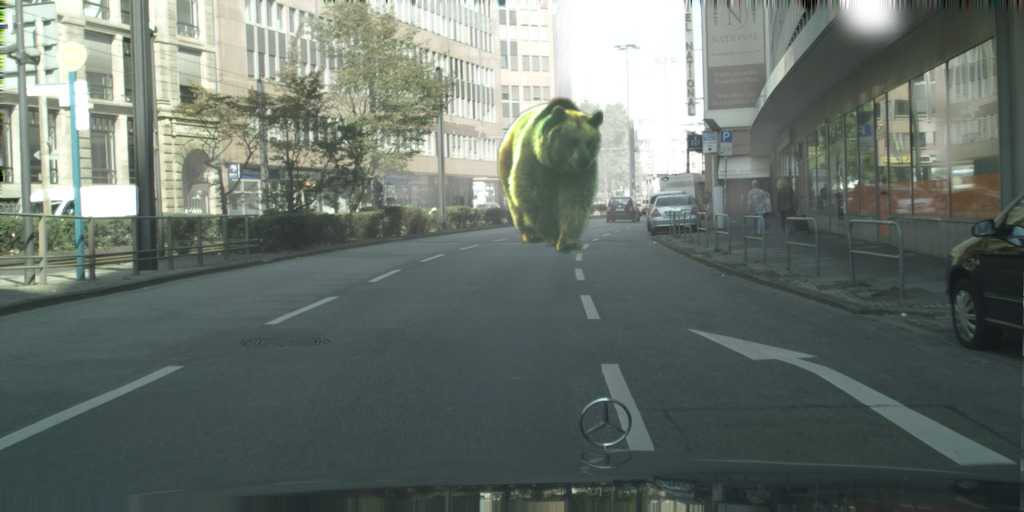}};
\node[inner sep=0pt, label=\scriptsize{DeepLabv3+ Prediction\strut}] at (\iwidth,-\iiheight)
    {\includegraphics[width=\iwidth]{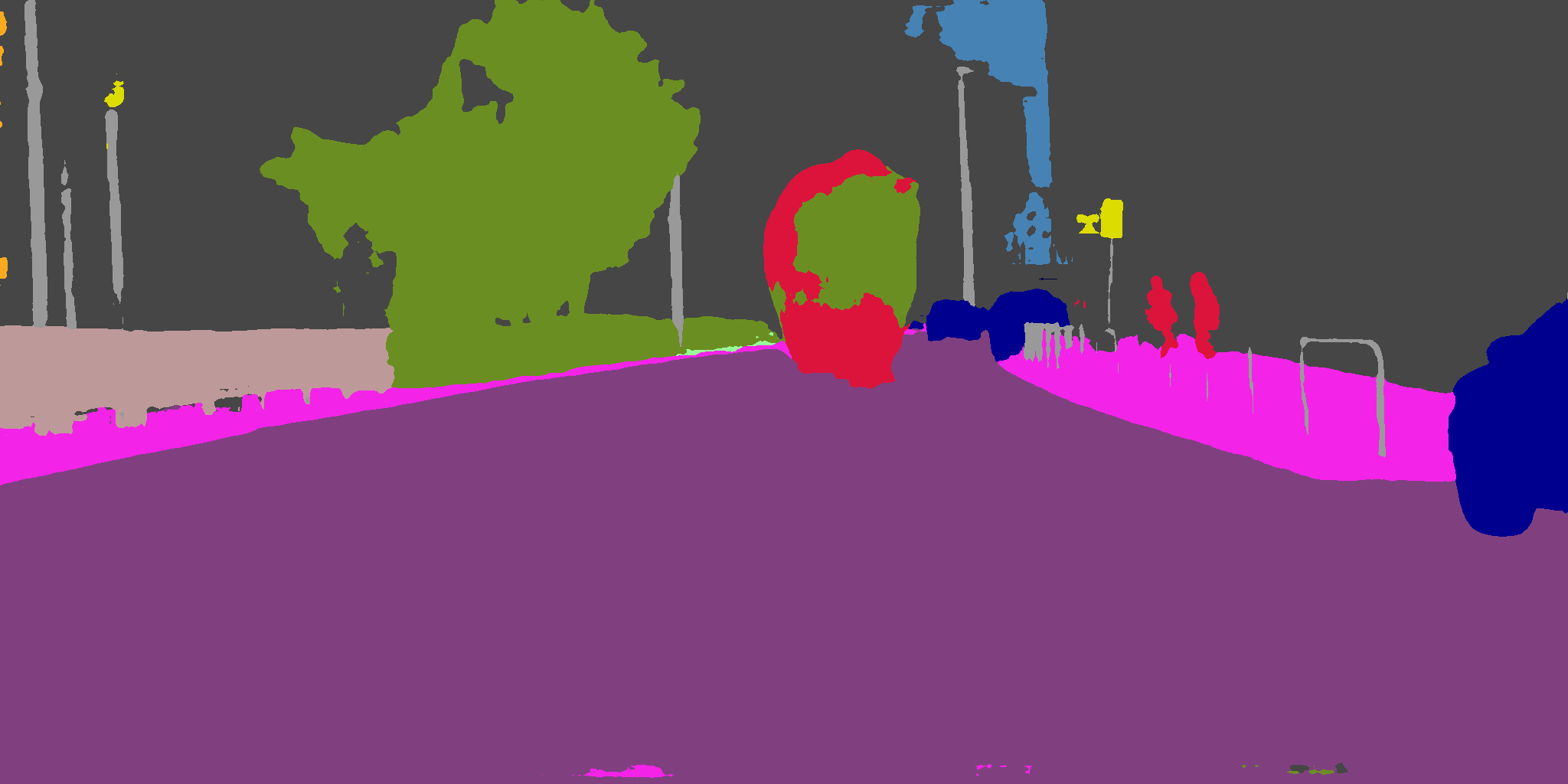}};
\node[inner sep=0pt, label=\scriptsize{Ground Truth\strut}] at (\iwidth*2,-\iiheight)
    {\includegraphics[width=\iwidth]{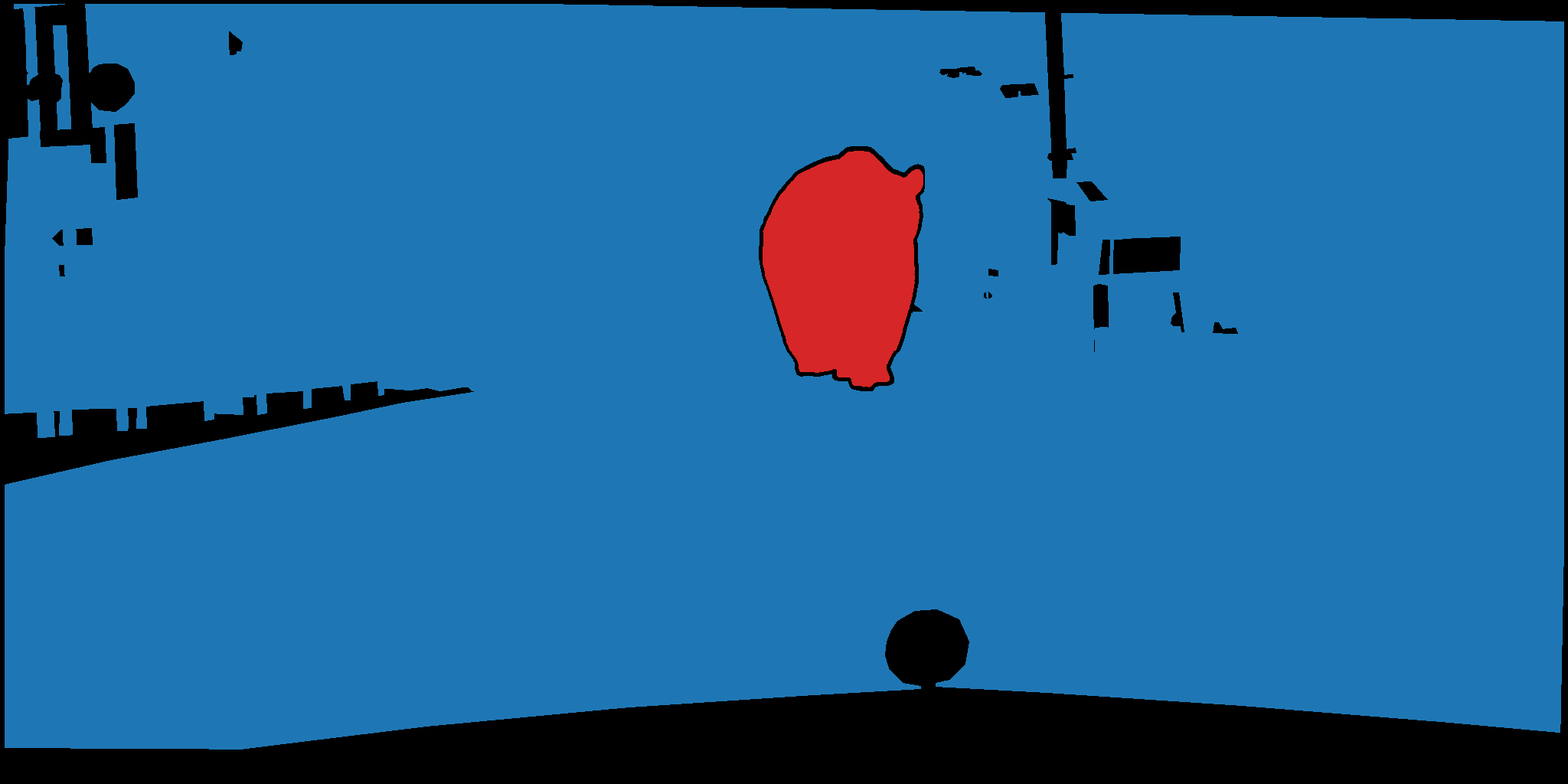}};
\node[inner sep=0pt, label=\scriptsize{Outlier Head (random patches)\strut}] at (\iwidth*3,-\iiheight)
    {\includegraphics[width=\iwidth]{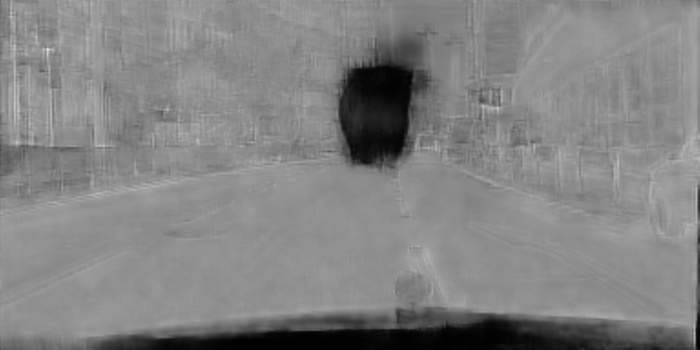}};
\node[inner sep=1pt,below left] at (\iwidth*3.5,.5*\iheight-\iiheight)
    {\tiny{\textcolor{blue}{93.8}}};
\node[inner sep=0pt, label=\scriptsize{MC Dropout (MI)\strut}] at (\iwidth*4,-\iiheight)
    {\includegraphics[width=\iwidth]{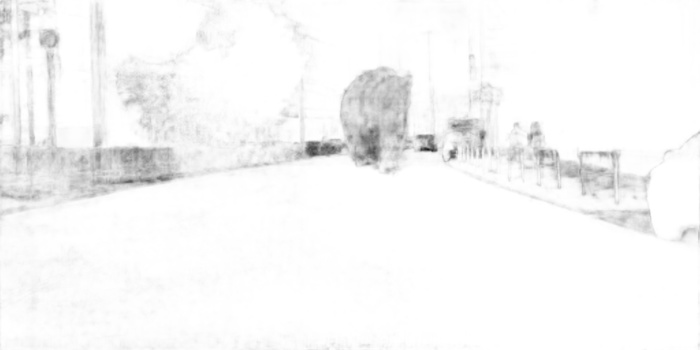}};
\node[inner sep=1pt,below left] at (\iwidth*4.5,.5*\iheight-\iiheight)
    {\tiny{\textcolor{blue}{62.6}}};
    
\node[inner sep=0pt] at (0,-\iiiheight)
    {\includegraphics[width=\iwidth]{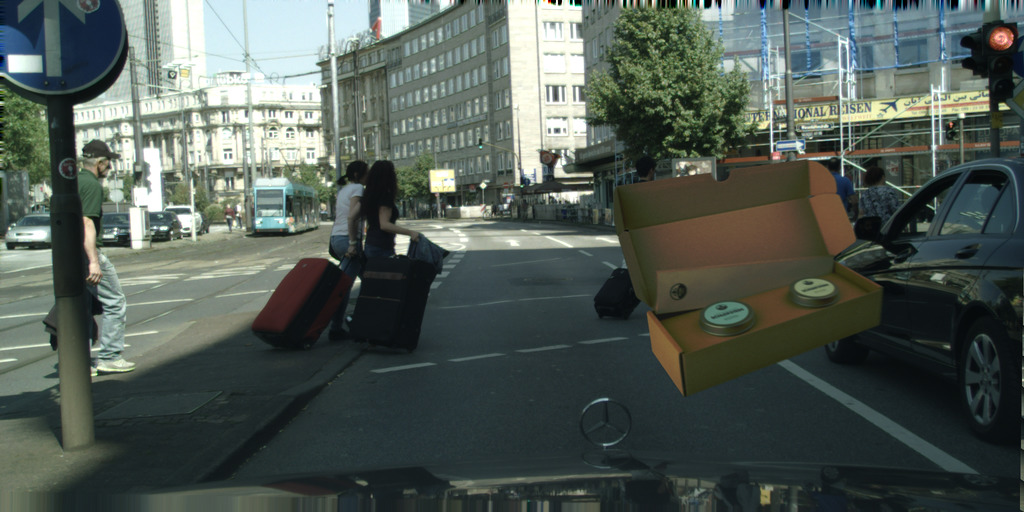}};
\node[inner sep=0pt] at (\iwidth,-\iiiheight)
    {\includegraphics[width=\iwidth]{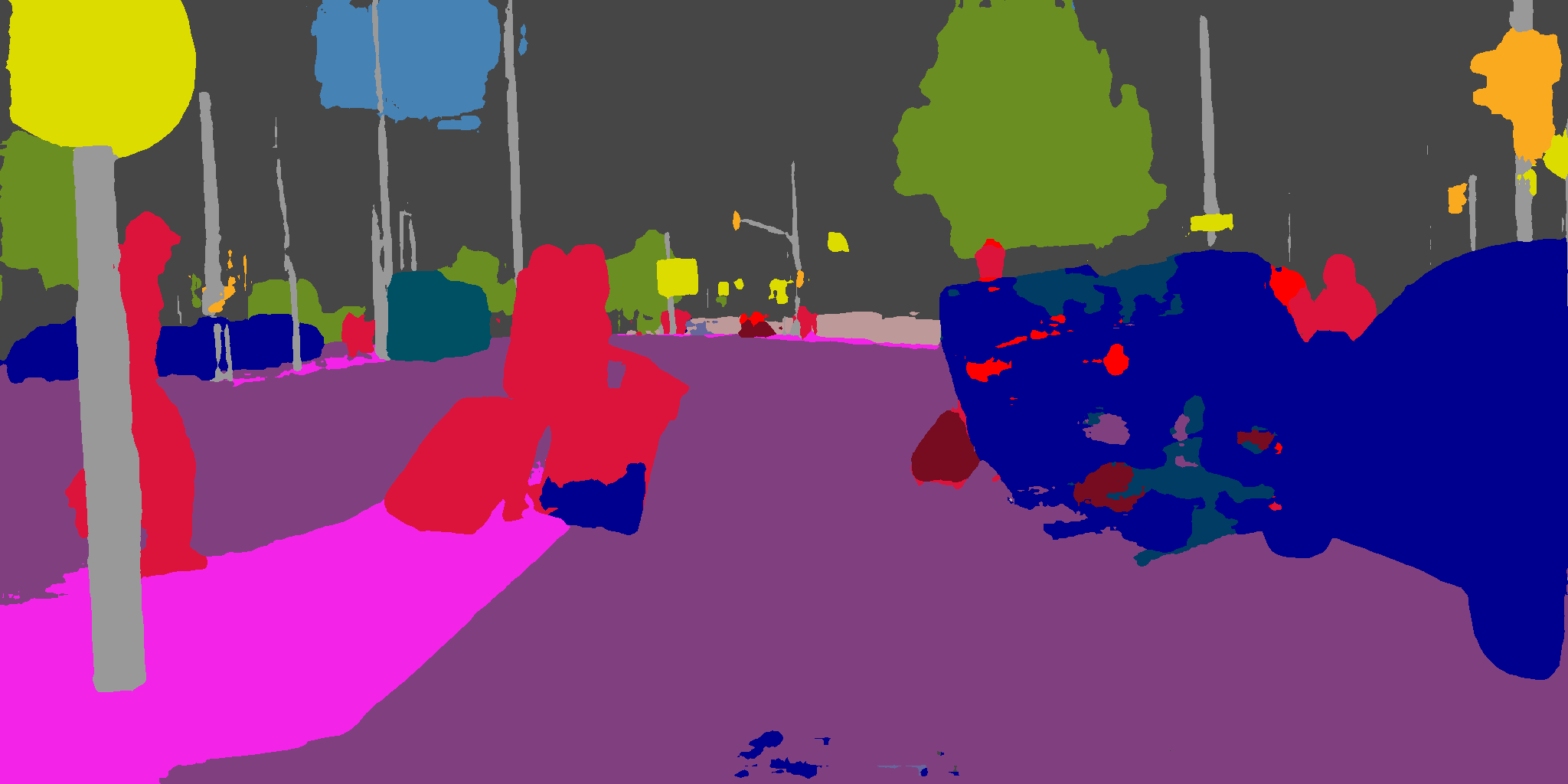}};
\node[inner sep=0pt] at (\iwidth*2,-\iiiheight)
    {\includegraphics[width=\iwidth]{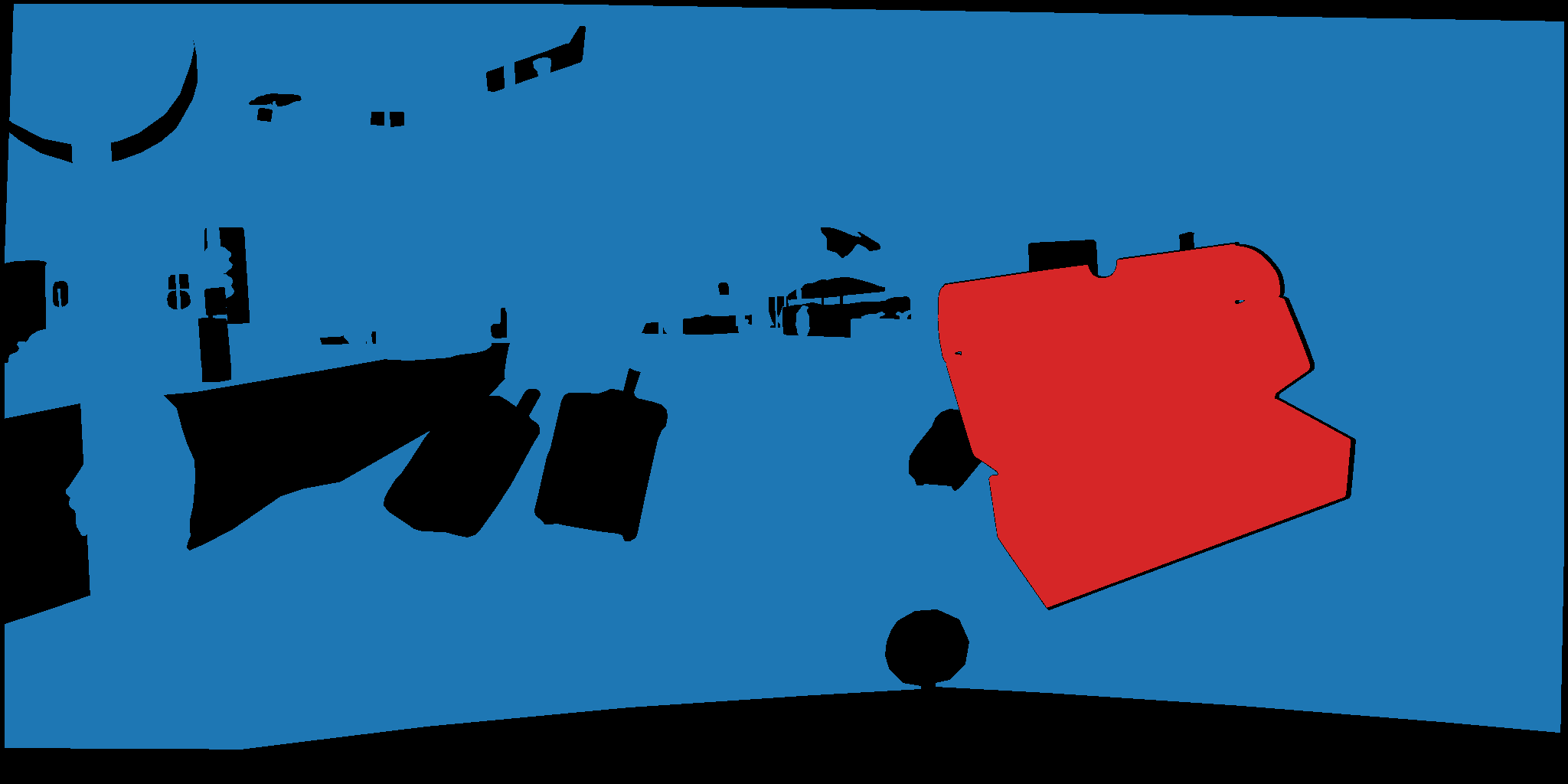}};
\node[inner sep=0pt] at (\iwidth*3,-\iiiheight)
    {\includegraphics[width=\iwidth]{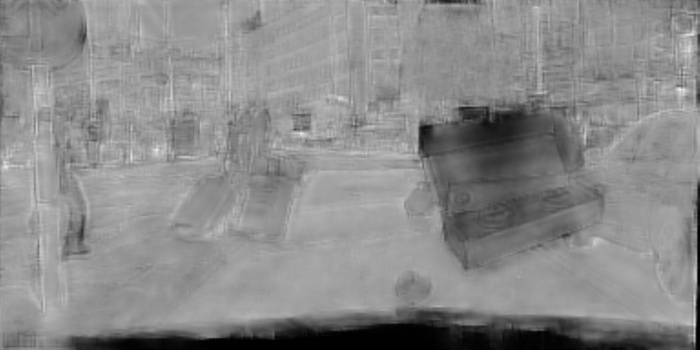}};
\node[inner sep=1pt,below left] at (\iwidth*3.5,.5*\iheight-\iiiheight)
    {\tiny{\textcolor{blue}{88.5}}};
\node[inner sep=0pt] at (\iwidth*4,-\iiiheight)
    {\includegraphics[width=\iwidth]{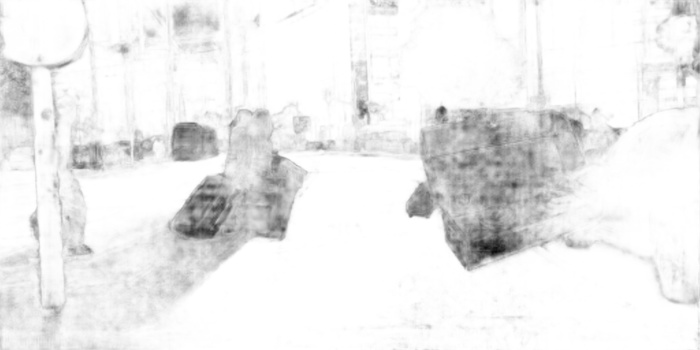}};
\node[inner sep=1pt,below left] at (\iwidth*4.5,.5*\iheight-\iiiheight)
    {\tiny{\textcolor{blue}{70.6}}};
    
\node[inner sep=0pt] at (0,-\iiiiheight)
    {\includegraphics[width=\iwidth]{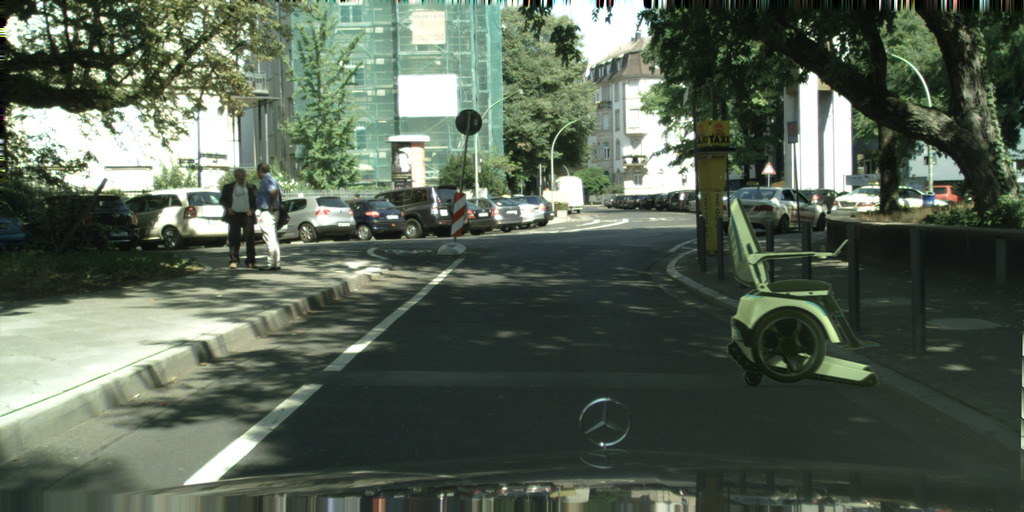}};
\node[inner sep=0pt] at (\iwidth,-\iiiiheight)
    {\includegraphics[width=\iwidth]{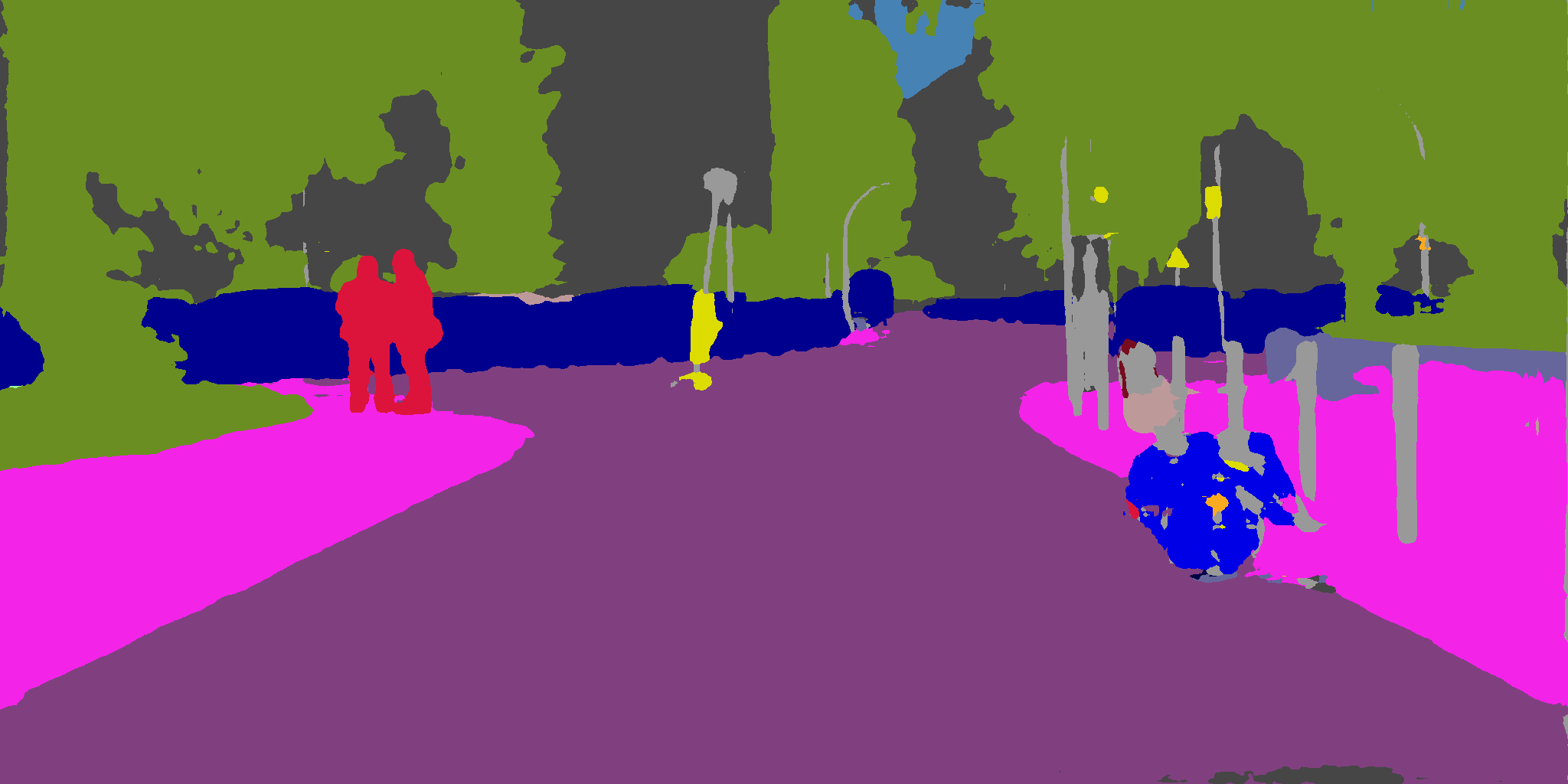}};
\node[inner sep=0pt] at (\iwidth*2,-\iiiiheight)
    {\includegraphics[width=\iwidth]{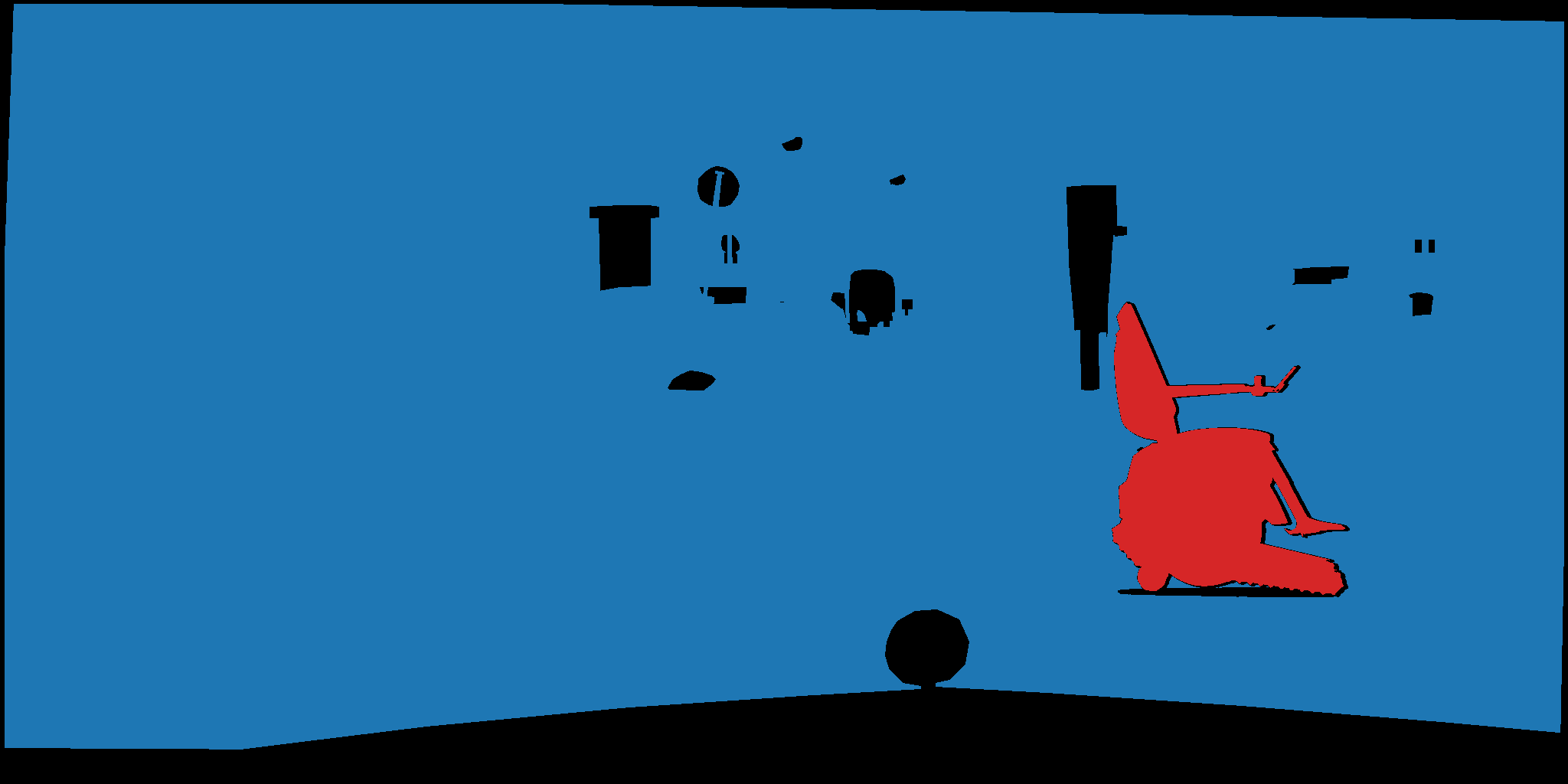}};
\node[inner sep=0pt] at (\iwidth*3,-\iiiiheight)
    {\includegraphics[width=\iwidth]{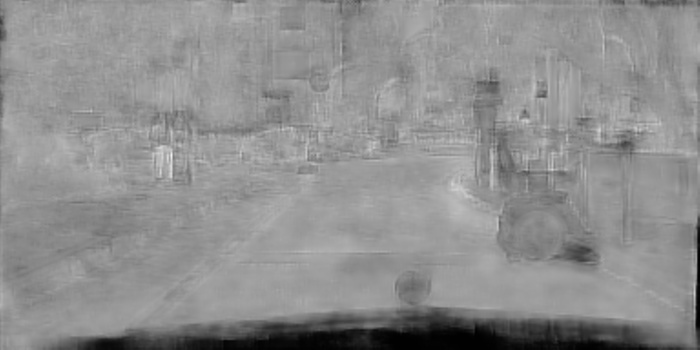}};
\node[inner sep=1pt,below left] at (\iwidth*3.5,.5*\iheight-\iiiiheight)
    {\tiny{\textcolor{blue}{37.3}}};
\node[inner sep=0pt] at (\iwidth*4,-\iiiiheight)
    {\includegraphics[width=\iwidth]{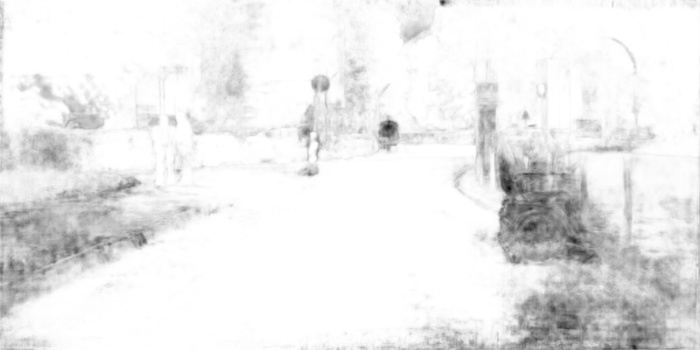}};
\node[inner sep=1pt,below left] at (\iwidth*4.5,.5*\iheight-\iiiiheight)
    {\tiny{\textcolor{blue}{51.8}}};
    
\node[inner sep=0pt] at (0,-\iiiibheight)
    {\includegraphics[width=\iwidth]{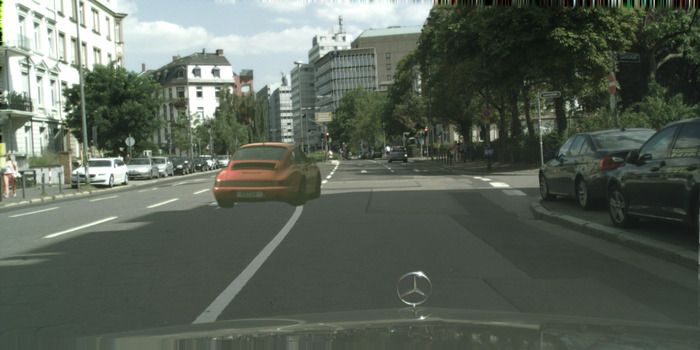}};
\node[inner sep=0pt] at (\iwidth,-\iiiibheight)
    {\includegraphics[width=\iwidth]{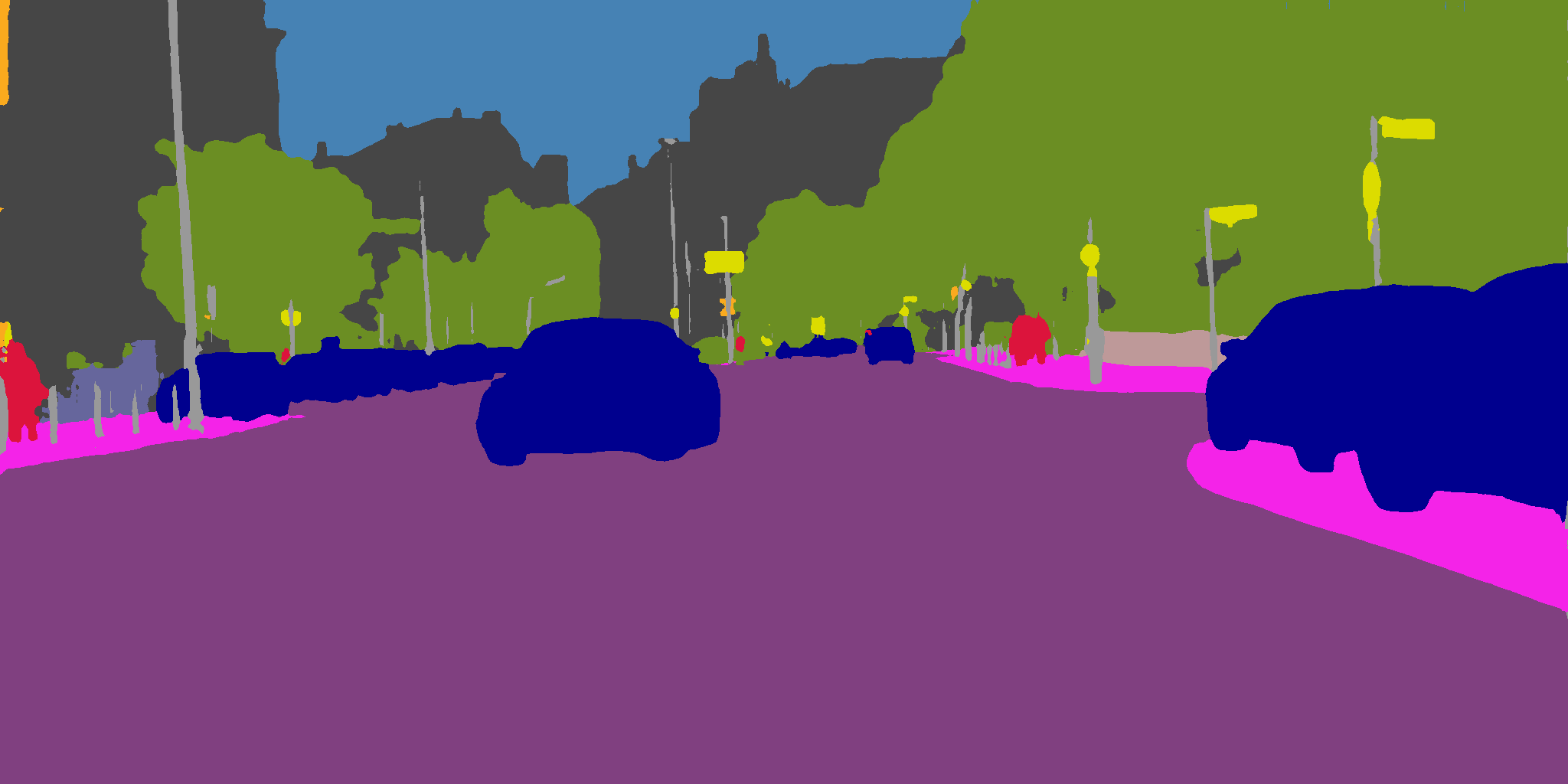}};
\node[inner sep=0pt] at (\iwidth*2,-\iiiibheight)
    {\includegraphics[width=\iwidth]{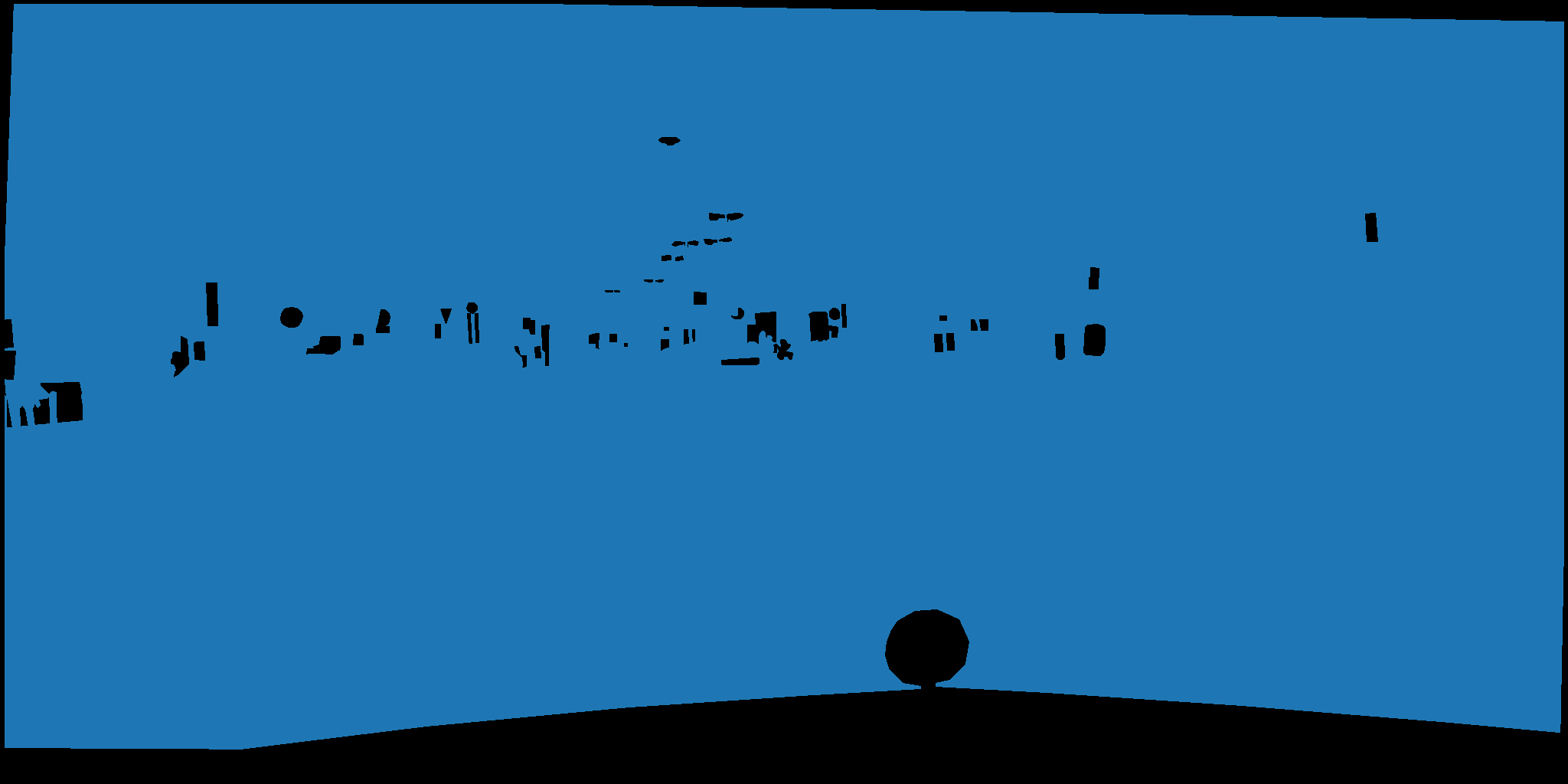}};
\node[inner sep=0pt] at (\iwidth*2,-\iiiibheight-.4)
    {\footnotesize{\textcolor{white}{blended inlier object}}};
\node[inner sep=0pt] at (\iwidth*3,-\iiiibheight)
    {\includegraphics[width=\iwidth]{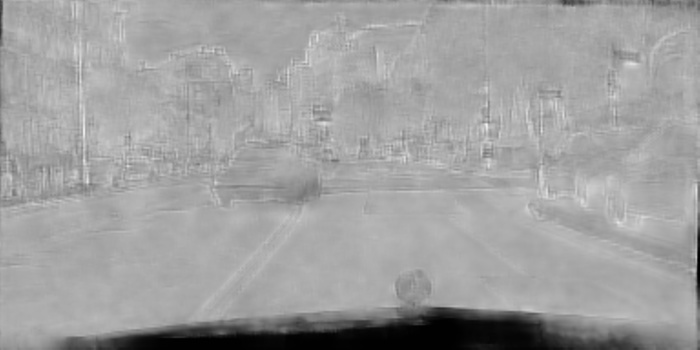}};
\node[inner sep=1pt,below left] at (\iwidth*3.5,.5*\iheight-\iiiibheight)
    {\tiny{\textcolor{blue}{100}}};
\node[inner sep=0pt] at (\iwidth*4,-\iiiibheight)
    {\includegraphics[width=\iwidth]{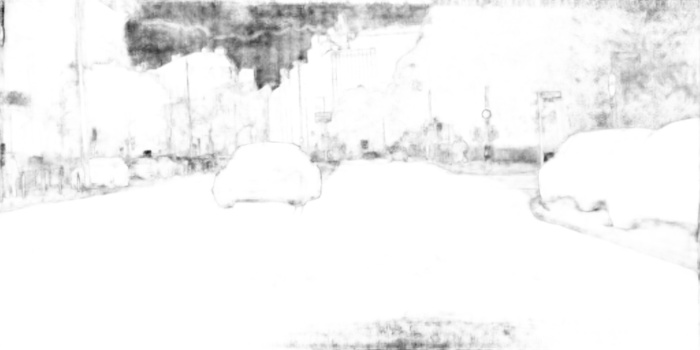}};
\node[inner sep=1pt,below left] at (\iwidth*4.5,.5*\iheight-\iiiibheight)
    {\tiny{\textcolor{blue}{100}}};
    
\node[inner sep=0pt, label=\scriptsize{Input\strut}] at (0,-\vheight)
    {\includegraphics[width=\iwidth]{laf_0008_rgb.jpg}};
\node[inner sep=0pt, label=\scriptsize{DeepLabv3+ Prediction\strut}] () at (\iwidth,-\vheight)
    {\includegraphics[width=\iwidth]{laf_0008_pred.png}};
\node[inner sep=0pt, label=\scriptsize{Ground Truth\strut}] () at (\iwidth*2,-\vheight)
    {\includegraphics[width=\iwidth]{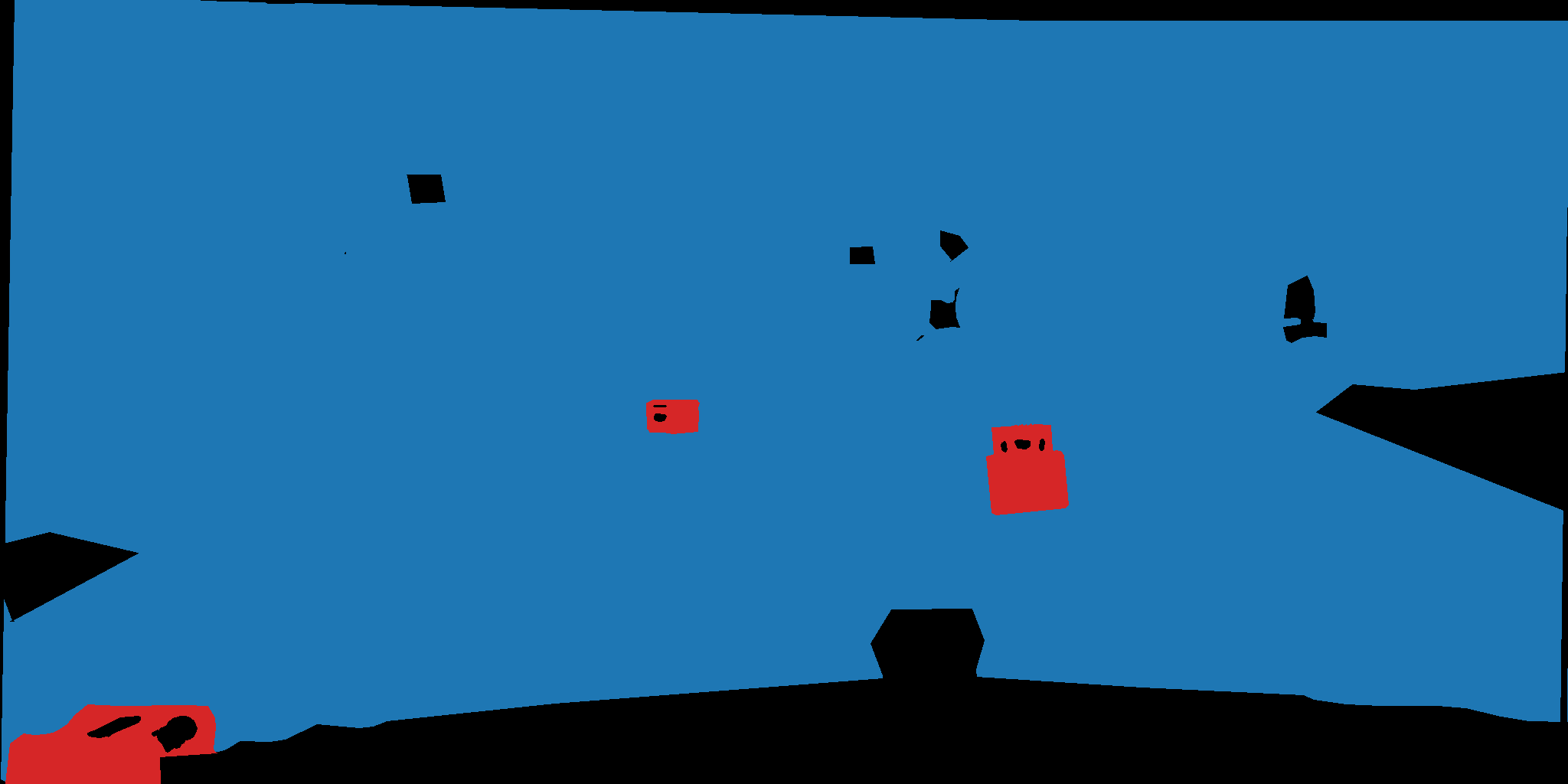}};
\node[inner sep=0pt, label=\scriptsize\strut SynBoost] () at (\iwidth*3,-\vheight)
    {\includegraphics[width=\iwidth]{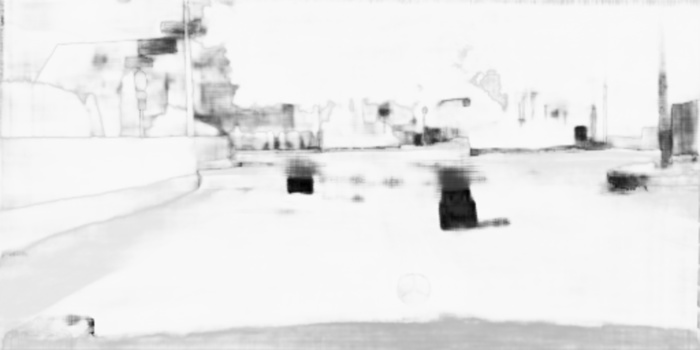}};
\node[inner sep=1pt,below left] at (\iwidth*3.5,.5*\iheight-\vheight)
    {\tiny{\textcolor{blue}{45.7}}};
\node[inner sep=0pt, label=\scriptsize\strut MC Dropout (MI)] () at (\iwidth*4,-\vheight)
    {\includegraphics[width=\iwidth]{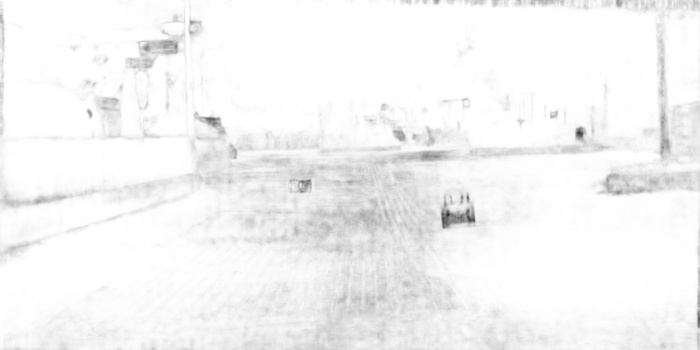}};
\node[inner sep=1pt,below left] at (\iwidth*4.5,.5*\iheight-\vheight)
    {\tiny{\textcolor{blue}{01.2}}};
    
\node[inner sep=0pt] at (0,-\viheight)
    {\includegraphics[width=\iwidth]{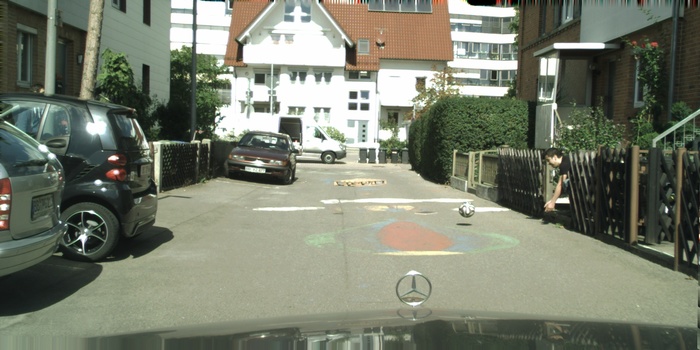}};
\node[inner sep=0pt] at (\iwidth,-\viheight)
    {\includegraphics[width=\iwidth]{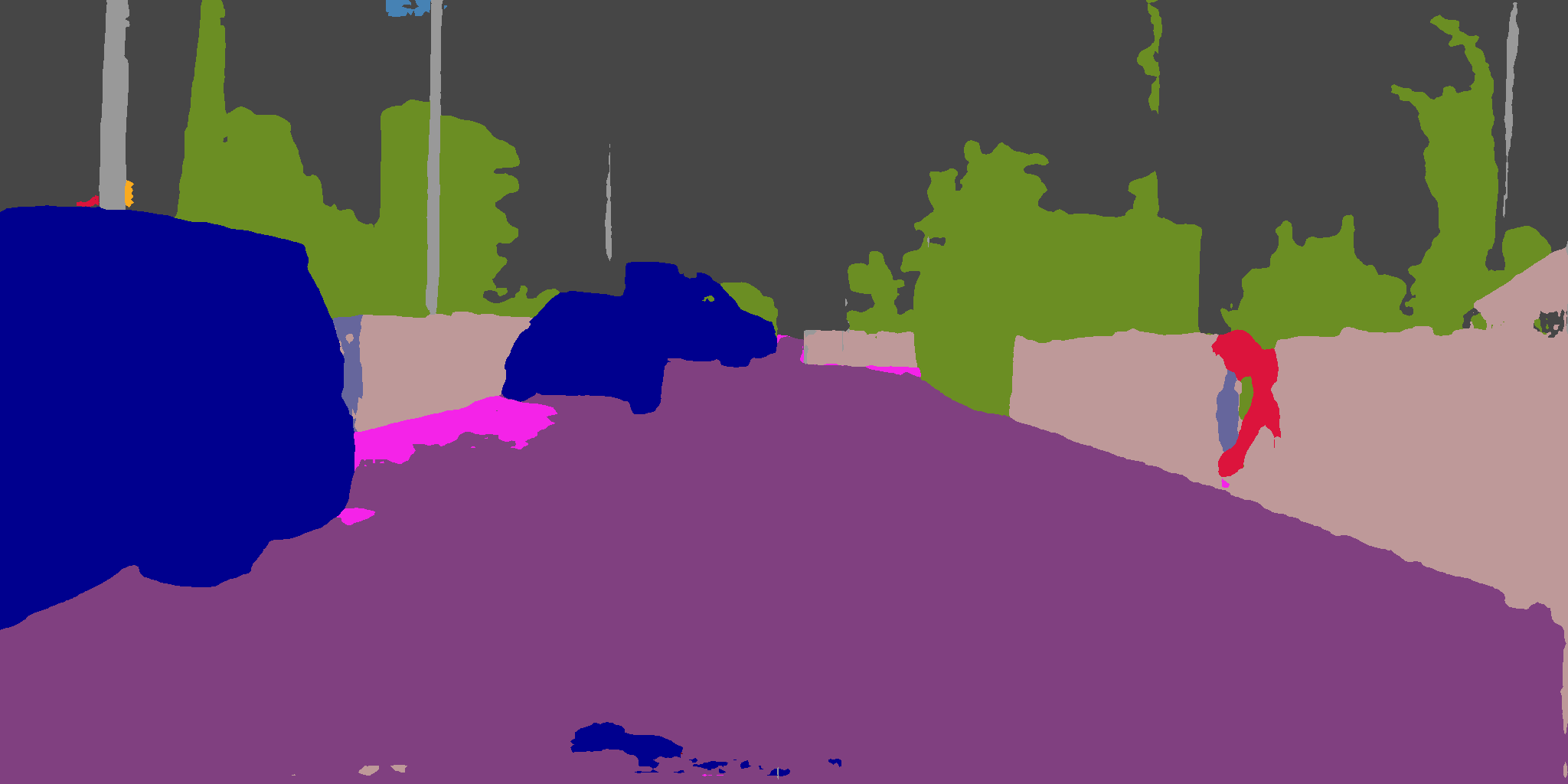}};
\node[inner sep=0pt] at (\iwidth*2,-\viheight)
    {\includegraphics[width=\iwidth]{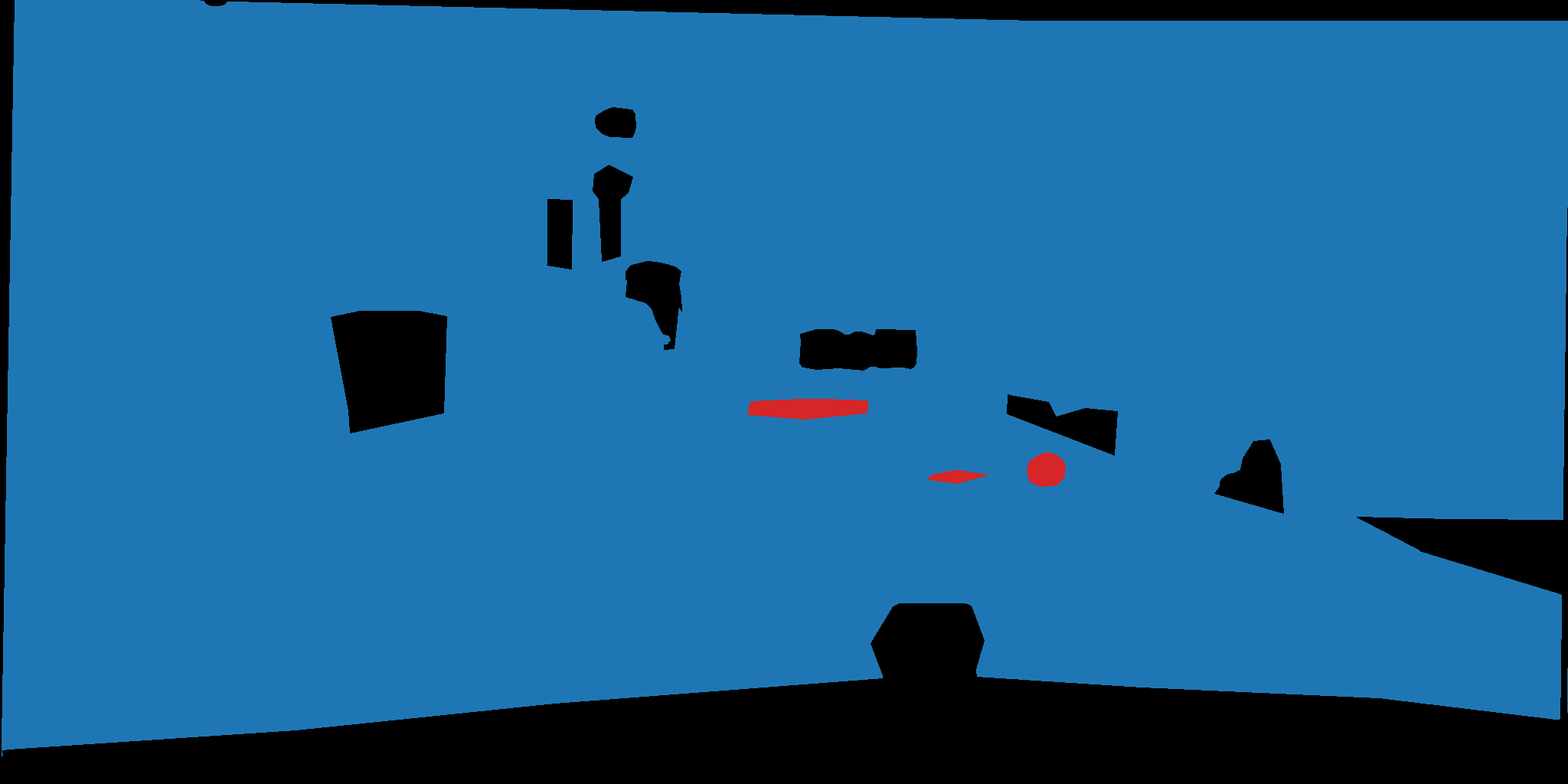}};
\node[inner sep=0pt] at (\iwidth*3,-\viheight)
    {\includegraphics[width=\iwidth]{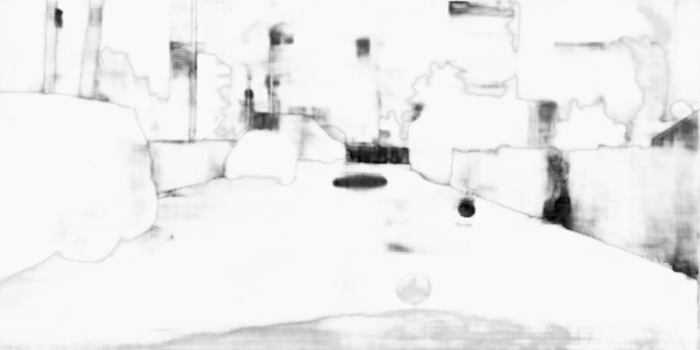}};
\node[inner sep=1pt,below left] at (\iwidth*3.5,.5*\iheight-\viheight)
    {\tiny{\textcolor{blue}{43.9}}};
\node[inner sep=0pt] at (\iwidth*4,-\viheight)
    {\includegraphics[width=\iwidth]{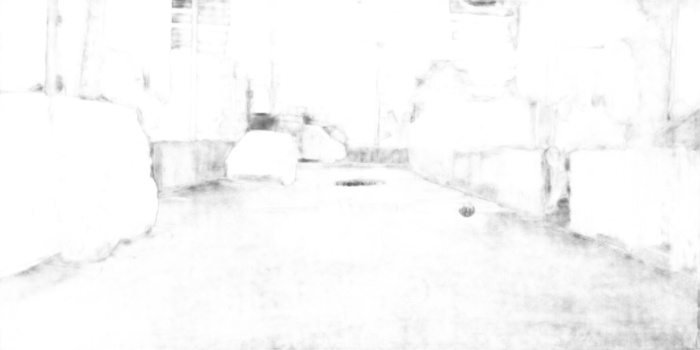}};
\node[inner sep=1pt,below left] at (\iwidth*4.5,.5*\iheight-\viheight)
    {\tiny{\textcolor{blue}{01.2}}};
    
\node[inner sep=0pt] at (0,-\viiheight)
    {\includegraphics[width=\iwidth]{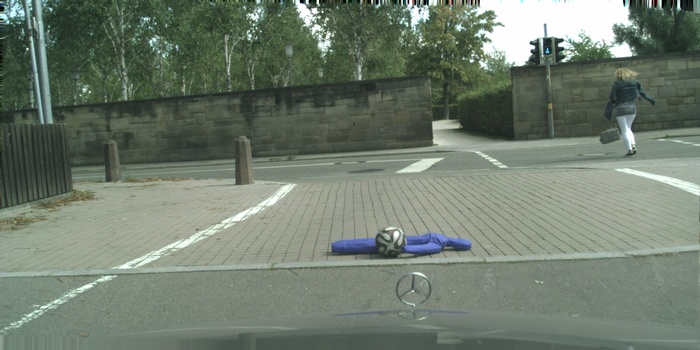}};
\node[inner sep=0pt] at (\iwidth,-\viiheight)
    {\includegraphics[width=\iwidth]{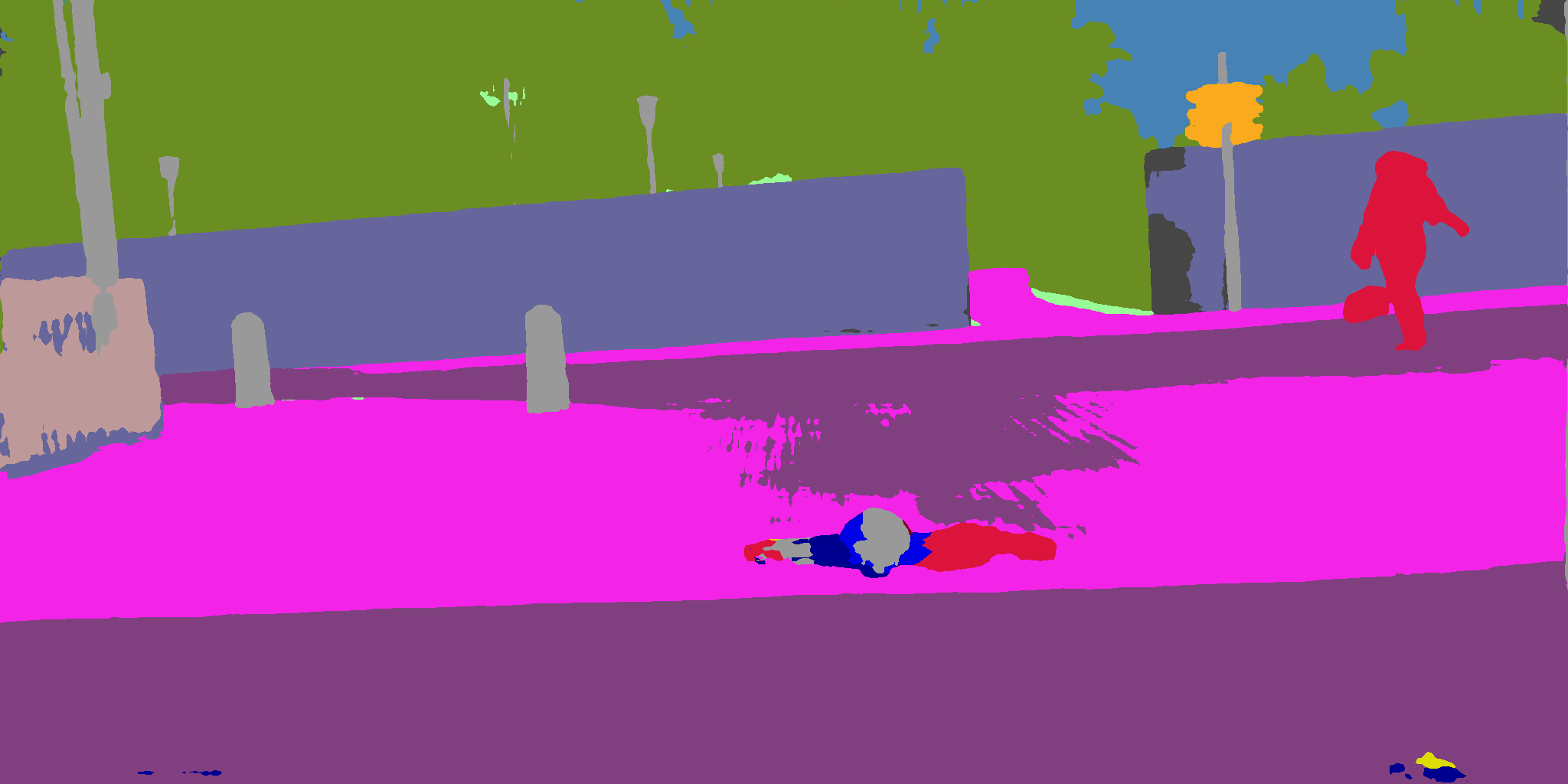}};
\node[inner sep=0pt] at (\iwidth*2,-\viiheight)
    {\includegraphics[width=\iwidth]{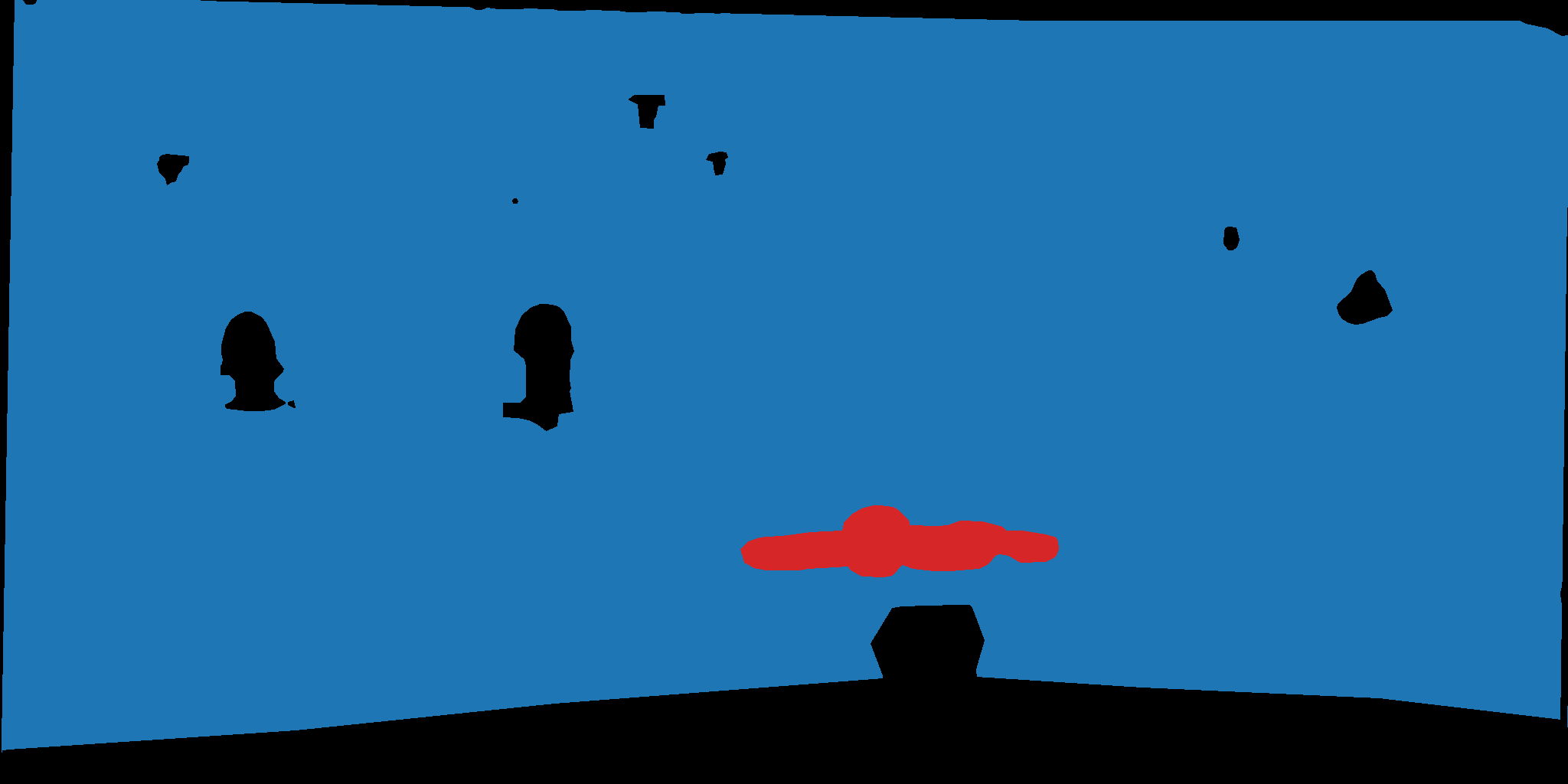}};
\node[inner sep=0pt] at (\iwidth*3,-\viiheight)
    {\includegraphics[width=\iwidth]{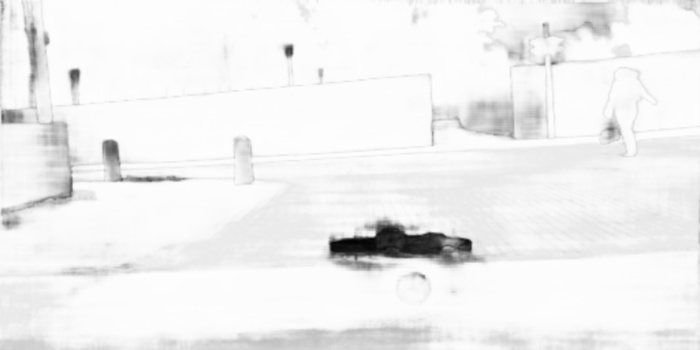}};
\node[inner sep=1pt,below left] at (\iwidth*3.5,.5*\iheight-\viiheight)
    {\tiny{\textcolor{blue}{96.6}}};
\node[inner sep=0pt] at (\iwidth*4,-\viiheight)
    {\includegraphics[width=\iwidth]{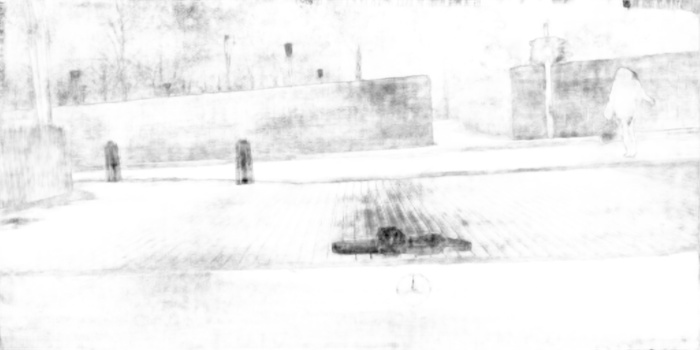}};
\node[inner sep=1pt,below left] at (\iwidth*4.5,.5*\iheight-\viiheight)
    {\tiny{\textcolor{blue}{02.4}}};
\end{tikzpicture}
\caption{\textbf{Qualitative examples} of Fishyscapes Static (rows 1-2) and Fishyscapes Web (rows 3-5) and Fishyscapes Lost \& Found (rows 6-8). The ground truth contains labels for \ac{id} (blue) and \ac{ood} (red) pixels, as well as ignored void pixels (black). We additionally show the output of the best method per dataset in column 4 and the best method without \ac{ood} training in the last column. We report the AP of each method output in its top right corner.}
\label{fig:data-overview}
\vspace{-4mm}
\end{figure*}

%% file: fsweb_over_time.tex
\begin{tikzpicture}

\definecolor{color0}{rgb}{0.12156862745098,0.466666666666667,0.705882352941177}
\definecolor{color1}{rgb}{1,0.498039215686275,0.0549019607843137}
\definecolor{color2}{rgb}{0.172549019607843,0.627450980392157,0.172549019607843}
\definecolor{color3}{rgb}{0.83921568627451,0.152941176470588,0.156862745098039}
\definecolor{color4}{rgb}{0.580392156862745,0.403921568627451,0.741176470588235}
\definecolor{color5}{rgb}{0.549019607843137,0.337254901960784,0.294117647058824}
\definecolor{color6}{rgb}{0.890196078431372,0.466666666666667,0.76078431372549}
\definecolor{color7}{rgb}{0.737254901960784,0.741176470588235,0.133333333333333}
\definecolor{color8}{rgb}{0.0901960784313725,0.745098039215686,0.811764705882353}

\begin{axis}[
axis x line=bottom,
axis y line=left,
axis line shift=5pt,
axis line style={-},
ticklabel style={scale=0.8},
tick align=outside,
tick pos=left,
x grid style={white!69.0196078431373!black},
xmin=-0.2, xmax=5.8,
xtick = {0,1,2,3,4},
xticklabels = {{March 19},{June 19},{Sept. 19}, {Jan. 20}, {Oct. 20}},
xtick style={color=black},
y grid style={white!69.0196078431373!black},
ylabel={\small average precision},
ymin=0, ymax=.8,
ytick style={color=black}
]
\addplot [semithick, black, dotted]
table {%
0 0.0264882976046282
1 0.0278138006786062
2 0.0232780714193781
3 0.0147103005334189
4 0.0163906898466801
};
\addplot [semithick, color0, mark=*, mark size=2, mark options={solid}]
table {%
0 0.235564270383052
1 0.237640272297105
2 0.260868126581718
3 0.155824268670576
3 0.155824268491529
4 0.166143072070021
4 0.166143088534487
};
\addplot [semithick, color2, mark=*, mark size=2, mark options={solid}]
table {%
0 0.503794559771594
1 0.364836796611718
2 0.355237590491565
3 0.223432880292585
4 0.248573367706262
};
\addplot [semithick, color3, mark=*, mark size=2, mark options={solid}]
table {%
0 0.788925381160755
1 0.418920836433195
2 0.401401353275534
3 0.302029789397414
4 0.32632842626898
};
\addplot [dashed, semithick, color4, mark=*, mark size=2, mark options={solid}]
table {%
0 0.529
1 0.568
2 0.572850547908232
3 0.42787984978344
4 0.430206034889669
};
\addplot [semithick, color5, mark=*, mark size=2, mark options={solid}]
table {%
0 0.521
1 0.547
2 0.532357000065614
3 0.338433002553015
4 0.358048934975696
};
\addplot [dashed, semithick, color6, mark=*, mark size=2, mark options={solid}]
table {%
0 0.277
1 0.435803366627373
2 0.434359750823132
3 0.339430460274597
4 0.300220255627381
};
\addplot [dashed, semithick, white!49.8039215686275!black, mark=*, mark size=2, mark options={solid}]
table {%
3 0.613828833083897
4 0.652713926491831
};
\addplot [semithick, color7, mark=*, mark size=2, mark options={solid}]
table {%
4 0.124583827112721
};
\addplot [semithick, color8, mark=*, mark size=2, mark options={solid}]
table {%
4 0.613115629425961
};
\draw (axis cs:4.1,0.0163906898466801) node[
  scale=0.7,
  anchor=west,
  text=black,
  rotate=0.0
]{data balance};
\draw (axis cs:4.1,0.166143088534487) node[
  scale=0.7,
  anchor=west,
  text=black,
  rotate=0.0
]{softmax entropy};
\draw (axis cs:4.1,0.248573367706262) node[
  scale=0.7,
  anchor=west,
  text=black,
  rotate=0.0
]{kNN embedding};
\draw (axis cs:4.1,0.32632842626898) node[
  scale=0.7,
  anchor=west,
  text=black,
  rotate=0.0
]{learned density};
\draw (axis cs:4.1,0.430206034889669) node[
  scale=0.7,
  anchor=west,
  text=black,
  rotate=0.0
]{OoD training};
\draw (axis cs:4.1,0.37) node[
  scale=0.7,
  anchor=west,
  text=black,
  rotate=0.0
]{Bayesian DeepLab};
\draw (axis cs:4.1,0.29) node[
  scale=0.7,
  anchor=west,
  text=black,
  rotate=0.0
]{Dirichlet DeepLab};
\draw (axis cs:4.1,0.652713926491831) node[
  scale=0.7,
  anchor=west,
  text=black,
  rotate=0.0
]{outlier head};
\draw (axis cs:4.1,0.124583827112721) node[
  scale=0.7,
  anchor=west,
  text=black,
  rotate=0.0
]{resynthesis};
\draw (axis cs:4.1,0.613115629425961) node[
  scale=0.7,
  anchor=west,
  text=black,
  rotate=0.0
]{   diss. ensemble};
\end{axis}

\end{tikzpicture}

%% file: good_and_bad_examples.tex
\begin{figure*}[htb!]
\centering
\def\iwidth{.19\linewidth}
\def\iicol{\iwidth*2.2}
\def\iheight{1.65}
\def\iiheight{\iheight*2.4}
\def\iiiheight{\iheight*3.4}
\def\iiiiheight{\iheight*4.4}
\def\vheight{\iheight*5.8}
\def\viheight{\iheight*6.8}
\def\viiheight{\iheight*7.8}

\begin{tikzpicture}[
method/.style={label={west:{\parbox[c]{1.5cm}{\raggedleft \scriptsize{#1}}}}}]
\node at (0, 1.1) {\scriptsize successful example\strut};
\node at (\iicol, 1.1) {\scriptsize incorrect example\strut};
    
\node[inner sep=0pt, method=Softmax Entropy] at (0,0)
    {\includegraphics[width=\iwidth]{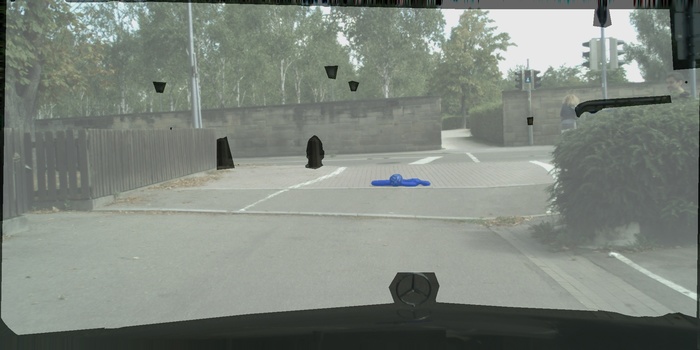}\includegraphics[width=\iwidth]{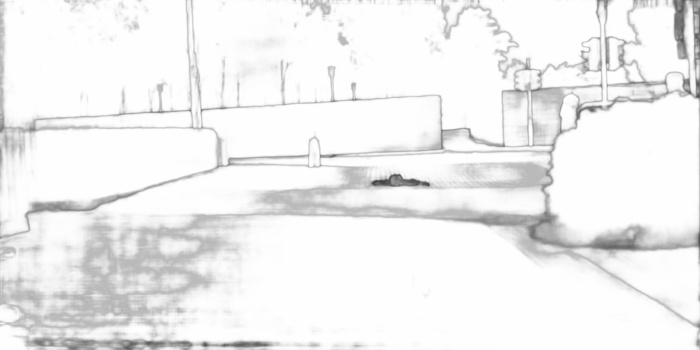}};
\node[inner sep=0pt] at (\iicol,0)
    {\includegraphics[width=\iwidth]{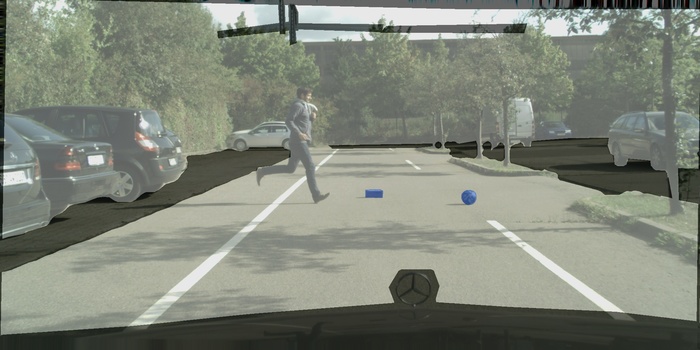}\includegraphics[width=\iwidth]{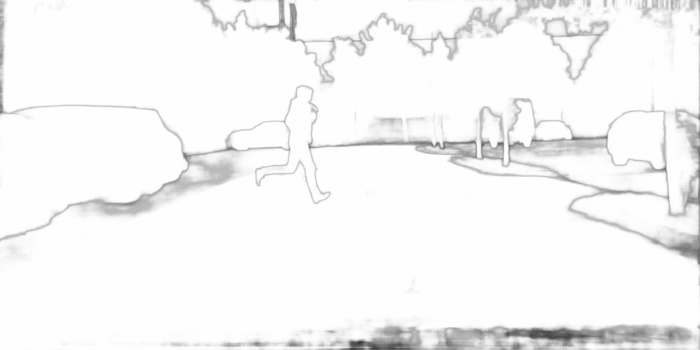}};
\draw[color=red!60, very thick] (\iicol+48.5,-.1) circle (0.22);
\draw[color=red!60, very thick] (\iicol+65,-.1) circle (0.22);

\node[inner sep=0pt, method=Void Classifier] at (0,-\iheight)
    {\includegraphics[width=\iwidth]{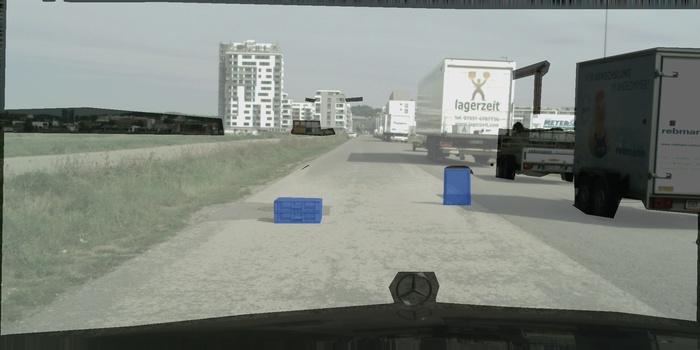}\includegraphics[width=\iwidth]{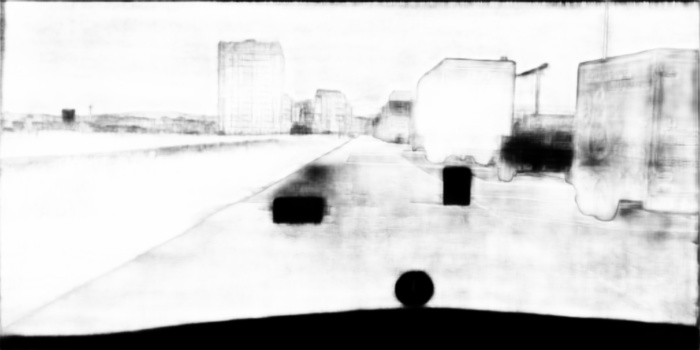}};
\node[inner sep=0pt] at (\iicol,-\iheight)
    {\includegraphics[width=\iwidth]{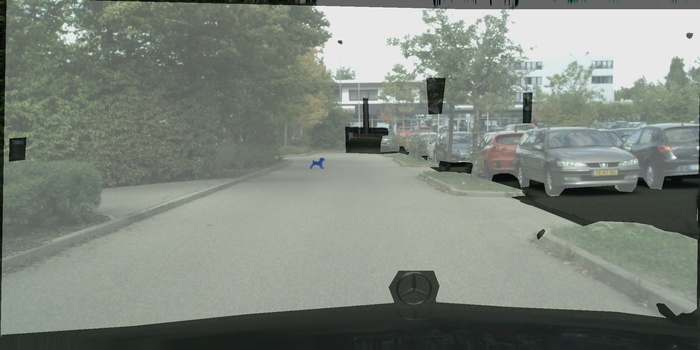}\includegraphics[width=\iwidth]{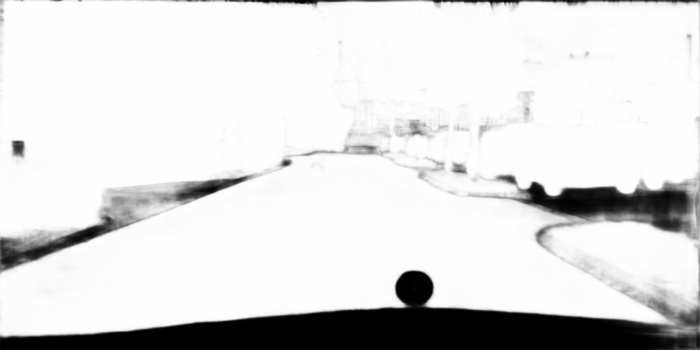}};
\draw[color=red!60, very thick] (\iicol+43,-\iheight+.05) circle (0.2);

\node[inner sep=0pt, method=Bayesian DeepLab] at (0,-\iheight*2)
    {\includegraphics[width=\iwidth]{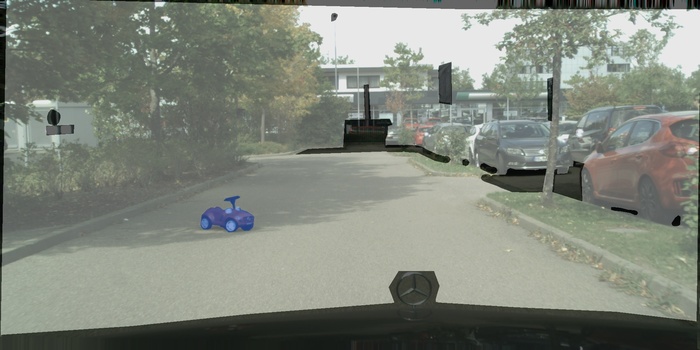}\includegraphics[width=\iwidth]{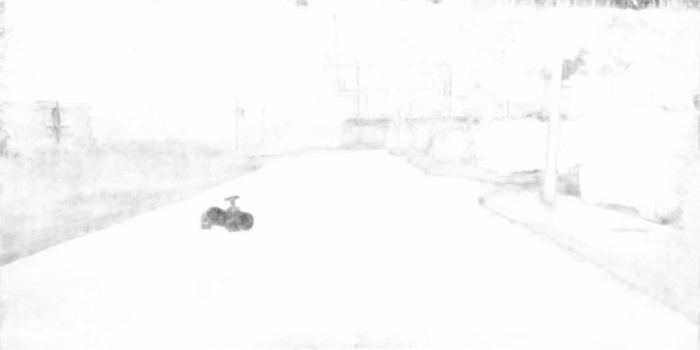}};
\node[inner sep=0pt] at (\iicol,-\iheight*2)
    {\includegraphics[width=\iwidth]{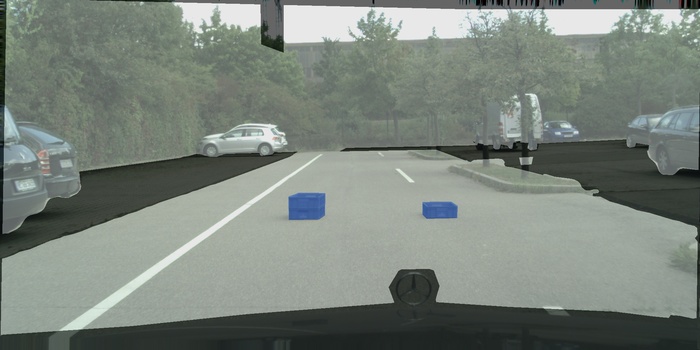}\includegraphics[width=\iwidth]{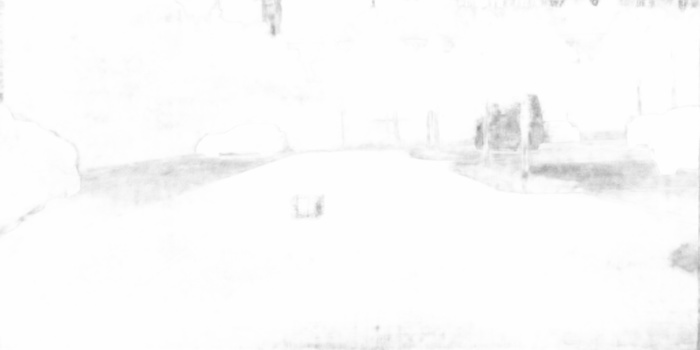}};
\draw[color=red!60, very thick] (\iicol+42,-\iheight*2-.12) circle (0.22);
\draw[color=red!60, very thick] (\iicol+60,-\iheight*2-.12) circle (0.22);
    
\node[inner sep=0pt, method=Dirichlet DeepLab] at (0,-\iheight*3)
    {\includegraphics[width=\iwidth]{laf_0040_overlay.jpg}\includegraphics[width=\iwidth]{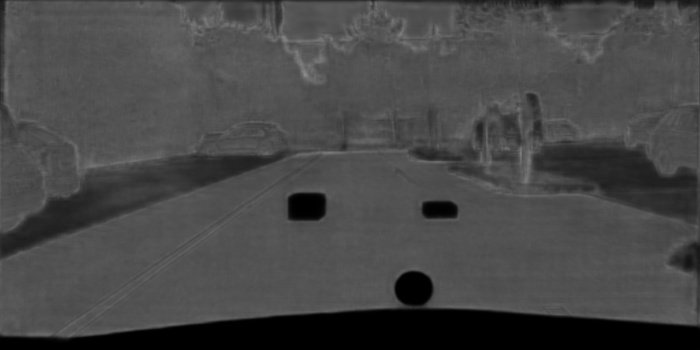}};
\node[inner sep=0pt] at (\iicol,-\iheight*3)
    {\includegraphics[width=\iwidth]{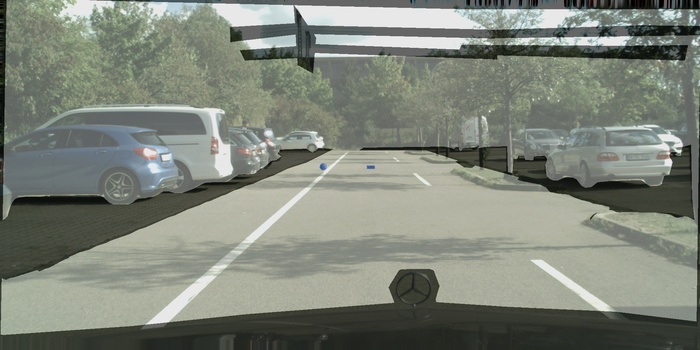}\includegraphics[width=\iwidth]{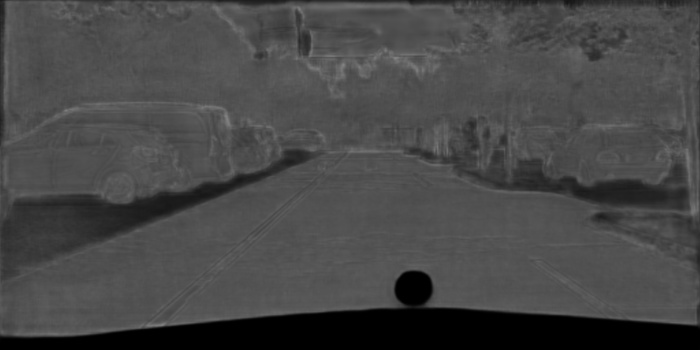}};
\draw[color=red!60, very thick] (\iicol+47,-\iheight*3+.1) circle (0.3);
    
\node[inner sep=0pt, method=kNN density] at (0,-\iheight*4)
    {\includegraphics[width=\iwidth]{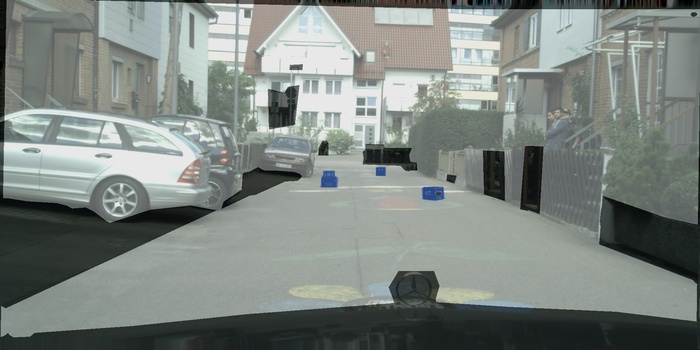}\includegraphics[width=\iwidth]{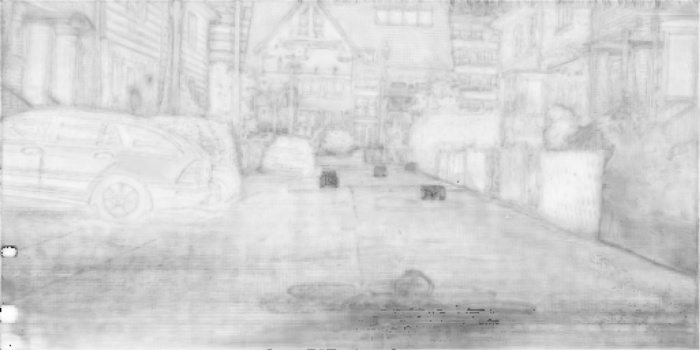}};
\node[inner sep=0pt] at (\iicol,-\iheight*4)
    {\includegraphics[width=\iwidth]{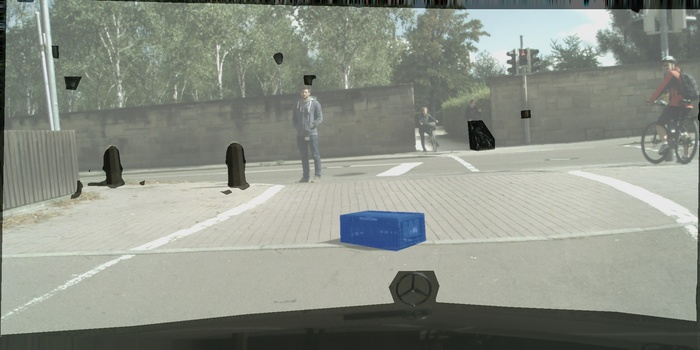}\includegraphics[width=\iwidth]{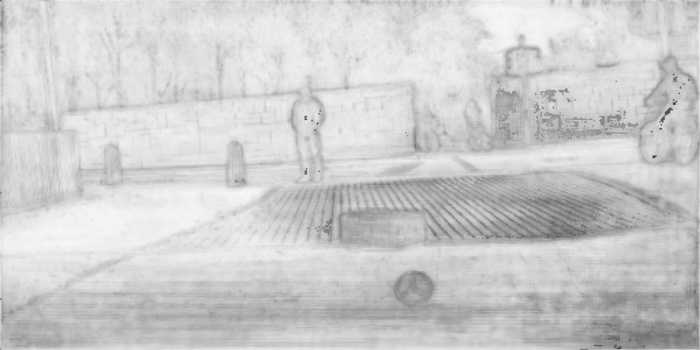}};
\draw[color=red!60, very thick] (\iicol+51,-\iheight*4-.2) circle (0.35);
    
\node[inner sep=0pt, method=learned density regression] at (0,-\iheight*5)
    {\includegraphics[width=\iwidth]{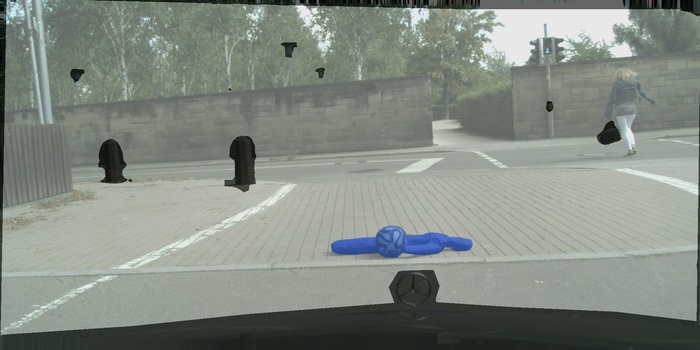}\includegraphics[width=\iwidth]{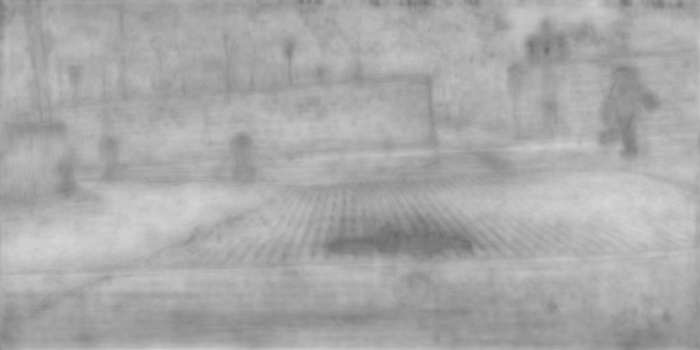}};
\node[inner sep=0pt] at (\iicol,-\iheight*5)
    {\includegraphics[width=\iwidth]{laf_0057_overlay.jpg}\includegraphics[width=\iwidth]{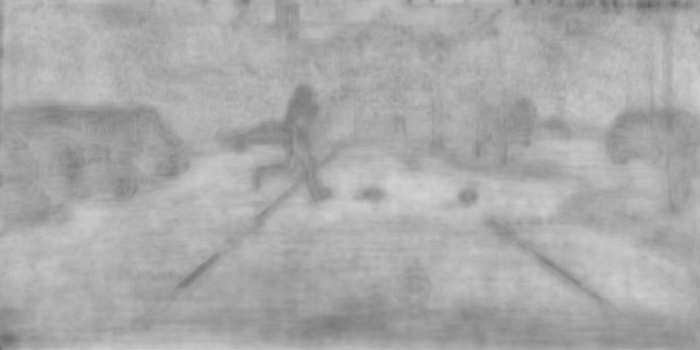}};
\draw[color=red!60, very thick] (\iicol+48.5,-\iheight*5-.1) circle (0.22);
\draw[color=red!60, very thick] (\iicol+65,-\iheight*5-.1) circle (0.22);
    
\node[inner sep=0pt, method=Image Resynthesis] at (0,-\iheight*6)
    {\includegraphics[width=\iwidth]{laf_0035_overlay.jpg}\includegraphics[width=\iwidth]{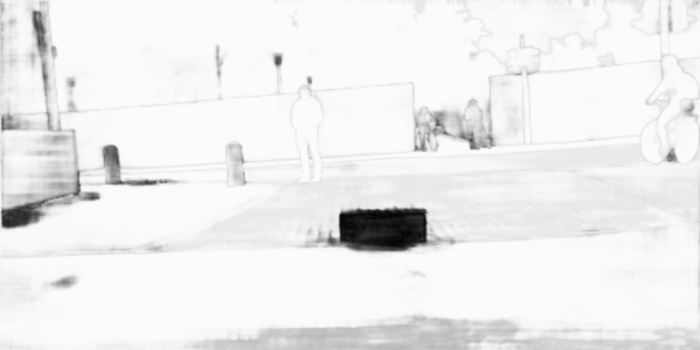}};
\node[inner sep=0pt] at (\iicol,-\iheight*6)
    {\includegraphics[width=\iwidth]{laf_0057_overlay.jpg}\includegraphics[width=\iwidth]{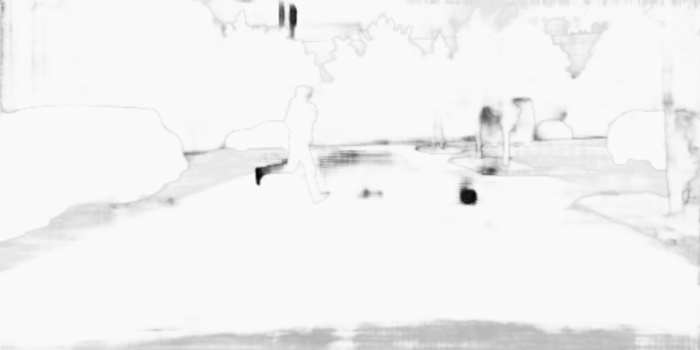}};
\draw[color=red!60, very thick] (\iicol+48.5,-\iheight*6-.1) circle (0.25);
    
\node[inner sep=0pt, method=Outlier Head Combined] at (0,-\iheight*7)
    {\includegraphics[width=\iwidth]{laf_0035_overlay.jpg}\includegraphics[width=\iwidth]{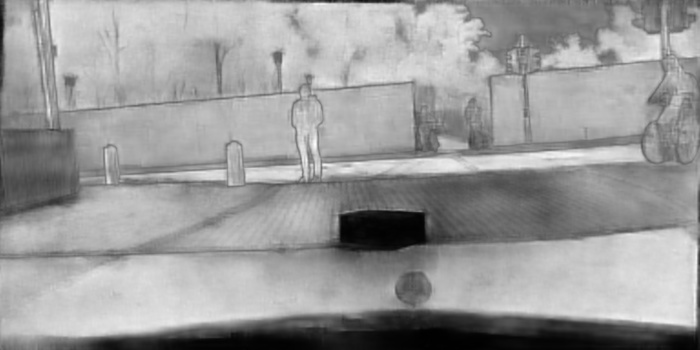}};
\node[inner sep=0pt] at (\iicol,-\iheight*7)
    {\includegraphics[width=\iwidth]{laf_0029_overlay.jpg}\includegraphics[width=\iwidth]{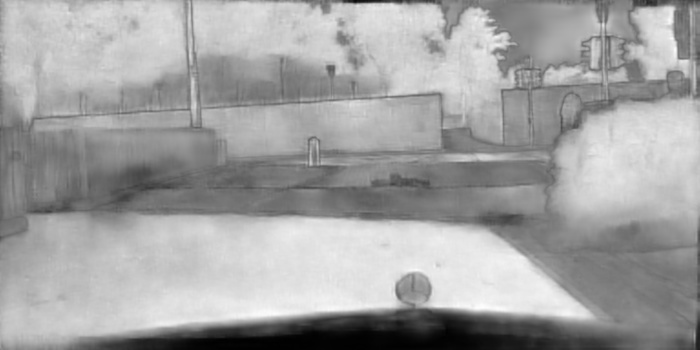}};
\draw[color=red!60, very thick] (\iicol+50,-\iheight*7) circle (0.3);

\node[inner sep=0pt, method=Synboost] at (0,-\iheight*8)
    {\includegraphics[width=\iwidth]{laf_0088_overlay.jpg}\includegraphics[width=\iwidth]{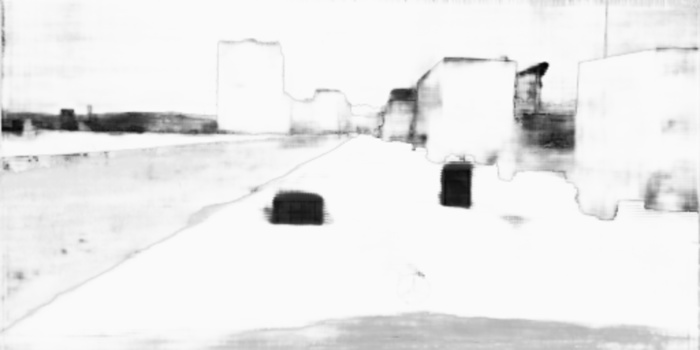}};
\node[inner sep=0pt] at (\iicol,-\iheight*8)
    {\includegraphics[width=\iwidth]{laf_0057_overlay.jpg}\includegraphics[width=\iwidth]{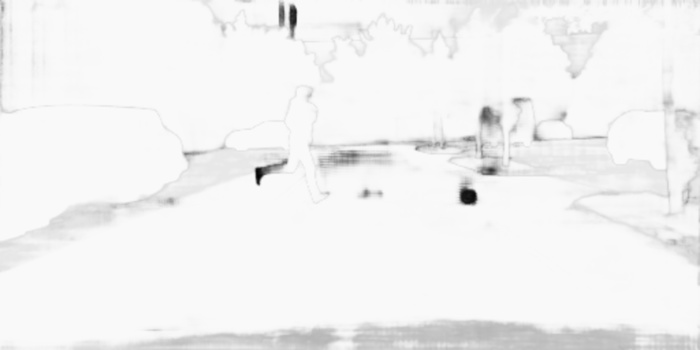}};
\draw[color=red!60, very thick] (\iicol+48.5,-\iheight*8-.1) circle (0.25);
\end{tikzpicture}
\caption{\textbf{Successful and failed examples} for all methods on the Fishyscapes Lost \& Found dataset. Input images overlayed with the evaluation labels are on the left, predicted anomaly scores on the right of each example pair. For every method, we show the best variant. The red circles highlight anomalies that are missed by the method or indistinguishable from noise.}
\label{fig:good-and-bad}
\vspace{-4mm}
\end{figure*}

%% file: appendix.tex
Here we provide additional experimental evaluations as well as details on the proposed datasets and the evaluated methods.

\section{Misclassification Detection}
\label{sec:misclassification}
Additionally to anomaly detection, we test some methods on the detection of misclassifications from the semantic segmentation output. Misclassification detection is another proxy classification task that correlates with uncertainty. However, misclassification mixes uncertainty from
\begin{itemize}[label={--},leftmargin=*, nosep]
    \item noise in the input (\emph{aleatoric} uncertainty) 
    \item model uncertainty
    \item shifts in data balance (softmax classification implicitly learns a prior distribution of the classes over the training set) 
\end{itemize}
Nevertheless, failure detection is an important problem for deployment on autonomous agents, e.g. as part of sensor fusion mechanisms, and misclassification detection is used in different related work~\cite{Hendrycks2016-ua,Lakshminarayanan2017-zk,Guo2017-kg,Jiang2018-rl} to benchmark uncertainty estimates.

\begin{figure*}
\centering
\def\iwidth{.19\linewidth}
\def\iheight{1.8}
\def\iiheight{3.65}
\begin{tikzpicture}
\node[rotate=90] at (-2,-\iheight) {\scriptsize\strut Fishyscapes Misclassifications};

\node[inner sep=0pt, label=\scriptsize{Input\strut}] at (0,0)
    {\includegraphics[width=\iwidth]{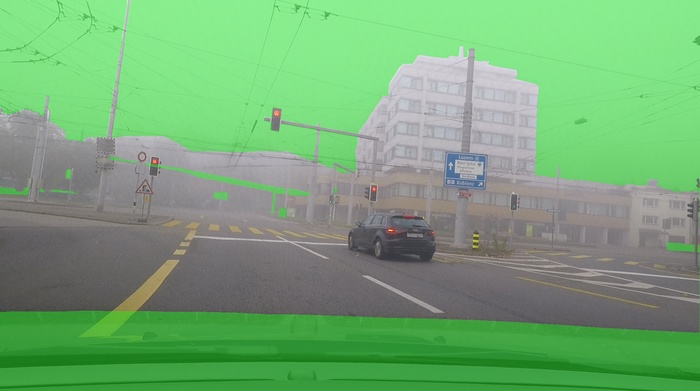}};
\node[inner sep=0pt, label=\scriptsize{Prediction\strut}] () at (\iwidth,0)
    {\includegraphics[width=\iwidth]{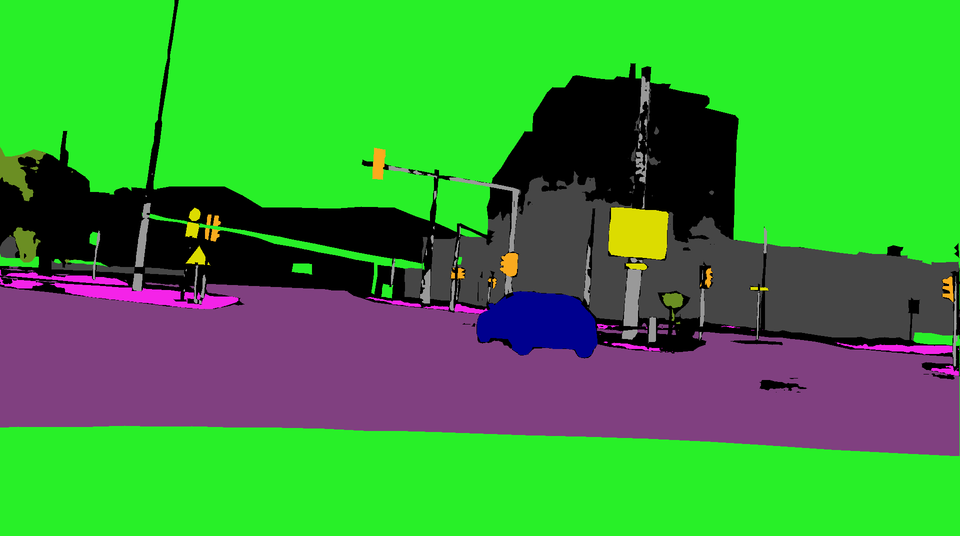}};
\node[inner sep=0pt, label=\scriptsize{Void Entropy\strut}] () at (\iwidth*2,0)
    {\includegraphics[width=\iwidth]{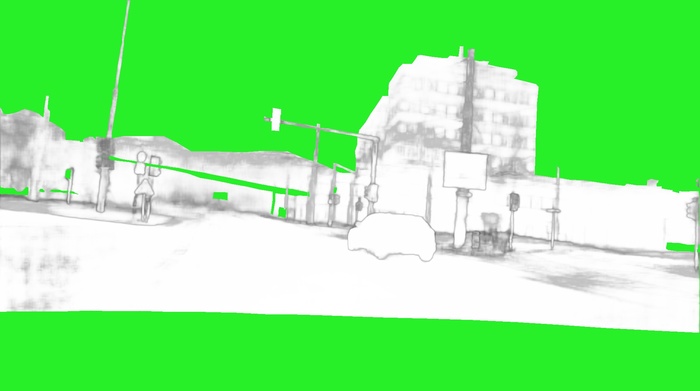}};
\node[inner sep=0pt, label=\scriptsize\strut Prediction] () at (\iwidth*3,0)
    {\includegraphics[width=\iwidth]{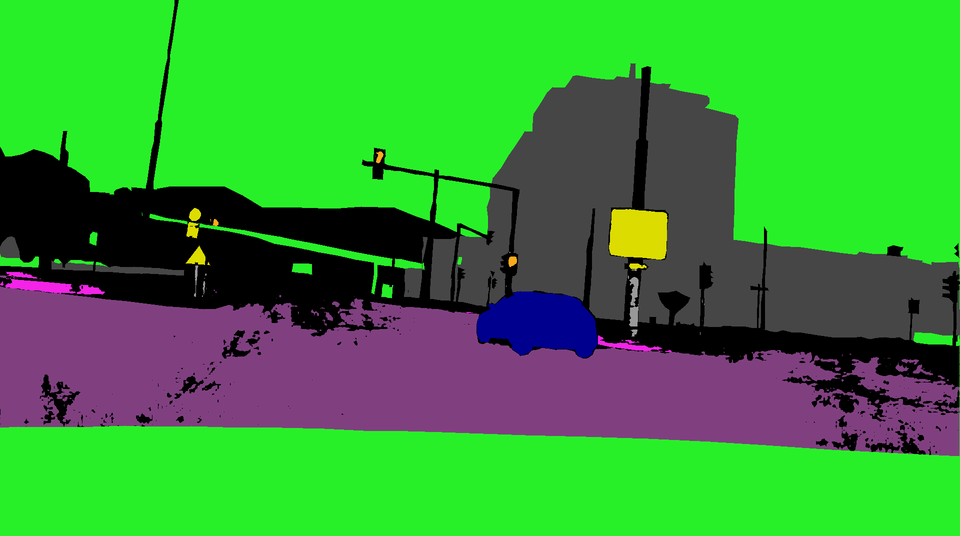}};
\node[inner sep=0pt, label=\scriptsize\strut Bayesian Predictive Entropy] () at (\iwidth*4,0)
    {\includegraphics[width=\iwidth]{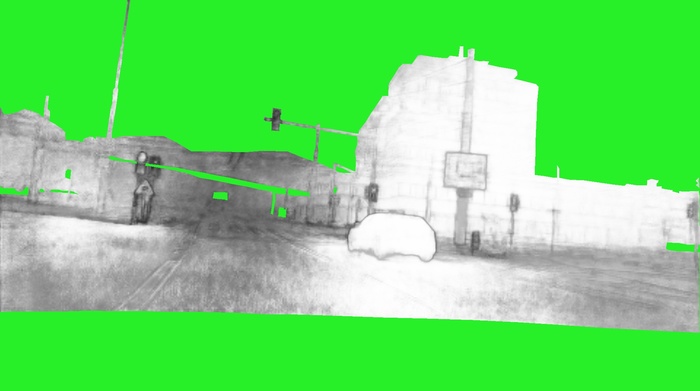}};
    
\node[inner sep=0pt] at (0,-\iheight)
    {\includegraphics[width=\iwidth]{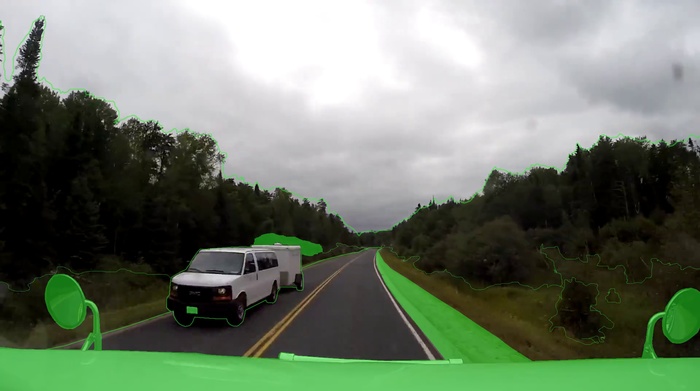}};
\node[inner sep=0pt] () at (\iwidth,-\iheight)
    {\includegraphics[width=\iwidth]{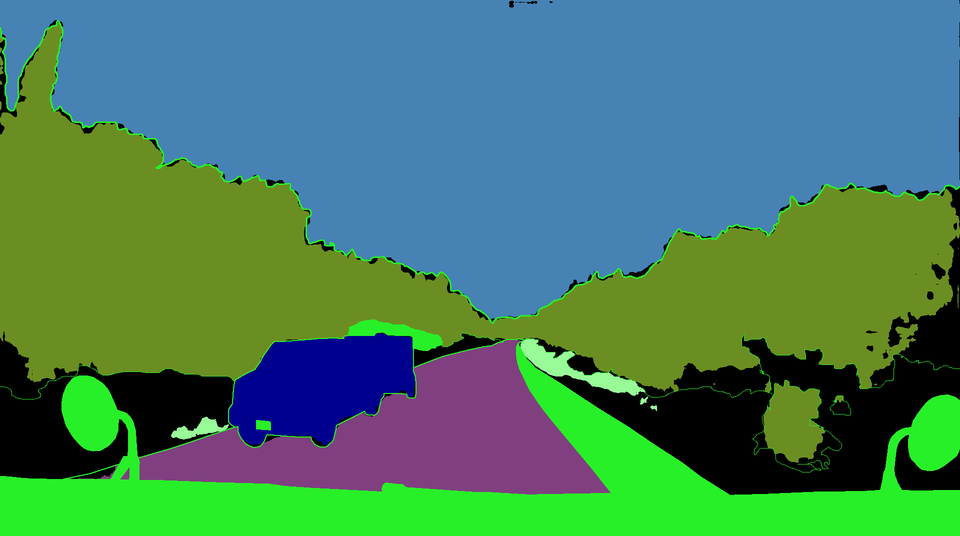}};
\node[inner sep=0pt] () at (\iwidth*2,-\iheight)
    {\includegraphics[width=\iwidth]{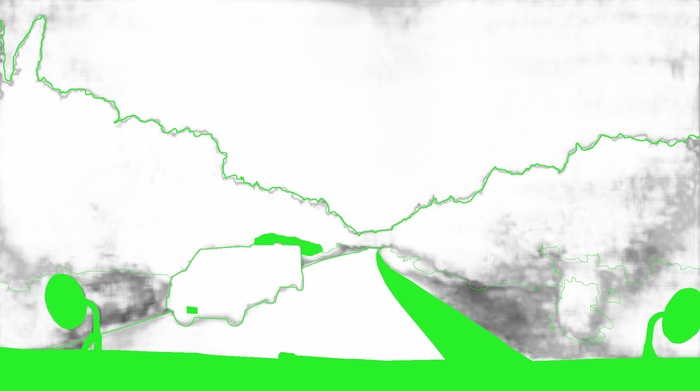}};
\node[inner sep=0pt] () at (\iwidth*3,-\iheight)
    {\includegraphics[width=\iwidth]{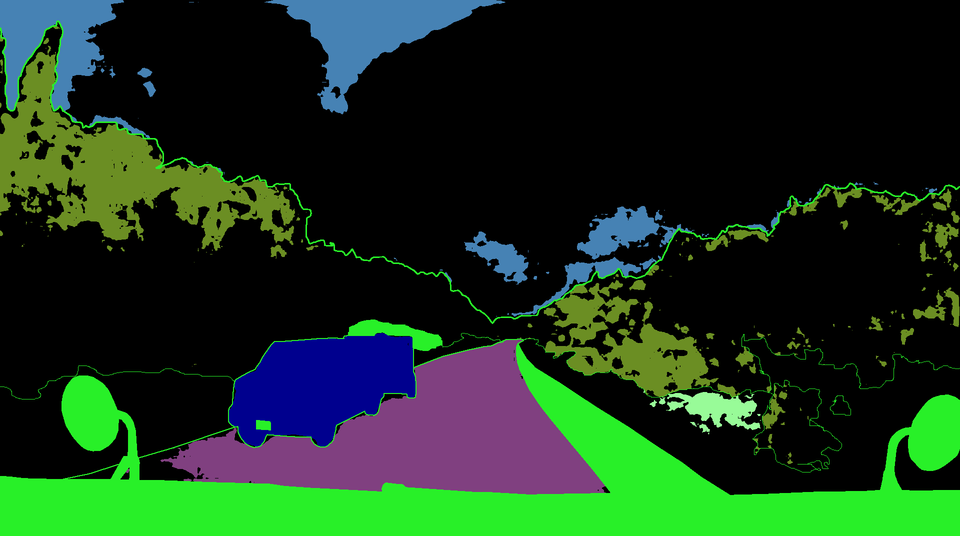}};
\node[inner sep=0pt] () at (\iwidth*4,-\iheight)
    {\includegraphics[width=\iwidth]{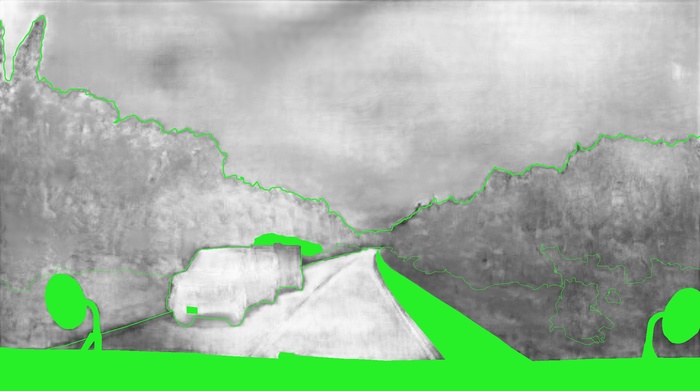}};
    
\node[inner sep=0pt] at (0,-\iiheight)
    {\includegraphics[width=\iwidth]{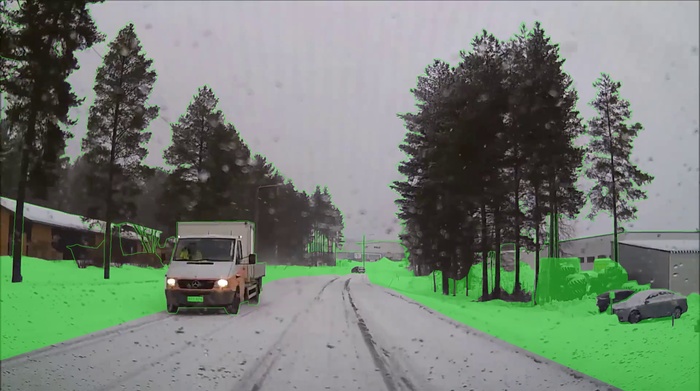}};
\node[inner sep=0pt] () at (\iwidth,-\iiheight)
    {\includegraphics[width=\iwidth]{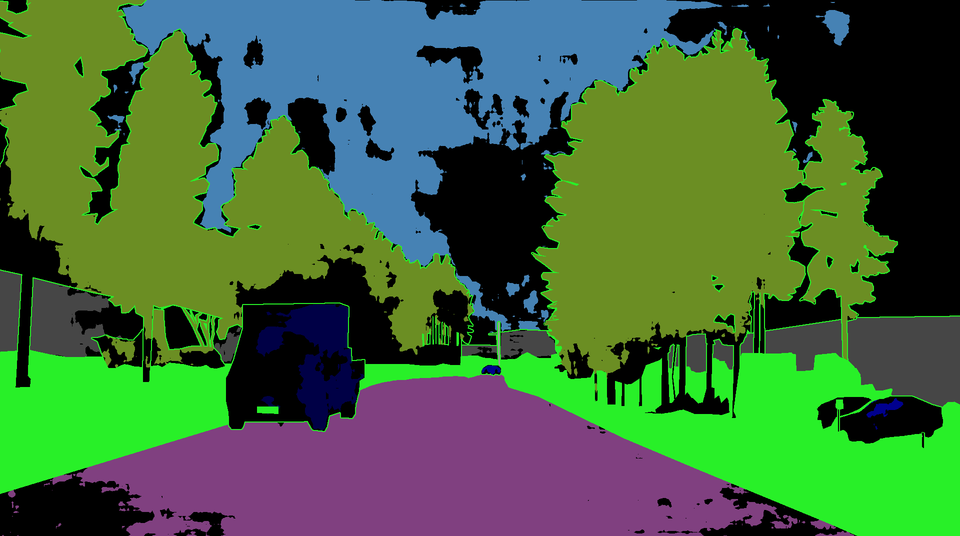}};
\node[inner sep=0pt] () at (\iwidth*2,-\iiheight)
    {\includegraphics[width=\iwidth]{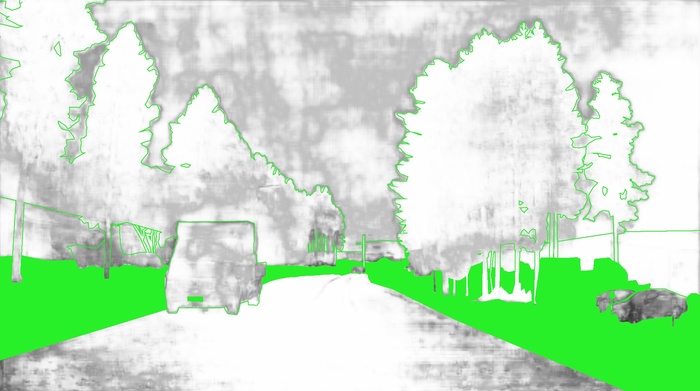}};
\node[inner sep=0pt] () at (\iwidth*3,-\iiheight)
    {\includegraphics[width=\iwidth]{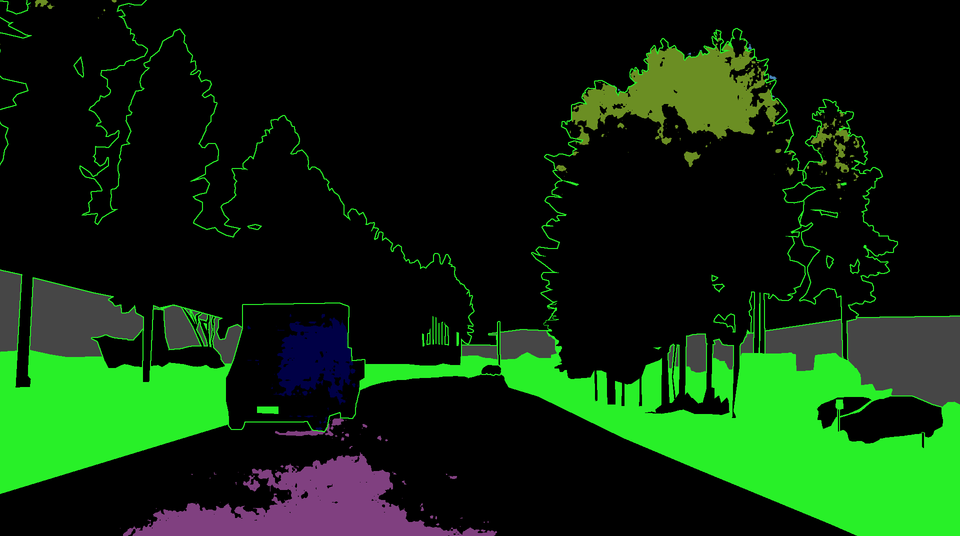}};
\node[inner sep=0pt] () at (\iwidth*4,-\iiheight)
    {\includegraphics[width=\iwidth]{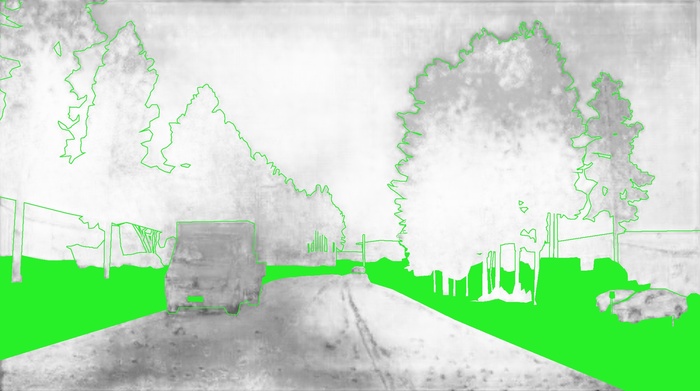}};
\end{tikzpicture}
\caption{\textbf{Qualitative examples of misclassification detection.} Predictions correspond to the uncertainty maps to their right. Misclassifications are marked in black, while ignored \emph{void} pixels are marked in bright green. Better methods should assign a high score (dark) to misclassified pixels. While the different trainings clearly lead to different classification performances, none of the methods captures all the misclassified pixels.}
\label{fig:misclass-overview}
\end{figure*}

\PAR{Dataset.} We test misclassification detection on a diverse mixture of different data sources that introduce sources of uncertainty in the input. From Foggy Driving~\cite{foggycityscapes1}, we select all images. From Foggy Zurich~\cite{foggycityscapes2}, we map classes \emph{sky} and \emph{fence} to \emph{void}, as their labelling is not accurate and sometimes areas that are not visible due to fog are simply labelled \emph{sky}. For WildDash~\cite{Zendel2018-ru}, we use all images. For Mapillary Vistas~\cite{Neuhold2017-ca}, we sample 50 random images from the validation set and apply the label mapping described in Table~\ref{tab:mapillary}.\\
During evaluation all pixels labelled as \emph{void} are ignored.

\begin{table}[ht]
\centering
\footnotesize
\begin{tabular}{ll}
\toprule
mapillary label & used label\\
\midrule
\emph{construction--barrier--fence} & \emph{fence}\\
\emph{construction--barrier--wall} & \emph{wall}\\
\emph{construction--flat--road} & \emph{road}\\
\emph{construction--flat--sidewalk} & \emph{sidewalk}\\
\emph{construction--structure--building} & \emph{building}\\
\emph{human--person} & \emph{person}\\
\emph{human--rider--*} & \emph{rider}\\
\emph{nature--sky} & \emph{sky}\\
\emph{nature--terrain} & \emph{terrain}\\
\emph{nature--vegetation} & \emph{vegetation}\\
\emph{object--support--pole} & \emph{pole}\\
\emph{object--support--utility-pole} & \emph{pole}\\
\emph{object--traffic-light} & \emph{traffic light}\\
\emph{object--traffic-sign--front} & \emph{traffic sign}\\
\emph{object--vehicle--bicycle} & \emph{bicycle}\\
\emph{object--vehicle--bus} & \emph{bus}\\
\emph{object--vehicle--car} & \emph{car}\\
\emph{object--vehicle--motorcycle} & \emph{motorcycle}\\
\emph{object--vehicle--on-rails} & \emph{train}\\
\emph{object--vehicle--truck} & \emph{truck}\\
\emph{marking--*} & \emph{road}\\
anything else & \emph{void}\\
\bottomrule
\end{tabular}
\vspace{3mm}
\caption{Mapping of Mapillary classes onto our used set of classes for misclassification detection.}
\label{tab:mapillary}
\end{table}

\begin{table}[ht]
    \centering
    \scriptsize{
    \setlength\tabcolsep{4pt}
    \begin{tabular}{ll|mcc}
    \toprule
    & & \multicolumn{3}{c}{FS Misclassification}\\
    
    method & score & max J$\,\uparrow$ & AP$\,\uparrow $ & mIoU\\
    \toprule
    Baseline & random uncertainty & 00.0 & 38.9 & 45.5 \\
    \midrule
    \multirow{2}{*}{Softmax}
        & max-probability & 43.6 & 67.4 & \multirow{2}{*}{45.5} \\
        & entropy  & 43.5 & 68.4 &\\
    \midrule
    \ac{ood} training & max-entropy & \textbf{44.3} & 71.3  & 35.8 \\
    \midrule
    \multirow{2}{*}{Bayesian DeepLab}
        & mutual information & 40.7 & 70.4  & \multirow{2}{*}{30.3}\\
        & predictive entropy  & 41.6 & \textbf{73.8} &\\
    \midrule
    Dirichlet DeepLab
        & prior entropy & 29.7 & 65.0 & 37.5 \\
    \midrule
    \multirow{2}{*}{\parbox[b]{8em}{kNN Embedding}}
        & density  & 40.7 & 68.0 & \multirow{2}{*}{45.5}\\
        & relative class density  & 31.7 & 58.0 & \\
    \bottomrule
    \end{tabular}
    }
    \vspace{5pt}
    \caption{\textbf{Misclassification Detection Results}. The gray column marks the primary metric.
    }
    \label{tab:misclassification_results}
    \vspace{-3mm}
\end{table}

\PAR{Evaluated Methods} From the methods evaluated on anomaly detection, we note that the void classifier produces meaningless results for misclassification detection since a high void output score produces the exact misclassification it is detecting. Furthermore, we did not evaluate the learned embedding density.

\PAR{Results} of our evaluation are presented in table~\ref{tab:misclassification_results} and qualitative examples in figure~\ref{fig:misclass-overview}. Differently from anomaly detection, the softmax score is expected to be a good indicator for classification uncertainty, and indeed shows competitive results. For Bayesian DeepLab, we find the predictive entropy to be a better indicator of misclassification, which was also observed by~\cite{Kendall2017-jy}.
The k\ac{nn} density shows results similar to the other methods, hinting that embedding-based methods cannot be entirely classified as \ac{ood}-specific, but may also be able to detect input noise that is very different from the training distribution. Overall, the experiments do not reveal a single method that performs significantly better than others.

\section{Details on the Methods}
\label{sec:details:methods}
In this section we provide implementation details on the evaluated methods to ease the reproducibility of the results presented in this paper.

\subsection{Semantic Segmentation Model}
We use the state-of-the-art model DeepLabv3+~\cite{Chen2018-bp} with Xception-71 backbone, image-level features, and dense prediction cell. When no retraining is required, we use the original model trained on Cityscapes\footnote{\url{https://github.com/tensorflow/models/blob/master/research/deeplab}}.

\subsection{Softmax}
ODIN~\cite{Liang2017-mj} applies input preprocessing and temperature scaling to improve the \ac{ood} detection ability of the maximum softmax probability. Early experiments on Fishyscapes showed that (i) temperature scaling did not improve much the results of this baseline, and (ii) input preprocessing w.r.t.\ the softmax score is not possible due to the limited GPU memory and the large size of the DeepLab model. As the maximum probability is anyway not competitive with respect to the other methods, we decided to not further develop that baseline.




\subsection{Bayesian DeepLab}
We reproduce the setup described by Mukhoti \& Gal~\cite{Mukhoti2018-af}. As such, we use the Xception-65 backbone pretrained on ImageNet, and insert dropout layers in its middle flow. We train for 90k iterations, with a batch size of 16, a crop size of $513 \times 513$, and a learning rate of $7\cdot10^{-3}$ with polynomial decay.

\subsection{Dirichlet DeepLab}
Following Malinin \& Gales~\cite{Malinin2018-pl}, we interpret the output logits of DeepLab as log-concentration parameters $\bm{\alpha}$ and train with the loss described by Equation~(\ref{eq:dirichlet}) and implemented with the TensorFlow Probability~\cite{dillon2017tensorflow} framework. For the first term, the target labels are smoothed with $\epsilon=0.01$ and scaled by $\alpha_0=100$ to obtain target concentrations. To ensure convergence of the classifier, we found it necessary to downweight both the first and second terms by $0.1$ and to initialize all but the last layer with the original DeepLab weigths.

We also tried to replace the first term by the negative log-likelihood of the Dirichlet distribution but were unable to make the training converge.

\subsection{kNN Embedding}
\PAR{Layer of Embedding.} As explained in Section~\ref{sec:method:baselines}, we had to restrict the kNN queries to one layer. A single layer of the network already has more than 10000 embedding vectors and we need to find k nearest neighbors for all of them. Querying over multiple layers therefore becomes infeasible. To select a layer of the network, we test multiple candidates on the FS Lost \& Found validation set. We experienced that our kNN fitting with \emph{hnswlib}\footnote{\url{https://github.com/nmslib/hnswlib}}~\cite{Malkov2016-fv} was not deterministic, therefore we provide the average performance on the validation set over 3 different experiments. Additionally, we had to reduce the complexity of kNN fitting by randomly sampling 1000 images from Cityscapes instead of the whole training set (2975 images).

For the kNN density, we provide the results for different layers in Table~\ref{tab:knn:layers}.

\begin{table}[htb]
\centering
\footnotesize
\begin{tabular}{lc}
\toprule
DeepLab Layer & \ac{ap} \\
\midrule
\scriptsize{\verb|xception_71/middle_flow/block1/unit_8|} & $1.00\pm .02$ \\
\scriptsize{\verb|xception_71/exit_flow/block2|}          & $1.80\pm .01$ \\
\scriptsize{\verb|aspp_features|}                         & $2.97\pm .47$\\
\scriptsize{\verb|decoder_conv0_0|}                       & $3.84\pm .19$ \\
\scriptsize{\verb|decoder_conv1_0|}                       & $2.46\pm .09$ \\
\bottomrule
\end{tabular}
\vspace{3mm}
\caption{Parameter search of the \textbf{embedding layer for kNN density}. The AP is computed on the validation set of FS Lost \& Found. Based on these results, we use the layer \texttt{decoder\_conv0\_0} in all our experiments.}
\label{tab:knn:layers}
\end{table}

For class-based embedding, we perform a similar search for the choice of layer. The result can be found in Table~\ref{tab:classknn:layers}.

\begin{table}[h!]
\centering
\footnotesize
\begin{tabular}{lc}
\toprule
DeepLab Layer & \ac{ap} \\
\midrule
\scriptsize{\verb|xception_71/middle_flow/block1/unit_8|} & $9.6\pm .0$ \\
\scriptsize{\verb|xception_71/middle_flow/block1/unit_10|} & $9.7\pm .0$ \\
\scriptsize{\verb|xception_71/exit_flow/block2|}          & $9.7\pm .1$ \\
\scriptsize{\verb|aspp_features|}                         & $2.3\pm .7$\\
\scriptsize{\verb|decoder_conv0_0|}                       & $2.8\pm .1$ \\
\scriptsize{\verb|decoder_conv1_0|}                       & $3.1\pm .2$ \\
\bottomrule
\end{tabular}
\vspace{3mm}
\caption{Parameter search of the \textbf{embedding layer for class based relative kNN density}. The AP is computed on the validation set of FS Static. Based on these results, we use the layer \texttt{xception\_71/exit\_flow/block2} in all our experiments.}
\label{tab:classknn:layers}
\end{table}

\PAR{Number of Neighbors.}We select $k$ according to Tables~\ref{tab:knn-k} and~\ref{tab:classknn-k}. All values are measured with the same kNN fitting. As the computational time for each query grows with $k$, small values are preferable. Note that by definition, the relative class density needs a sufficiently high $k$ such that not all neighbors are from the same class. 

\begin{table}[htb]
\centering
\footnotesize
\begin{tabular}{rc}
\toprule
k & \ac{ap} \\
\midrule
      1 & 42.3\\
      2 & 44.6 \\
      5 & 47.7\\
      10 & 50.9 \\
      20 & 52.2\\
      50 & 52.7 \\
      100 & 52.5\\
\bottomrule
\end{tabular}
\vspace{3mm}
\caption{Parameter search for the \textbf{number of nearest neighbors for kNN embedding density}. As computing time increases with $k$, we select $k=20$.}
\label{tab:knn-k}
\end{table}

\begin{table}[htb]
\centering
\footnotesize
\begin{tabular}{rc}
\toprule
k & \ac{ap} \\
\midrule
      5  & 5.4\\
      10 & 6.7 \\
      20 & 7.9\\
      50 & 9.3 \\
      100 & 9.9\\
      200 & 10.0\\
\bottomrule
\end{tabular}
\vspace{3mm}
\caption{Parameter search for the \textbf{number of nearest neighbors for the class based kNN relative density}. As computing time increases with $k$, we select $k=100$.}
\label{tab:classknn-k}
\end{table}

\subsection{Learned Embedding Density}
\label{sec:appendix:methods:learned-embedding-density}

\PAR{Flow architecture.} The normalizing flow follows the simple architecture of Real-NVP. We stack 32 steps, each one composed of an affine coupling layer, a batch normalization layer, and a fixed random permutation. As recommended by~\cite{Kingma2018-jp}, we initialize the weights of the coupling layers such that they initially perform identity transformations.

\PAR{Flow training.} For a given DeepLab layer, we export the embeddings computed on all the images of the Cityscapes training set. The number of such datapoints depends on the stride of the layer, and amounts to 22M for a stride of 16. We keep 2000 of them for validation and testing, and train on the remaining embeddings for 200k iterations, with a learning rate of $10^{-4}$, and the Adam optimizer. Note that we can compare flow models based on how well they fit the in-distribution embeddings, and thus do not require any \ac{ood} data for hyperparameter search.

\PAR{Layer selection.} \ac{ood} data is only required to select the layer at which the embeddings are extracted. The corresponding feature space should best separate \ac{ood} and \ac{id} data, such that \ac{ood} embeddings are assigned low likelihood. We found that it is critical to extract embeddings before ReLU activations, as some dimensions might be negative for all training points, thus making the training highly unstable. We show in Table~\ref{tab:learned:layers} the \ac{ap} on the FS Lost \& Found validation set for different layers. We first observe that we did not achieve training convergence for those layers that showed best results in the k\ac{nn} method. This may be due to the high dimensionality of these layers, and/or because the flow is not well suited to approximate these distributions. We also notice that overall layers in the encoder middle flow work best, while Mukhoti \& Gal~\cite{Mukhoti2018-af} insert dropout layers at this particular stage. While we do not know the reason behind their design decision, we hypothesize the they found these layers to best model the epistemic uncertainty.

\begin{table}[htb]
    \centering
    \footnotesize
    \begin{tabular}{lc}
      \toprule
      DeepLab layer & AP \\
      \midrule
      \scriptsize{\verb|xception_71/entry_flow/block5|} & 1.27 \\
      \scriptsize{\verb|xception_71/middle_flow/block1/unit_4|} & 2.14 \\
      \scriptsize{\verb|xception_71/middle_flow/block1/unit_6|} & 2.38 \\
      \scriptsize{\verb|xception_71/middle_flow/block1/unit_8|} & 2.41 \\
      \scriptsize{\verb|xception_71/middle_flow/block1/unit_10|} & 2.52 \\
      \scriptsize{\verb|xception_71/middle_flow/block1/unit_12|} & 2.22 \\
      \scriptsize{\verb|aspp_features|}                         & - \\
      \scriptsize{\verb|decoder_conv0_0|}                       & 0.16 \\
      \scriptsize{\verb|decoder_conv1_0|}                       & \textbf{2.77} \\
      \bottomrule
    \end{tabular}
    \vspace{3mm}
    \caption{Cross-validation of the \textbf{embedding layer for the learned density.} The AP is computed on the validation set of FS Lost \& Found. Based on these results, we use the layer \texttt{decoder\_conv1\_0} in all our experiments. We could not manage to make the training of the \texttt{aspp\_features} layer converge, most likely due to a very peaky distribution that induces numerical instabilities.}
    \label{tab:learned:layers}
\end{table}

\PAR{Effect of input preprocessing.} As previously reported by~\cite{Liang2017-mj,Lee2018-si}, we observe that this simple input preprocessing brings substantial improvements to the detection score on the test set. We show in Table~\ref{tab:learned:noise} the AP for different noise magnitudes $\epsilon$.

\begin{table}[htb]
    \centering
    \footnotesize
    \begin{tabular}{ccc}
      \toprule
      \multirow{2}{*}{Noise $\epsilon$} & \multicolumn{2}{c}{AP on FS Static} \\
      & validation & test \\
      \midrule
      None & 36.0 & 52.5 \\
      0.1 & 38.4 & - \\
      0.2 & 39.1 & - \\
      0.25 & \textbf{39.2} & \textbf{55.4} \\
      0.3 & \textbf{39.2} & - \\
      0.35 & \textbf{39.2} & - \\
      0.4 & 39.1 & - \\
      0.5 & 39.0 & - \\
      1.0 & 36.6 & - \\
      \bottomrule
    \end{tabular}
    \vspace{3mm}
    \caption{Cross-validation of the \textbf{input preprocessing for the learned density.} Based on these results, we apply noise with magnitude $\epsilon=0.25$ in all our experiments.}
    \label{tab:learned:noise}
\end{table}